\documentclass[reqno,letterpaper]{amsart}
\usepackage{amsmath}	
\usepackage{amssymb}	
\usepackage{amsfonts}	
\usepackage{amsthm}	
\usepackage[centercolon=true]{mathtools}

\usepackage[
  paper  = letterpaper,
  margin=1.5in,
  bottom = 1.25in,
  ]{geometry}

\usepackage{graphicx}

\usepackage[dvipsnames, svgnames]{xcolor}
\colorlet{MyRed}{Crimson!60!DarkRed}
\colorlet{MyBlue}{DodgerBlue!75!black}
\colorlet{MyGreen}{DarkGreen}
\colorlet{MyViolet}{DarkMagenta}

\colorlet{MyLightBlue}{DodgerBlue!20}
\colorlet{MyLightGreen}{MyGreen!20}

\colorlet{PrimalColor}{MyBlue}
\colorlet{PrimalFill}{MyLightBlue}
\colorlet{DualColor}{MyRed}

\colorlet{AlertColor}{MyRed}	
\colorlet{BadColor}{MyRed}	
\colorlet{GoodColor}{MyGreen}	
\colorlet{LinkColor}{MediumBlue}	
\colorlet{MacroColor}{MyViolet}
\colorlet{RevColor}{MediumBlue}	

\usepackage{hyperref}
\hypersetup{
final,
colorlinks=true,
linktoc=all,
pdfstartview=FitH,
breaklinks=true,
pdfpagemode=UseNone,
pageanchor=true,
pdfpagemode=UseOutlines,
plainpages=false,
bookmarksnumbered,
bookmarksopen=false,
bookmarksopenlevel=1,
hypertexnames=true,
pdfhighlight=/O,
frenchlinks,
urlcolor=Maroon,linkcolor=MyBlue!60!black,citecolor=DarkGreen!70!black, 
pdftitle={},
pdfauthor={},
pdfsubject={},
pdfkeywords={},
pdfcreator={pdfLaTeX},
pdfproducer={LaTeX with hyperref}
}

\usepackage[utf8]{inputenc}	
\usepackage[T1]{fontenc}	
\usepackage{dsfont}

\usepackage[sf,mono=false]{libertine}
\usepackage[charter,vvarbb]{newtxmath}
\usepackage[cal=euler]{mathalfa}
\usepackage{anyfontsize} 
\usepackage[table,dvipsnames,svgnames]{xcolor}
\usepackage{booktabs}       
\usepackage{xspace}
\usepackage{siunitx}
\usepackage{amsfonts}       
\usepackage{amsmath}        
\usepackage{multirow}
\usepackage[nameinlink, capitalise]{cleveref}

\usepackage{pifont}
\usepackage{subfigure}
\usepackage{algorithm}
\usepackage{algorithmicx}
\usepackage[noend]{algpseudocode}
\algnewcommand{\Initialize}[1]{%
  \State \textbf{Initialize:}
  \Statex \hspace*{\algorithmicindent}\parbox[t]{.8\linewidth}{\raggedright #1}
}

\definecolor{asparagus}{rgb}{0.53, 0.66, 0.42}
\definecolor{cornflowerblue}{rgb}{0.39, 0.58, 0.93}
\definecolor{mediumtealblue}{rgb}{0.0, 0.33, 0.71}
\definecolor{amethyst}{rgb}{0.6, 0.4, 0.7}
\newcommand{\yl}[1]{{\color{black} #1}} 
\newcommand{\todo}[1]{{\color{black} #1}} 
\makeatletter
\newcommand{\st}[1]{{\color{black} #1}}
\makeatother
\makeatletter
\newcommand{\sv}[1]{{\color{black} #1}}

\makeatother
\newtheorem{assumption}{Assumption}
\newcommand{\defineq}{\triangleq}

\newcommand{\RR}{\mathbf{R}}

\newcommand{\cA}{{\mathcal{A}}}

\newcommand{\cC}{{\mathcal{C}}}
\newcommand{\cD}{{\mathcal{D}}}

\newcommand{\cF}{{\mathcal{F}}}

\newcommand{\cO}{{\mathcal{O}}}
\newcommand{\cP}{{\mathcal{P}}}

\newcommand{\cS}{{\mathcal{S}}}

\newcommand{\cU}{{\mathcal{U}}}
\newcommand{\cV}{{\mathcal{V}}}

\newcommand{\cX}{{\mathcal{X}}}
\newcommand{\cY}{{\mathcal{Y}}}

\newcommand{\expectation}{\mathbb{E}}
\DeclareMathOperator{\indicator}{\mathbbm{1}}
\DeclareMathOperator{\argmin}{argmin}
\DeclareMathOperator{\Jac}{Jac}
\DeclareMathOperator{\proj}{Proj}
\DeclareMathOperator{\clip}{\textbf{Clip}}
\DeclareMathOperator{\stonabla}{\widetilde{\nabla}}
\DeclareMathOperator{\stoerror}{\Delta}

\newcommand{\myparagraph}[1]{\vspace*{0.5em}\par\noindent\textbf{{#1}.}}

\newcommand{\paretofront}{\cP}
\newcommand{\weight}{\lambda}
\newcommand{\model}{w}
\newcommand{\simplex}[1]{\Delta_{#1}}
\newcommand{\inputspace}{\cX}
\newcommand{\outputspace}{\cY}
\newcommand{\attributespace}{\cA}
\newcommand{\attributenumber}{\cS}

\newcommand{\upperobj}{\cU}
\newcommand{\lowerobj}{\cF}
\newcommand{\implicitobj}{\varphi}
\newcommand{\lowervar}{w}
\newcommand{\uppervar}{\lambda}
\newcommand{\dualvar}{v}
\newcommand{\lowersol}{\lowervar^{\star}}
\newcommand{\dualsol}{\dualvar^{\star}}
\newcommand{\starterrordual}{\varepsilon_{\dualvar, 0}}
\newcommand{\filtration}{\mathfrak{S}}
\newcommand{\apgstep}{D}
\newcommand{\agrad}{D^{in}}
\newcommand{\pgstep}{\bar{D}}
\newcommand{\lyap}{\psi}

\newcommand{\constant}{\cC}

\newcommand{\constantdualdrift}{M_{\dualvar^\star}}

\newcommand{\frameworkname}{\textsc{badr}\xspace}
\newcommand{\algoname}{\textsc{badr-gd}\xspace}
\newcommand{\stoalgoname}{{\textsc{badr-sgd}}\xspace}

\newcommand{\fairnessmetric}{\cF\text{air}}

\newcommand{\cmark}{\ding{51}} 
\newcommand{\xmark}{\ding{55}} 

\usepackage{tikz}
\usepackage{fontawesome5}
\usetikzlibrary{
  shapes.geometric,
  shapes.misc,
  positioning,
  fit,
  backgrounds,
  shadows.blur,
  arrows.meta
}

\definecolor{SoftBlue}{RGB}{197,221,243}
\definecolor{SoftGreen}{RGB}{198,234,183}
\definecolor{SoftRed}{RGB}{252,185,186}
\definecolor{SoftOrange}{RGB}{255,218,185}
\definecolor{SoftGray}{RGB}{240,240,240}

\tikzset{
  module/.style={
    draw=black!50, thick,
    rectangle, rounded corners=2mm,
    minimum width=3.5cm,
    minimum height=1cm,     
    fill=#1!50,
    font=\sffamily\bfseries,
    drop shadow={shadow blur steps=5,shadow xshift=0pt,shadow yshift=-1pt}
  },
  helper/.style={
    draw=black!30, thick,
    rectangle, rounded corners=2mm,
    minimum width=2cm,
    minimum height=0.8cm,
    fill=#1!10,
    font=\sffamily,
    align=center
  },
  arr/.style={
    -{Latex[length=3mm]}, thick, draw=black!70,
    font=\sffamily\small
  },
  group/.style={
    draw=black!40, thick,
    rounded corners=3mm,
    inner sep=10pt,
    dashed,
    fill=white!20
  },
  grouplabel/.style={
    font=\sffamily\large\bfseries,
    anchor=south west,
    inner sep=3pt
  }
}

\newcommand{\tbadr}{\texttt{badr}\xspace}

\newcommand{\tpandas}{\texttt{pandas}\xspace}

\newcommand{\tsklearn}{\texttt{scikit-learn}\xspace}

\newcommand{\tjax}{\texttt{jax}\xspace}

\usepackage{listings}

\definecolor{codegreen}{rgb}{0,0.6,0}
\definecolor{codegray}{rgb}{0.5,0.5,0.5}
\definecolor{codepurple}{rgb}{0.58,0,0.82}
\definecolor{backcolour}{rgb}{0.95,0.95,0.92}

\usepackage{pythonhighlight}
\definecolor{keywordcolour}{RGB}{207,34,46}
\definecolor{stringcolour}{RGB}{26,62,115}
\definecolor{literatecolour}{RGB}{20,90,179}
\definecolor{specmethodcolour}{RGB}{137,90,225}
\crefname{example}{example}{examples}
\Crefname{example}{Example}{Examples}

\crefname{theorem}{theorem}{theorems}
\Crefname{theorem}{Theorem}{Theorems}

\crefname{lemma}{lemma}{lemmas}
\Crefname{lemma}{Lemma}{Lemmas}

\crefname{proposition}{proposition}{propositions}
\Crefname{proposition}{Proposition}{Propositions}

\crefname{remark}{remark}{remarks}
\Crefname{remark}{Remark}{Remarks}

\crefname{corollary}{corollary}{corollaries}
\Crefname{corollary}{Corollary}{Corollaries}

\crefname{definition}{definition}{definitions}
\Crefname{definition}{Definition}{Definitions}

\crefname{conjecture}{conjecture}{conjectures}
\Crefname{conjecture}{Conjecture}{Conjectures}

\crefname{axiom}{axiom}{axioms}
\Crefname{axiom}{Axiom}{Axioms}

\expandafter\def\expandafter\normalsize\expandafter{%
    \normalsize%
    \setlength\abovedisplayskip{0pt}%
    \setlength\belowdisplayskip{8pt}%
    \setlength\abovedisplayshortskip{-2pt}%
    \setlength\belowdisplayshortskip{4pt}%
}

\allowdisplaybreaks[4]

\newtheorem{theorem}{Theorem}
\newtheorem{lemma}[theorem]{Lemma} 
\newtheorem{proposition}[theorem]{Proposition} 
\newtheorem{remark}[theorem]{Remark}
\newtheorem{corollary}[theorem]{Corollary}
\newtheorem{definition}[theorem]{Definition}

\usepackage{tikz}	
\usepackage{tikz-cd}	
\usetikzlibrary{calc,patterns}	

\usepackage{array}	
\usepackage{booktabs}	
\usepackage[inline,shortlabels]{enumitem}	
\setlist[1]{topsep=\smallskipamount,itemsep=\smallskipamount,left=\parindent}
\setlist[2]{left=0pt}

\usepackage[sort&compress]{natbib}	

\setcitestyle{numbers,square}

\usepackage{lipsum}
\usepackage{bbm}

\makeatletter
\renewcommand{\tocsection}[3]{%
  \indentlabel{\@ifnotempty{#2}{\bfseries\ignorespaces#1 #2\quad}}\bfseries#3}

\renewcommand{\tocsubsection}[3]{%
  \indentlabel{\@ifnotempty{#2}{\ignorespaces#1 #2\quad}}#3}

\newcommand\@dotsep{4.5}

\def\@tocline#1#2#3#4#5#6#7{\relax
  \ifnum #1>\c@tocdepth
  \else
    \par \addpenalty\@secpenalty\addvspace{#2}%
    \begingroup \hyphenpenalty\@M
    \@ifempty{#4}{%
      \@tempdima\csname r@tocindent\number#1\endcsname\relax
    }{%
      \@tempdima#4\relax
    }%
    \parindent\z@ \leftskip#3\relax
    \advance\leftskip\@tempdima\relax
    \rightskip\@pnumwidth plus1em \parfillskip-\@pnumwidth
    #5\leavevmode\hskip-\@tempdima{#6}\nobreak
    \leaders\hbox{$\m@th\mkern \@dotsep mu\hbox{.}\mkern \@dotsep mu$}\hfill
    \nobreak
    \hbox to\@pnumwidth{\@tocpagenum{\ifnum#1=1\bfseries\fi#7}}\par
    \nobreak
    \endgroup
  \fi}

\AtBeginDocument{%
  \expandafter\renewcommand\csname r@tocindent0\endcsname{0pt}%
}
\def\l@subsection{\@tocline{2}{0pt}{2.5pc}{5pc}{}}

\makeatother
\usepackage[foot]{amsaddr}

\title[Fairness-informed Pareto Optimization : An Efficient Bilevel Framework]{Fairness-informed Pareto Optimization \\ An Efficient Bilevel Framework}
\author[S.~Tanji]{Sofiane Tanji$^{\ast}$}
\address{$^{\ast}$INMA/ICTEAM, Universit\'e Catholique de Louvain, B-1348 Louvain-la-Neuve, Belgium}
\author[S.~Vaiter]{Samuel Vaiter$^{\diamond}$}
\address{$^{\diamond}$Lab. Jean Alexandre Dieudonn\'e, CNRS, Nice, 06000, France}
\author[Y.~Laguel]{Yassine Laguel$^{\dagger}$}
\thanks{Corresponding author: \href{mailto:yassine.laguel@univ-cotedazur.fr}{yassine.laguel@univ-cotedazur.fr}}
\address{$^{\dagger}$Lab. Jean Alexandre Dieudonn\'e, Universit\'e C\^{o}te d'Azur, Nice, 06000, France}

\begin{document}
\begin{abstract}
    \yl{
Despite their promise, fair machine learning methods often yield Pareto-inefficient models, in which the performance of certain groups can be improved without degrading that of others. 
This issue arises frequently in traditional in-processing approaches such as fairness-through-regularization.  
In contrast, existing Pareto-efficient approaches are biased towards a certain perspective on fairness and fail to adapt to the broad range of fairness metrics studied in the literature.
In this paper, we present \frameworkname, a simple framework to recover the \emph{optimal} Pareto-efficient model for any fairness metric. 
Our framework recovers its models through a Bilevel Adaptive Rescalarisation procedure. 
\sv{The lower level is a weighted empirical risk minimization task where the weights are a convex combination of the groups, while the upper level optimizes the chosen fairness objective.}
We equip our framework with two novel large-scale, single-loop algorithms, \algoname and \stoalgoname, and establish their convergence guarantees. 
We release \tbadr, an open-source Python toolbox implementing our framework for a variety of learning tasks and fairness metrics.
Finally, we conduct extensive numerical experiments demonstrating the advantages of \frameworkname over existing Pareto-efficient approaches to fairness.
}
\end{abstract}
\maketitle

\section{Introduction}\label{sec:introduction}
\st{
Data-driven decision systems are now commonly used in high-stakes domains such as employment, education, law enforcement, and health care \citep{citron2014scored}.
\yl{Their widespread deployment} which assists and sometimes takes over human decisions, has been accompanied by increased concerns about unfairness and bias based on sensitive attributes such as gender, race or religion \citep{cross2024bias, mehrabi2021survey, obermeyer2019dissecting}.
\yl{
Motivated by evidence that learned models can perform unevenly across sensitive groups \citep{mehrabi2021survey}, research on algorithmic fairness has proposed a range of formulations and metrics \citep{hardt2016equality,metric_kearns,feldman2015certifying} to quantify and alleviate the bias of standard learning algorithms.
Yet, optimizing these criteria typically comes at a cost in predictive performance, inducing an inherent fairness-accuracy trade-off \citep{menon2018cost}.
This tension is particularly pronounced in in-processing methods that incorporate constraints or penalties into the learning algorithm \citep{preprocess_fairglm}, often yielding models that are \emph{inefficient} from a performance standpoint.
Simply put, it is often possible to improve the performance of one group without harming others, indicating that the model is not operating on the \emph{Pareto front} of groups performances.
}

\yl{
This limitation has spurred interest in group fairness formulations \citep{dwork2012fairness} that consider aggregations of group performance alternative to the standard empirical risk minimization (ERM) objective.
For instance, the minimax group fairness framework \citep{martinez2020minimax} minimizes the worst-group loss, ensuring that no group's performance can be improved without degrading another's.
More broadly, fairness risk measures such as the Conditional Value at Risk (CVaR) \citep{rockafellar2000optimization} interpolate between the robust minimax approach and the expectation-based ERM loss.
While these methods guarantee Pareto efficiency, as detailed in Section~\ref{subsec:minmax}, their impact on specific fairness metrics remains unpredictable and uncontrollable.
}

\yl{
In this work, we introduce a general framework -- \frameworkname for Bilevel Adaptive Rescalarization -- that learns on the Pareto front of group performance the model that optimizes a user-specified fairness metric.
We cast the task as a bilevel formulation: the lower level enforces Pareto efficiency across groups, and the upper level selects the point that minimizes the chosen fairness metric.
We propose \algoname, a large-scale single-loop stochastic algorithm designed specifically for this bilevel problem, and prove convergence under standard assumptions.
We release an open-source Python toolbox that implements our approach for several fairness metrics.
Experiments on real-world datasets show improvements over existing methods on the targeted fairness metric while preserving Pareto-efficient group performance.
}
}

\subsubsection*{Related work}\yl{

\paragraph{Multi-objective optimization} 

\yl{
In this work, we tackle fair machine learning within a multi-objective optimization framework. 
That is, group performances are measured through their respective losses and success is measured through Pareto efficiency \citep{ehrgott2005multicriteria}. 
The set of Pareto efficient models, coined the Pareto front can be searched through different methods, including scalarization methods \citep{ehrgott2005multicriteria,eichfelder2008adaptive}, the $\varepsilon$-constraint method~\citep{mavrotas2009effective}, Particle Swarm Optimization~\citep{coello2004handling}, etc. 
Here, we build upon a standard scalarization method, the weighted sum approach~\citep{zadeh1963optimality, geoffrion1968proper}, which swaps the multi-objective formulation for an arbitrary convex combination of the objectives.
This approach is widely used in practice and is guaranteed under mild assumptions to output Pareto efficient solutions~\citep{geoffrion1968proper,ehrgott2005multicriteria}.
Yet the interplay between the convex weights and the quality of the resulting solution remains poorly understood in general \citep{das1997closer,ehrgott2005multicriteria}.
Some methods attempt to explore the entire Pareto front by varying the weights~\citep{kim2004adaptive,ryu2019multiobjective}, while other methods compare weights based on their relative impact on the objective~\citep{ayan2023comprehensive}. 
In this work, we present a novel data-driven approach for the weight selection in the weighted sum method.
Our method optimizes a user-specified fairness metric over the Pareto front of group performances, thereby tailoring the solution to the stakeholders' fairness concerns.
Beyond fair learning, our approach is applicable to general multi-objective problems whenever a utility function is given to measure the quality of pareto-efficient solutions. 
}

\paragraph{Efficient Fair Learning} 

\st{
To mitigate unfairness, three families of interventions are common: pre-processing \citep{zemel2013learning, xu2023fair, ruoss2020learning, lahoti2019ifair, calmon2017optimized}, which modifies the data or labels before training (e.g., reweighting, resampling, or learning fair representations); in-processing \st{see, e.g., \citep{zafar2017fairness, zafar2019fairness, donini2018empirical, cruz2022fairgbm, cotter2019optimization, celis2019classification},} which 
\yl{alter the training process itself to promote fairness in the learned model}; and post-processing \citep{lohia2019bias}, which adjusts scores or decision thresholds of a trained model to satisfy a target criterion.
Within in-processing, the arguably most standard approach is to incorporate fairness requirements in the optimization problem with either explicit fairness constraints or 
\yl{penalization} terms \citep{preprocess_fairglm}.
By formulating the initial optimization problem to penalize violations of a chosen fairness criterion (e.g., adding a constraint on demographic parity or a penalty term for equalized odds), the model is encouraged to be fairer during learning.
\yl{However, from a Pareto perspective this approach cannot guarantee efficient solutions -- see our discussion in Section~\ref{sec:preliminaries}: a model trained with such penalties may still be improvable for some groups without harming others.

Most Pareto-efficient methods in the fairness literature rely on data reweighting and have been cast as both pre- and in-processing techniques.
Regarding pre-processing methods,~\citep{sivarajkumar2023fair} proposes for instance to reweight data per group with weights inversely proportional to group frequencies. 
We note also several works consider reweighting jointly per group and class -- e.g.~\citep{kamiran2012data} -- to reduce certain fairness metrics such as demographic parity. 
Yet allowing weight to depend on class takes the method out of our multi-objective framework : the resulting weighted loss is no longer a convex combination of group losses, and Pareto efficiency is not guaranteed.
Regarding in-processing methods, in the same vein as~\citep{kamiran2012data}, some training procedures dynamically estimate weights jointly on group and class to mitigate certain fairness metrics~\citep{krasanakis2018adaptive,jiang2019identifyingcorrectinglabelbias}.
Alternatively, several methods have considered adversarial reweighting schemes to promote fairness during training.
This line of works, designated as minimax fairness~\citep{martinez2020minimax,hashimoto2018fairness,diana2021minimax,abernethy2020active}, has expanded to distributionally robust optimization approaches~\citep{williamson2019fairness,pillutla2024federated,tsang2025unified}, and has the benefit of systematically yielding Pareto efficient models,  see \cref{subsec:minmax} for details.
Yet, given a specific fairness metric, one can hardly guarantee the quality of these robust models. 
}

\yl{
Finally we note that several works have explored bilevel optimization to dynamically adjust group-wise weights. 
For instance~\citep{ozdayi2021fair,roh2020fairbatch} consider a bilevel formulation where the upper level minimizes arbitrary fairness metrics but the lower level allows for weights to depend on both group and class, thereby losing Pareto efficiency.
Alternatively, \citep{martinez2020minimax, abernethy2020active, shekhar2021adaptive} propose a bilevel framework where the lower level minimizes a weighted group loss, and the upper level optimizes the worst-case group loss.
Furthermore, some bilevel methods, such as those proposed in~\citep{kamani2021pareto}, have been developed to address multi-objective problems that aim to balance different fairness metrics and accuracy. 
These approaches lead to formulations and considerations that differ from ours : competing loss functions do not share the same regularity properties and are typically supported by weaker convergence guarantees.
Finally, we note none of these papers provide large-scale optimization methods with convergence guarantees that effectively solve their bilevel formulations.
}
}
}

\paragraph{Bilevel Optimization} 
\yl{
Bilevel optimization~\citep{dempe2002foundations,dempe2020bilevel} plays an increasing role in machine learning applications including including hyperparameter optimization~\citep{bertrand2022implicit}, meta-learning~\citep{chayti2024new}, or fair machine learning, as discussed in the above paragraph. 
Among existing approaches for bilevel learning problems, we note two main categories: penalization-based approaches~\citep{kwon2023fully} that transforms the bilevel problem into a single-level one by adding a penalty term to the upper-level objective, and implicit methods~\citep{domke2012generic,maclaurin2015gradient} that approximate hypergradients through implicit differentiation or unrolling techniques.
Early deterministic works relied on double-loop algorithms, where the inner loop iteratively solved the lower-level problem to near-optimality~\citep{pedregosa2016hyperparameter,ghadimi2018approximation,gong2024accelerated}.
More recent studies have proposed single-loop schemes that update both levels simultaneously~\citep{dagreou2022framework,chen2022single}, thereby avoiding the need to tune the number of inner iterations.
To enable implicit differentiation and theoretical guarantees, these methods usually assume the lower-level problem is smooth (with Lipschitz-continuous Hessians) and strongly convex~\citep{pedregosa2016hyperparameter,ghadimi2018approximation,dagreou2022framework}, or at least satisfy the Polyak--\L ojasiewicz condition~\citep{ji2021bilevel}.
Yet, in our setting, the multiplicative interaction between model weights and group-level losses induces unbounded cross derivatives, violating the smoothness and convexity assumptions commonly required in the analysis.
}

\subsubsection*{Contributions}\st{
We have four sets of contributions.
\myparagraph{The \frameworkname framework}
We introduce the \frameworkname framework for fair learning, which trains models to minimize a user-specified fairness metric subject to group-wise Pareto efficiency.
\frameworkname treats group-wise empirical losses as multiple objectives, seeking solutions where no group loss can be decreased without increasing at least one other.
Within this efficient set, \frameworkname minimizes the selected fairness metric.
The key idea is to learn the scalarization parameters (group weights) via a bilevel optimization formulation.
The lower level enforces efficiency, whereas the upper level optimizes the fairness criterion, rather than fixing the weights a priori.
This perspective provides a unified basis to compare Pareto-efficient baselines for fair learning (e.g., balanced reweighting and minimax fairness) and enables metric-driven improvements in the learned model.
\myparagraph{\yl{Large-scale bilevel optimization methods}}
\sv{Our second contribution is the development of \yl{two single-loop bilevel algorithms}, \st{\algoname and \stoalgoname}, tailored to our framework \frameworkname in the deterministic and stochastic settings, respectively.}
These algorithms jointly update the model parameters, the dual variable associated with the lower-level optimality conditions, and the group-weight vector through coupled gradient steps. \yl{The proposed updates} require only first-order information on the fairness metric and approximate second-order information on group losses.
\yl{In contrast to} traditional bilevel approaches that rely on costly inner-outer loops or strong smoothness assumptions, \algoname \yl{~and \stoalgoname} operate under mild regularity conditions.
\yl{We establish theoretical convergence guarantees in this general bilevel setting, achieving complexity rates of $O(1/T)$ for both settings. 
Notably, our convergence metric jointly captures stationarity of the implicit upper-level objective and suboptimality of the lower-level solution.
A central challenge in our analysis arises from the absence of \emph{global} smoothness of the lower-level objective with respect to the upper-level variable. In the deterministic setting, we overcome this difficulty by deriving a novel uniform bound on the dual variable generated by the algorithm. In the stochastic setting, such a uniform bound is no longer attainable. To address this issue, we introduce two novel technical arguments: (i) a clipping mechanism that controls the influence of the dual variable in the estimation of the implicit function, and (ii) a new Lyapunov function that extends those previously considered in the literature~\citep{dagreou2022framework} by incorporating an additional term specifically designed to handle the unboundedness of the dual variable.
These results constitute a contribution to the bilevel optimization literature beyond the fairness application of this paper.}

\myparagraph{Open-source Python toolbox}
We release \tbadr\footnote{Repository link : \url{https://github.com/AdaptiveDecisionMakingGroup/badr}}$^{,}$\footnote{Throughout the paper, \tbadr (lowercase) denotes the Python toolbox we release, whereas \frameworkname refers to our overall framework.}, an open-source Python toolbox that implements our general framework.
\tbadr conforms to the \tsklearn interface, which makes the training and analysis of fair models simple and straightforward.
The toolbox is shipped with seven unfairness metrics and supports user-defined custom metrics.
Eleven widely used tabular datasets can be readily fetched, and any \tpandas dataframes can also be used, removing the need for users to write their own custom dataloaders.
The library directly integrates existing learning models such as linear regression, logistic regression and support vector machines.
Finally, we provide abstractions for the oracles, allowing any user to plug in their own optimization algorithm to solve the underlying problem.
We offer an in-depth empirical comparison of several optimization methods (including performance profiles and scalability experiments), highlighting the effectiveness of our proposed algorithm\yl{s} \algoname \yl{~and \stoalgoname}.

\myparagraph{Extensive numerical experiments}
Using the \tbadr toolbox, we conduct extensive numerical experiments on real-world datasets and seven fairness metrics.
We provide numerical evidence that our framework \yl{\frameworkname} improves the targeted fairness metric over other Pareto-efficient baselines, namely uniform and balanced sampling, minimax group fairness, and one-fit group fairness.
We visualize how the unfairness metrics behave along the Pareto front for two and three groups, thereby supporting the need to optimize over the Pareto front.
We also compare \yl{\frameworkname} with these baselines for larger numbers of groups and again show that \yl{it} improves the targeted fairness metric without (1) compromising the group performance Pareto-efficiency nor (2) degrading accuracy (for classification) or RMSE (for regression) on the test set.

}
\subsubsection*{Organization of the paper}\st{
The rest of the paper is organized as follows}. \yl{Section~\ref{sec:preliminaries} motivates multi-objective approaches to fairness in machine learning and introduces our bilevel framework, \frameworkname, for computing optimal Pareto-efficient models with respect to a given fairness metric. Section~\ref{sec:bilevel} presents scalable algorithms for solving the resulting bilevel optimization problem, along with their convergence analysis. In Section~\ref{sec:toolbox}, we introduce \tbadr, our open-source Python toolbox, which enables efficient implementation of our approach and evaluates its numerical performance on several standard learning tasks. Finally, Section~\ref{sec:experiments} is devoted to extensive empirical studies demonstrating the superiority of our method over state-of-the-art approaches to Pareto-efficient fair learning.
}

\section{Fairness through a Pareto lens}\label{sec:preliminaries}

\yl{We consider a general prediction problem in which the data consists of triplets $(x_i, y_i, a_i)$. 
Specifically, $(x_i, y_i) \in \inputspace \times \outputspace$ represents the input-output pair for individual $i$, and $a_i \in \attributespace = \{1, \ldots, S\}$ indicates their sensitive attribute (e.g. gender, race, or similar characteristics).
Traditionally, predictors $f(\model, \cdot): \inputspace \to \outputspace$ are estimated through the empirical risk minimization (ERM) principle. Given training data $(x_i, y_i, a_i)_{i\in \{1, \ldots, n\}}$, this entails solving
\begin{equation}\label{eq:erm}
    \min_{\model \in \mathbb{R}^d} \frac{1}{n} \sum_{i=1}^n \left[\ell(f(\model, x_i)), y_i \right], 
\end{equation}
where $\model \in \RR^d$ parametrizes the predictor.  
Notably, ERM disregards individuals' sensitive attribute, potentially resulting in biases across the communities involved~\citep{blodgett2016demographic}.
In this work we seek to develop a novel learning method that reduces the disparity of performances across communities. 
Our approach aligns with the framework of in-process fairness methods~\citep{Caton2024}. 
In this section, we present an adaptive reweighting scheme to enhance fairness while guaranteeing efficiency across groups performance.
}
\subsection{Inefficiency of fairness by regularization approaches}
\begin{figure}[t]
    \centering
    \includegraphics[width=\textwidth]{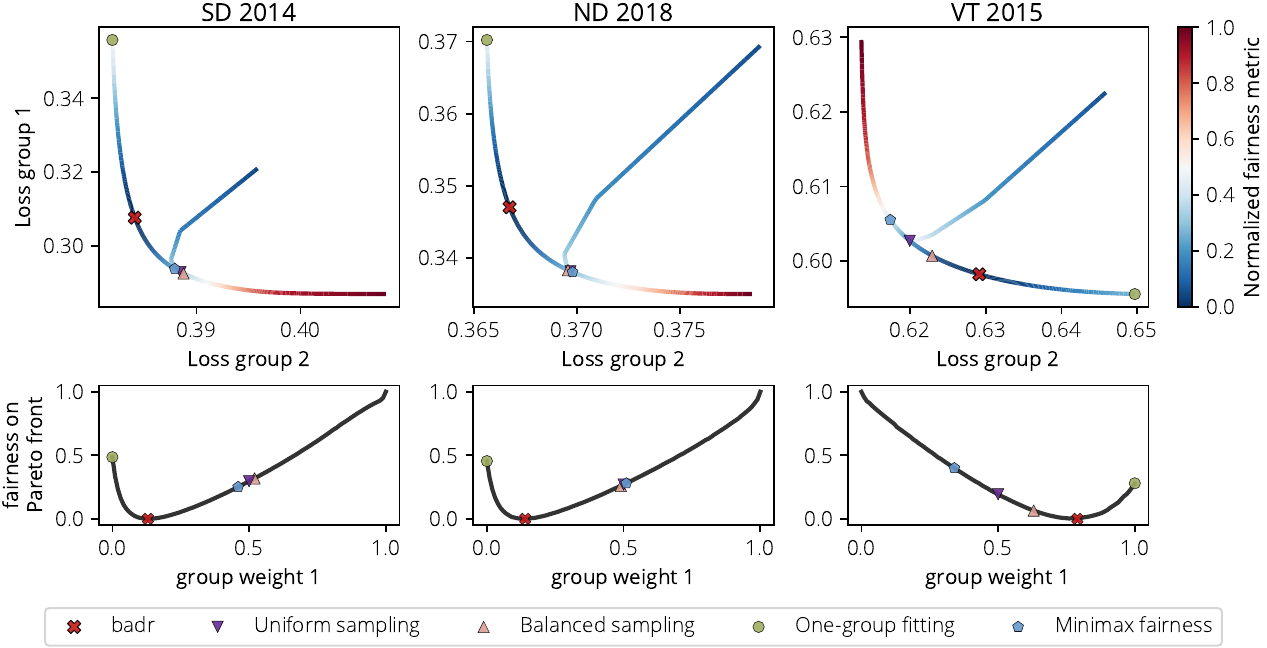}  
    \caption{{\small \yl{Illustration of the impact of fairness regularization on Pareto efficiency for a logistic regression task evaluated on three subsamples of the ACSEmployment dataset \citep{dataset_folktables}.
    The first row reports the group-wise losses for the two sensitive groups (male vs. female). The two curves correspond to the Pareto front and the regularization path induced by the fairness-by-penalization approach, respectively.
    The second row reports the values of the considered fairness metric (individual fairness \citep{metric_kearns}) on the Pareto front, together with the performance of several fairness Pareto-efficient methods compared against \frameworkname.}
    }}
    \label{fig:penalization}
\end{figure}
\yl{
A natural starting point for quantifying performance disparities across groups is to explicitly examine their respective losses.
Specifically, for each sensitive attribute $a \in \attributespace$, we define the \emph{group loss} $F_a : \model \mapsto F_{a}(w) = n_a^{-1} \sum_{i : a_i = a} \ell(f(\model, x_i), y_i)$, where $n_a = \#\{i : a_i = a\}$ denotes the number of data points corresponding to the sensitive attribute $a$.
Group losses allow to extend the original empirical risk minimization problem~\eqref{eq:erm} into a multi-objective formulation:
\begin{equation}\label{eq:MOproblem}
    \min_{\model \in \mathbb{R}^d} \left[F_1(\model), \ldots,  F_S(\model)\right].
\end{equation}
Considering fairness through group losses underpins many fairness-oriented learning methods. 
A prominent strategy, often termed minmax fairness, reallocates sampling weight toward the worst-performing groups, thereby reducing \eqref{eq:MOproblem} to a single-objective problem \st{(see \cref{subsec:minmax})}.
Nonetheless, the full multi-objective formulation in~\eqref{eq:MOproblem} admits a broader spectrum of solutions, offering greater flexibility in pursuing fairness. 
Such solutions are typically characterized by the concept of Pareto efficiency:
\begin{definition}[Pareto Efficiency]
A model $\bar{\model}$ is called \emph{Pareto efficient} for the group losses $(F_{a})_{a \in \attributespace}$, if for all $w \in \mathbb{R}^d$, one of the two following assertions is satisfied

\[
\begin{aligned}
    &(i)  & F_a(w) \geq F_a(\bar w), \quad \text{for all } a \in \attributespace, \\
    &(ii) & F_a(w) > F_a(\bar w), \quad \text{for some } a \in \attributespace. \\
\end{aligned}
\]
The set of Pareto efficient point for~\eqref{eq:MOproblem}, denoted subsequently $\paretofront$, is called the \emph{Pareto front}.
\end{definition}
Intuitively, Pareto-efficient solutions characterize models for which performance cannot be improved for any group without adversely affecting at least one other. 
While this requirement seems natural, it is not necessarily achieved by current in-processing fairness methods. 
This can be observed for instance on fairness by constraint methods, which typically involve augmenting the empirical risk minimization (ERM) problem~\eqref{eq:erm} with a fairness constraint,

\begin{equation}\label{eq:fairnesss_constrained_pb}    
\begin{aligned}
    \min_{\model \in \inputspace} \quad & \frac{1}{n} \sum_{i=1}^n \ell(f(\model, x_i), y_i) \; \text{s.t.} \; \fairnessmetric(\model) \leq \varepsilon,\\
\end{aligned}
\end{equation}
or its relaxation through penalization~\citep{preprocess_fairglm}, 

\begin{equation}\label{eq:fairnesss_cpenalized_pb}   
    \min_{\model \in \inputspace}  \quad \frac{1}{n} \sum_{i=1}^n \ell(f(\model, x_i), y_i)  + \frac{1}{\nu}\; \fairnessmetric(w).
\end{equation}
Here, $\varepsilon > 0$ (resp. $\nu > 0$) is a user-defined tolerance level that governs the permissible degree of unfairness, and $\fairnessmetric:\inputspace \to \RR$ denotes the fairness metric used to quantify model bias - we review a number of standard fairness metrics in Appendix~\ref{app:fairnessmetrics}.

In Figure \ref{fig:penalization} we illustrate the effect of fairness regularization on Pareto efficiency across three classification tasks. 
%
%
We train a logistic regression model on \st{three distinct subsamples of the ACSEmployment dataset \citep{dataset_folktables}.}
In each case, gender (male vs female) is the sensitive attribute. 
The model is augmented with an individual-fairness penalty~\citep{metric_kearns}, and we vary the penalty weight $\nu$ over a logarithmic grid from $10^{-6}$ to $10^{-0.5}$.
\st{The top row plots the Pareto front of group losses and the regularization path starting from uniform weights.
Both are colored by the individual fairness metric, with the color scale shown by the colorbar.}
%
\st{In the bottom row, we show how the fairness metric changes with the group weights.
This view makes changes in the metric more apparent than the color map.}
Across all three datasets, the resulting trajectories significantly depart from the Pareto frontier displayed by the subgroup losses (in blue).
This demonstrates that naive penalization produces solutions that are dominated-and therefore inefficient from a Pareto perspective.
}
\subsection{Scalarization and the weighted-sum method}
%
\yl{
Problem~\eqref{eq:MOproblem} may be addressed by a broad literature of optimization approaches aiming to recover its Pareto front.
These include scalarization methods, a priori methods, or no-preferences methods - for general references, see~\citep{branke2008multiobjective,marler2004survey}.
In this work, we will build upon a standard scalarization strategy in the \st{multiobjective} literature, the weighted-sum method \citep{gass1955computational,geoffrion1968proper}, which given a distribution of weights $\weight_a \in \simplex{S}$, turns~\eqref{eq:MOproblem} into the single optimization problem:

\begin{equation}\label{eq:scalazized_problem}
    \min_{\model \in \mathbb{R}^d} \sum_{a \in \attributespace} \weight_a F_a(w)
\end{equation}
Under mild conditions, solutions of~\eqref{eq:scalazized_problem} are guaranteed to be Pareto efficient. 
Their closure also spans the entire Pareto front when the Group losses are strongly convex. 
\begin{proposition}[See~\citep{geoffrion1968proper}]
Let $\lambda \in \simplex{S}$ be fixed. Any solution $w_\lambda$ of \eqref{eq:scalazized_problem} is Pareto-efficient for the multi-objective problem~\eqref{eq:MOproblem}. 
Conversely, if the group losses $F_{a}$ are convex and at least one of them is strictly convex, then any Pareto efficient point for~\eqref{eq:MOproblem} belongs to the closure of $\{w_{\lambda}, \lambda \in \simplex{S}\}$.
\end{proposition}
When the learning weights $\weight_a$ correspond to the proportion $n_a/n$ of each group in the training sample, one recovers the standard ERM problem~\eqref{eq:erm}, which therefore outputs by the above Proposition a Pareto-efficient model. 
Alternatively, the fairness literature has considered a broad family of data reweighting strategies to alleviate the unfairness of ERM's output. 
%
%
Yet these approaches are mostly restricted to binary classification tasks -- e.g.~\citep{calders2009building,krasanakis2018adaptive} -- and further require weights to jointly depend on sensitive attribute $a$ and output label $y$. 
This is for instance the case for~\citep{calders2009building} which adopts an importance sampling approach to achieve independence on training data between sensitive attribute and output variable. 
As a consequence, group loss performance may no longer exhibit Pareto efficiency. 
}
%
\subsection{Minmax fairness and fairness risk measures as robust scalarizations}
\label{subsec:minmax}
%

\yl{
From a group losses perspective, min-max fairness~\citep{martinez2020minimax,hashimoto2018fairness,diana2021minimax,abernethy2020active} constitutes a popular approach to fairness that focuses on improving performance over worst-performing groups

\begin{equation}\label{eq:def_minmax_fairness_pb}
    \min_{\model} \max_{a \in \attributespace} F_a(\model).    
\end{equation}
Solutions of~\eqref{eq:def_minmax_fairness_pb} offer robust guarantees on the output model~$\model$ by minimizing the worst-case group error, often resulting in a reduced performance gap across sub-groups.
This highlights their advantage over equal error rate approaches~\citep{diana2021minimax}, which may enforce parity by uniformly degrading performance-even at the expense of disadvantaged groups.

Finding such solutions also present important computational advantages over other in-processing methods, benefiting from the extended literature on stochastic min-max optimization~\citep{beznosikov2023smooth}. 
In particular, optimization algorithms for~\eqref{eq:def_minmax_fairness_pb} are frequently supported with theoretical guarantees including the preservation of convexity for linear tasks (logistic or linear regression) and convergence rates toward optimal or critical point in the non-convex setting. 
These perks have spurred the interest of min-max fairness approaches on a broad span of learning tasks, notably in federated learning~\citep{mohri2019agnostic}. 

A recurrent reproach to formulation~\eqref{eq:def_minmax_fairness_pb} is its potential over-emphasis on the performance of a single-group, which may reveal inadequate in situations where many groups are involved. 
To this end, extensions of~\eqref{eq:def_minmax_fairness_pb} have been considered to provide robustness on distributions over groups, leading to the min-max formulation

\begin{equation}\label{eq:def_dro_fairness}
    \min_{\model} \max_{Q \in \cD} \mathbb{E}_{\boldsymbol{a}\sim Q}[F_{\boldsymbol{a}}(\model)].    
\end{equation}
where $\cD$ denotes a set of probability distributions supported on the simplex $\simplex{S}$.
Benefits of these extensions have been studied in a number of papers and for various formulations of the ambiguity set~\citep{pillutla2024federated}, bridging connections with the theory of risk measures~\citep{williamson2019fairness,tsang2022unified}.
It turns out problem~\eqref{eq:def_minmax_fairness_pb} and its extension~\eqref{eq:def_dro_fairness} both output Pareto-efficient models when group losses are convex:
\begin{proposition}\label{prop:dro_pareto}
    Assume the group losses $(F_{a})_{a \in \attributespace}$ are convex. Then any solution of \eqref{eq:def_minmax_fairness_pb} or \eqref{eq:def_dro_fairness} is Pareto-efficient. 
\end{proposition}
Proposition~\ref{prop:dro_pareto} constitutes the basis of~\citep{martinez2020minimax} to motivate for min-max fairness, although the formulation presented there remains expressed as $\min_{w \in \paretofront} \max_{a \in \attributespace} F_{a}(w)$ \st{with an explicit restriction to $\paretofront$}.
Subsequent works~\citep{abernethy2020active,diana2021minimax} have relaxed the Pareto front constraint, which directly follows from Proposition~\ref{prop:dro_pareto}. 
%
A proof of Proposition~\eqref{prop:dro_pareto} is provided in \citep[Proposition 1]{sagawa2019distributionally} for the analogous setting in which the ambiguity set $\cD$ consists of distributions supported on the training set, rather than on the set of sensitive attributes. 
%
}

%
%
\subsection{\frameworkname : an \emph{optimal} Pareto-efficient approach to group fairness}
%

%
\yl{
In this paper, we propose to learn optimal Pareto efficient models for~\eqref{eq:MOproblem} given an arbitrary measure of unfairness $\fairnessmetric : \RR^d \to \mathbb{R}$. 
Leveraging the implicit characterization in~\eqref{eq:scalazized_problem}, we cast this task as a bilevel program, which we term \frameworkname (Bilevel Adaptive Rescalarization), and formulate as follows: 
\begin{equation}\label{eq:bilevel_formulation}
\begin{aligned}
    &\min_{\lambda~\in \simplex{S}} \quad && \fairnessmetric (w_\lambda) \\
    &\text{s.t.} \quad && w_\lambda \in \arg\min_{w \in \RR^d} \sum_{a \in \attributespace} \lambda_a F_a(w)
\end{aligned}
\end{equation}
When the group losses $F_a$ are strongly convex, solutions of~\eqref{eq:bilevel_formulation} are guaranteed to be Pareto-efficient. 
Furthermore, provided the unfairness metric $\fairnessmetric$ is differentiable - or admits a tractable smooth approximation - problem~\eqref{eq:bilevel_formulation} is amenable to standard bilevel programming methods for large-scale learning~\citep{pedregosa2016hyperparameter,dagreou2022framework}. 
In particular, for any initial point $\lambda$ in the simplex, the unfairness of the associated model $w_\lambda$ can be decreased until a critical point of the implicit loss $\lambda \mapsto \fairnessmetric(w_\lambda)$ is reached.

Several bilevel approaches to in-processing fairness have been explored in the literature. 
\citep{roh2020fairbatch} proposes to learn a sampling distribution depending jointly on sensitive attributes and labels. 
In particular, the inner level objective may not correspond to a weighted sum of group losses, which can compromise Pareto efficiency of the output model. 
\citep{martinez2020minimax,shekhar2021adaptive} study the particular instance of $\eqref{eq:bilevel_formulation}$ where the unfairness metric $\fairnessmetric$ corresponds to maximum group loss. 
\citep{ozdayi2021fair} studies a similar problem to~\eqref{eq:bilevel_formulation}, where the upper level additionally includes the population loss $w \mapsto \sum_{i\in \cA} \frac{n_i}{n} F_a(w)$ as a penalty term.  
While several fairness metrics are considered therein, no connection to Pareto efficiency is established. 
On a different note, multi-objective approaches for group fairness have also been considered in~\citep{liu2022accuracy}, where \st{the} group losses and fairness metrics are taken as competing objectives. 

Our general formulation~\eqref{eq:bilevel_formulation} offers two key advantages over existing approaches in the literature.
To begin with, it is the first formulation that can accommodate a broad class of fairness metrics while simultaneously ensuring Pareto efficiency across groups.
Second, as shown in the following section, it enables efficient optimization with provable guarantees.
}

\section{Large scale methods}\label{sec:bilevel}
\yl{In this section, we develop a single-loop method we coin \algoname, for Bilevel ADaptive Rescalarization, to solve the bilevel program~\eqref{eq:bilevel_formulation}. 
We present in Section~\ref{sec:algorithm} the algorithmic details of \algoname.
We further establish in Section~\ref{sec:analysis} its convergence properties under standard assumptions in the bilevel optimization literature.
This method complement the algorithmic arsenal we developed for small to medium-scale problems in \cref{sec:toolbox}. 

\subsection{Algorithms}\label{sec:algorithm}

We present in Algorithm~\ref{alg:badr} a simple method, we name~\algoname to solve the bilevel problem~\eqref{eq:bilevel_formulation} for large-scale instances.\
\algoname iteratively updates three variables : the learning model $\lowervar$, the dual variable $\dualvar$ associated to the optimality condition of the lower-level problem in~\eqref{eq:bilevel_formulation}, and the fairness reweighting vector $\uppervar$.
Specifically, at each iteration $t$, it performs a gradient descent step on the reweighted objective $\lowervar \mapsto \sum_{a \in \attributespace} \lambda_{t, a} F_a(\lowervar)$ (line 3) and \st{a gradient step} on the quadratic loss $\dualvar \mapsto \frac{1}{2} \dualvar\top Q_t \dualvar_t + \nabla \fairnessmetric(w_t)^\top \dualvar$, where $Q_t$ denotes the weighted hessian $Q_t \defineq \sum_{a \in \attributespace} \lambda_{t, a} \nabla^2 F_a(\lowervar_t)$ (line 4).
Finally, the fairness weights $\uppervar$ are updated via an approximate projected gradient step on the objective $\lambda \mapsto \fairnessmetric(\lowersol(\lambda))$ (line 5).
{\small
\begin{algorithm}[t]
\caption{\algoname}\label{alg:badr}
\begin{algorithmic}[1]
    \Require{Starting points $\lowervar_0,\dualvar_0,\uppervar_0$; step sizes $\tau,\rho,\gamma$}
    \For{$t = 0$ to $T-1$}    
    \State $\lowervar_{t+1} \gets \lowervar_t - \tau\, \nabla \boldsymbol{F}(\lowervar_t)^\top \lambda_t$ \hfill \Comment{\small $\nabla \boldsymbol{F}(\lowervar_t) \in \RR^{\attributenumber \times d} \text{ with } a^{th} \text{ row equal to } \nabla F_a(\lowervar_t)$}
    \State $\dualvar_{t+1} \gets \dualvar_t - \rho
                                                    ( \nabla \fairnessmetric(\lowervar_t)
                                                        + \sum_{a \in \attributespace} \lambda_a \nabla^2 F_a(\lowervar)\, \dualvar_t )$    
                                                            \State $\uppervar_{t+1} \gets  \text{Proj}_{\simplex{\attributenumber}} 
                                                                                                (
                                                                                                    \uppervar_t - \gamma  \nabla \boldsymbol{F}(\lowervar_t)\, \dualvar_t 
                                                                                                )$  
    \EndFor
    \State \textbf{return} $\lowervar_T,\ \uppervar_T$
\end{algorithmic}
\end{algorithm}
}
A notable advantage of this method is its simplicity, as it only requires (approximate) second-order information on the group losses $F_a$ and first-order information on the fairness metric $\fairnessmetric$. 
Furthermore, its single loop nature makes it particularly well-suited for large-scale problems, as it avoids the costly inner-loop optimization of the lower-level problem in~\eqref{eq:bilevel_formulation}.

Still, in typical learning scenarios, one can hardly expect to access exact gradients for the group losses $F_a$ or the fairness metric $\fairnessmetric$. 
We therefore extend in \cref{alg:stochastic_badr} our method to the case where only stochastic estimates of these gradients are available. 
Specifically, we assume access to stochastic estimates $\stonabla \fairnessmetric(\lowervar)$,  $\stonabla F_a(\lowervar)$, and $\stonabla \fairnessmetric(\lowervar)$ of $\nabla \fairnessmetric(\lowervar)$, $\nabla F_a(\lowervar)$ and $\nabla^2 F_a(\lowervar)$ respectively, which may be computed through mini-batch samples.
A notable difference with~\algoname is the use of a clipping operator on the gradient estimate for the upper variable $\uppervar$ (line 5 of Algorithm~\ref{alg:stochastic_badr}) that will be instrumental in the derivation of our convergence results in Section~\ref{sec:analysis}.

\begin{algorithm}[t]
\caption{\stoalgoname}\label{alg:stochastic_badr}
\begin{algorithmic}[1]
    \Require{Starting points $\lowervar_0,\dualvar_0,\uppervar_0$; step sizes $\tau,\rho,\gamma$; clipping threshold \st{$C_\gamma$}}
    \For{$t = 0$ to $T-1$}    
    \State $\lowervar_{t+1} \gets \lowervar_t - \tau\, \stonabla \boldsymbol{F}(\lowervar_t)^\top \lambda_t$ \hfill \Comment{\small $\stonabla \boldsymbol{F}(\lowervar_t) \in \RR^{\attributenumber \times d} \text{ with } a^{th} \text{ row equal to } \stonabla F_a(\lowervar_t)$}
    \State $\dualvar_{t+1} \gets \dualvar_t - \rho
                                                    ( \stonabla \fairnessmetric(\lowervar_t)
                                                        + \sum_{a \in \attributespace} \lambda_a \stonabla^2 F_a(\lowervar)\, \dualvar_t )$    
                                                            \State $\uppervar_{t+1} \gets  \text{Proj}_{\simplex{\attributenumber}} 
                                                                                                (
                                                                                                    \uppervar_t - \gamma \clip\left(\stonabla \boldsymbol{F}(\lowervar_t)\, \dualvar_t\right) 
                                                                                                )$  
    \EndFor
    \State \textbf{return} $\lowervar_T,\ \uppervar_T$
\end{algorithmic}
\end{algorithm}

\subsection{Convergence analysis}\label{sec:analysis}
To analyze the bilevel formulation~\eqref{eq:bilevel_formulation}, we rewrite it in the form of a general bilevel optimization problem :

\begin{equation}
    \label{eq:general_bilevel}
    \begin{aligned}
        &\min_{\lambda \in \cC} \quad && \upperobj(w(\lambda), \lambda) \\
        &\text{s.t.} \quad && w_\lambda \in \arg\min_{w \in \RR^d} \lowerobj(w, \lambda)
    \end{aligned}    
\end{equation}
where $\cC
$ 
denotes a convex compact subset, $\upperobj : \RR^d \times \cC \to \RR$ and $\lowerobj : \RR^d \times \cC \to \RR$ are continuously differentiable functions.
Applied to Problem~\eqref{eq:general_bilevel}, our algorithm~\algoname reads :
\begin{equation}
    \label{eq:badr_updates}
    \begin{aligned}
        \lowervar_{t+1} &\leftarrow \lowervar_t - \tau\, \nabla_{\lowervar}\, \lowerobj(\lowervar_t,\uppervar_t) \\
        \dualvar_{t+1} &\leftarrow \dualvar_t - \rho\left(\nabla_{\lowervar, \lowervar} \lowerobj(\lowervar_t,\uppervar_t)\, \dualvar_t + \nabla_{\lowervar} \upperobj(\lowervar_t,\uppervar_t)\right) \\
        \uppervar_{t+1} &\leftarrow \text{Proj}_{\simplex{\attributenumber}} \left( \uppervar_t - \gamma \left( \nabla^{2}_{\lowervar, \uppervar} \lowerobj(\lowervar_t,\uppervar_t)^\top\, \dualvar_t + \nabla_{\uppervar} \upperobj(\lowervar_t,\uppervar_t) \right) \right)
    \end{aligned}
\end{equation}
As such it can be seen as a variant of the SOBA method~\citep{dagreou2022framework}, adapting its parameterization to the structure of~\eqref{eq:bilevel_formulation}.
Indeed, \st{in addition to} the novel constraint holding on the upper variable $\uppervar$, the main distinction with existing works is the absence of global smoothness of the lower-level objective $\lowerobj$ with respect to $\uppervar$.
In fact, in the specific setting of Problem~\eqref{eq:bilevel_formulation}, $\lowerobj$ admits the form $\lowerobj(\lowervar, \uppervar) = \sum_{s \in \attributespace} \uppervar_s F_s(\lowervar)$. 
This leads to crossed derivatives of the form: 
$\nabla_{w, \uppervar}^2 \lowerobj(\lowervar, \uppervar) = 
    \begin{bmatrix}
        \nabla F_1(\lowervar) & \nabla F_2(\lowervar) & \cdots & \nabla F_S(\lowervar)
    \end{bmatrix}.$
%
Hence, uniform smoothness cannot hold, since the gradients of group losses $\nabla F_s$ are unbounded over $\RR^d$ due to the strong convexity of the functions $F_s$.
This in turn invalidates the previous convergence analysis of SOBA~\citep{dagreou2022framework}, in several places including the Lipschitz structure of the inner solution $\lowersol : \uppervar \mapsto \argmin_{\lowervar} \lowerobj(\lowervar, \uppervar)$, the smoothness of the implicit function $\implicitobj : \uppervar \mapsto \upperobj(\lowersol(\uppervar), \uppervar)$ or the control of the error on the gradient of its gradient.
In contrast our approach makes use of weaker assumptions on the lower-level objective $\lowerobj$ that remain midly acceptable in practice, and that we clarify below.
%

\begin{assumption}
    \label{assumption:upperobj}    
    The upper objective $\upperobj$ satisfies the following conditions :
    \begin{enumerate}        
        \item $\upperobj$ is Lipschitz continuous on $\RR^d \times \cC$ with constant $L_{\upperobj, 0} > 0$, i.e. for all $\lowervar_1, \lowervar_2 \in \RR^d$ and $\uppervar_1, \uppervar_2 \in \cC$:
        \[
            |\upperobj(\lowervar_1, \uppervar_1) - \upperobj(\lowervar_2, \uppervar_2)| \leq L_{\upperobj, 0} \|(\lowervar_1, \uppervar_1) - (\lowervar_2, \uppervar_2)\|
        \]        
        \item $\nabla \upperobj$ is Lipschitz continuous on $\RR^d \times \cC$ with Lipschitz constant $L_{\upperobj, 1} > 0$, i.e. for all $\lowervar_1, \lowervar_2 \in \RR^d$ and $\uppervar_1, \uppervar_2 \in \cC$:
        \[
            \|\nabla_{\lowervar} \upperobj(\lowervar_1, \uppervar_1) - \nabla_{\lowervar} \upperobj(\lowervar_2, \uppervar_2)\| \leq L_{\upperobj, 1} \left(\|\lowervar_1- \lowervar_2\| +  \| \uppervar_1 - \uppervar_2 \|\right)
        \]        
    \end{enumerate}
\end{assumption}

Assumption~\ref{assumption:upperobj} is standard in the bilevel optimization literature~\citep{ghadimi2018approximation,arbel2021amortized,dagreou2022framework} and essential to ensure global smoothness of the implicit function $\implicitobj : \uppervar \mapsto \upperobj(\lowersol(\uppervar), \uppervar)$.
We complement it with the following assumption on the lower-level objective.
\begin{assumption}
    \label{assumption:lowerobj}
    The lower objective $\lowerobj$ is $\cC^2$-differentiable and satisfies the following conditions :
    \begin{enumerate}
        \item $\lowerobj$ is strongly convex with respect to $\lowervar$ on $\RR^d \times \cC$ with strong convexity constant $\mu_{\lowerobj} > 0$. That is, for all $\lowervar_1, \lowervar_2 \in \RR^d$ and $\uppervar \in \cC$:
        \[
            \lowerobj(\lowervar_2, \uppervar) \geq \lowerobj(\lowervar_1, \uppervar) + \langle \nabla_{\lowervar} \lowerobj(\lowervar_1, \uppervar), \lowervar_2 - \lowervar_1 \rangle + \frac{\mu_{\lowerobj}}{2} \|\lowervar_2 - \lowervar_1\|^2.
        \]
        \item $\nabla_{\lowervar} \lowerobj$ is Lipschitz with respect to $\lowervar$ on $\RR^d$ with Lipschitz constant $L_{\lowerobj, 1}^{\lowervar, \lowervar} \geq 0$, uniformly over $\cC$. That is, for all $\lowervar_1, \lowervar_2 \in \RR^d$ and $\uppervar \in \cC$:
        \[
            \|\nabla_{\lowervar} \lowerobj(\lowervar_1, \uppervar) - \nabla_{\lowervar} \lowerobj(\lowervar_2, \uppervar)\| \leq L_{\lowerobj, 1}^{\lowervar, \lowervar} \|\lowervar_1 - \lowervar_2\|.  
        \]        
        \item $\nabla^2_{\lowervar,\lowervar} \lowerobj$ is Lipschitz Hessian with respect to $\lowervar$ on $\RR^d \times \cC$ with Lipschitz constant $L_{\lowerobj, 2}^{\lowervar, \lowervar} \geq 0$. That is, for all $\lowervar_1, \lowervar_2 \in \RR^d$ and $\uppervar \in \cC$ : 
        \[
            \|\nabla^2_{\lowervar,\lowervar} \lowerobj(\lowervar_1, \uppervar) - \nabla^2_{\lowervar,\lowervar} \lowerobj(\lowervar_2, \uppervar)\| \leq L_{\lowerobj, 2}^{\lowervar, \lowervar} (\|\uppervar_1 - \uppervar_2\| + \|\lowervar_1 - \lowervar_2\|). 
        \]
        \item $\nabla_{\lowervar, \uppervar}^2 \lowerobj$ is uniformly upper bounded by a constant $M_\cC > 0$ over the solution set~\\ 
        $\{(\lowersol(\uppervar_1), \uppervar_2), (\uppervar_1,\uppervar_2)  \in \cC^2\}$, i.e. for all $\uppervar_1, \uppervar_2 \in \cC$,         
        \[
            \|\nabla_{\lowervar, \uppervar}^2 \lowerobj (\lowersol(\uppervar_1), \uppervar_2)\| \leq M_\cC.
        \]  
        \item $\nabla_{\lowervar, \uppervar}^2 \lowerobj$ is Lipschitz with constant $L_{\lowerobj, 2}^{\lowervar, \uppervar} > 0$. That is for all $\lowervar_1, \lowervar_2 \in \RR^d$ and $\uppervar_1, \uppervar_2 \in \cC$:
        \[
            \|\nabla_{\lowervar, \uppervar}^2 \lowerobj (\lowervar_1, \uppervar_1) - \nabla_{\lowervar, \uppervar}^2 \lowerobj (\lowervar_2, \uppervar_2)\| \leq L_{\lowerobj, 2}^{\lowervar, \uppervar} \left(\|\lowervar_1 - \lowervar_2\| + \|\uppervar_1 - \uppervar_2\|\right).
        \]
    \end{enumerate}
\end{assumption}
%
%
First three conditions of Assumption~\ref{assumption:lowerobj} are also common to ensure the well-posedness of the solution mapping $\uppervar \mapsto \lowersol(\uppervar)$ and the smoothness of the implicit function $\implicitobj$. 
A notable distinction with existing works~\citep{arbel2021amortized,lu2023slm,dagreou2022framework} is the absence of global smoothness of the lower objective with respect to $\uppervar$.
%
Instead, we relax this assumption and only require an upperbound for $\nabla_{w, \uppervar}^2 \lowerobj$ over the solution set $\{(\lowersol(\uppervar), \uppervar), \uppervar \in \cC\}$.
Finally, we further require this crossed derivative to be Lipschitz continuous, which would directly follow from a smoothness assumptions on the gradients $\nabla F_s$ in \eqref{eq:bilevel_formulation}.
%
We note that these assumptions are typically satisfied in learning problems with strongly convex loss functions and standard regularization schemes, such as ridge regression or logistic regression, provided the data are bounded\footnote{Our extension to the stochastic setting in Theorem~\ref{thm:convergence_stochastic} relaxes this requirement by allowing data with only bounded second-order moments.}. 

Our first convergence result is summarized in the following theorem.
The convergence metric we consider is a positive combination of the squared norm of the generalized gradient 
\[
    \pgstep_{\uppervar, t} \defineq \frac{1}{\gamma} \left( \uppervar_t - \text{Proj}_{\simplex{\attributenumber}} \left( \uppervar_t - \gamma \nabla \varphi(\uppervar_t) \right) \right)  
\]
and the squared distance of the lower variable to optimality $\|\lowervar_t - \lowersol(\uppervar_t)\|^2$. 
Finally, we denote by $\varepsilon_{\lowervar, 0} \defineq \|\lowervar_0 - \lowersol(\uppervar_0)\|$ and $\varepsilon_{\dualvar, 0} \defineq \|\dualvar_0 - \nabla \implicitobj(\uppervar_0)\|$ the initialization errors on the lower and dual variables respectively.
\begin{theorem}
    \label{thm:convergence_deterministic}     
    Consider running the iteration~\eqref{eq:badr_updates} for $T$ steps with step sizes $\tau, \rho, \gamma$ satisfying $\tau=\rho=1/L_{\lowerobj, 1}^{\lowervar, \lowervar}$ and $\gamma = \min(\kappa_{\lowerobj, \lowervar}^{-1}, \bar \gamma)$ where $\kappa_{\lowerobj, \lowervar} \defineq L_{\lowerobj, 1}^{\lowervar, \lowervar}/\mu_{\lowerobj}$ and $\bar \gamma$ is a constant made explicit in~\eqref{eq:def_bar_gamma}. 
    Then, under Assumptions~\ref{assumption:upperobj} and~\ref{assumption:lowerobj}, the sequence $(\lowervar_t, \uppervar_t)_{t \geq 0}$ generated by~\eqref{eq:badr_updates} satisfies for all $T \geq 1$,
    {\small
    \[
        \frac{1}{T} \sum_{t=0}^{T-1} \|\pgstep_{\uppervar, t}\|^2 + \bar \alpha \; \|\lowersol(\uppervar_t) - \lowervar_t\|^2 \leq \frac{8}{\gamma T} \left( \implicitobj(\uppervar_0) - \inf_{\uppervar \in \cC} \implicitobj(\uppervar) \right) + \frac{4 \kappa_{\lowerobj, \lowervar} }{T} \left(\bar \alpha \;\varepsilon_{\lowervar, 0}^2 +  \bar \beta  \varepsilon_{\dualvar, 0}^2 \right)
    \]
    }
    where 
    {\small
    \begin{align*}
        \bar \alpha & = 12 \left(L_{\upperobj, 1}^{\lowervar, \uppervar} + L_{\lowerobj, 2}^{\lowervar, \uppervar} \left(2 \kappa_{\lowerobj, \lowervar} \frac{L_{\upperobj, 0}}{\mu_{\lowerobj}} + \varepsilon_{\dualvar, 0}\right)\right)^2 
                    + 2568 \kappa_{\lowerobj, \lowervar}^2 M_\cC^2 \Bigl( L_{\upperobj, 0} \frac{L_{\lowerobj, 2}^{\lowervar, \lowervar}}{\mu_{\lowerobj}^2} + \frac{L_{\upperobj, 1}}{\mu_{\lowerobj}} \Bigr)^2 \\
        \bar \beta & = 6 M_\cC^2.   
    \end{align*}
    }
\end{theorem}
We defer the proof of this theorem to Section~\ref{sec:proof_convergence} of the Appendix.
\begin{remark}
Our convergence rate is of order $\cO(1/T)$ on squared gradient norms, which matches the standard rate for non-convex smooth optimization~\citep{beck2017first} as well as the rate achieved for bilevel algorithms~\citep{dagreou2022framework}.  
Furthermore, our convergence metric incorporates optimality at the lower-level, \st{an improvement over} the analysis derived in~\citep{dagreou2022framework}.
\end{remark}

\begin{remark}
An important component of our analysis is the control of the dual variable $\dualvar_t$ along the iterations. 
Specifically, we manage to upper bound $\|\dualvar_t\|$ uniformly over time, using the strong convexity of the lower-level problem.
This in turn enables us to control the error $\|\nabla \implicitobj(\uppervar_t) - (\nabla^{2}_{\lowervar, \uppervar} \lowerobj(\lowervar_t,\uppervar_t)^\top\, \dualvar_t + \nabla_{\uppervar} \upperobj(\lowervar_t,\uppervar_t))\|$ on the gradient of the implicit function.
\end{remark}
We complement our analysis with convergence guarantees for~\stoalgoname, our stochastic variant of~\algoname laid down in Algorithm~\ref{alg:stochastic_badr}.\
\stoalgoname allows for stochastic estimates of the oracles involved in~\eqref{eq:bilevel_formulation}.
Unlike in the deterministic setting, we cannot uniformly control the dual variable $\dualvar_t$ along the iterations as in Lemma~\ref{lem:dual_bounded}; instead, we can only bound its second-order moment (see Lemma~\ref{lem:dual_bounded_stochastic}). 
Although uniform boundedness of the dual variable plays a key role in controlling the stochastic error of the implicit gradient estimate $\stonabla^{2}_{\lowervar, \uppervar} \lowerobj(\lowervar_t,\uppervar_t)^\top\, \dualvar_t + \stonabla_{\uppervar} \upperobj(\lowervar_t,\uppervar_t)$, we show that this issue can be effectively handled by introducing an additional clipping mechanism.
Following our formalism, we consider now iterations of the form : 
\begin{equation}
    \label{eq:badr_stochastic_updates}
    \begin{aligned}
        \lowervar_{t+1} &\leftarrow \lowervar_t - \tau\, \stonabla_{\lowervar}\, \lowerobj(\lowervar_t,\uppervar_t) \\
        \dualvar_{t+1} &\leftarrow \dualvar_t - \rho\left(\stonabla_{\lowervar, \lowervar} \lowerobj(\lowervar_t,\uppervar_t)\, \dualvar_t + \stonabla_{\lowervar} \upperobj(\lowervar_t,\uppervar_t)\right) \\
        \uppervar_{t+1} &\leftarrow \text{Proj}_{\simplex{\attributenumber}} \left( \uppervar_t - \gamma \clip \left( \stonabla^{2}_{\lowervar, \uppervar} \lowerobj(\lowervar_t,\uppervar_t)^\top\, \dualvar_t + \stonabla_{\uppervar} \upperobj(\lowervar_t,\uppervar_t) \right) \right)
    \end{aligned}
\end{equation}
where $\stonabla$ denotes a stochastic estimate of the corresponding gradient, and $\clip : g \mapsto \min(1, C_\gamma/\|g\|)$ denotes a clipping operator and $C_\gamma$ an additional hyperparameter of our method.
%

We assume that stochastic estimates are revealed independently at each iteration , and we denote by $\filtration_t$ the filtration generated by the random variables up to iteration $t$. 
Let us further introduce the intermediate \sv{$\sigma$-algebras} $\filtration_{t, \lowervar} = \sigma(\filtration_t, \lowervar_{t+1})$ and $\filtration_{t, \dualvar} = \sigma(\filtration_{t, \lowervar}, \dualvar_{t+1})$.
We complement the previous assumptions with the following standard conditions on our stochastic estimates.
%
\begin{assumption}
    \label{assumption:stochastic_gradients}
    The stochastic gradients $\stonabla_{\lowervar} \lowerobj$, $\stonabla_{\lowervar, \lowervar} \lowerobj$, $\stonabla_{\lowervar, \uppervar}^2 \lowerobj$, $\stonabla_{\lowervar} \upperobj$ and $\stonabla_{\uppervar} \upperobj$ are independent unbiased estimates of the corresponding true gradients, with bounded variance. That is, for all $\lowervar \in \RR^d$ and $\uppervar \in \cC$:
    {\small
    \begin{align*}
        &\expectation[\stonabla_{\lowervar} \lowerobj(\lowervar_t, \uppervar_t) \mid \filtration_t] = \nabla_{\lowervar} \lowerobj(\lowervar, \uppervar) &&\expectation[\|\stonabla_{\lowervar} \lowerobj(\lowervar, \uppervar) - \nabla_{\lowervar} \lowerobj(\lowervar, \uppervar)\|^2 \mid \filtration_t] \leq \cV_{\lowerobj, 1},\\
        &\expectation[\stonabla_{\lowervar, \lowervar}^2 \lowerobj(\lowervar, \uppervar) \mid \filtration_{t, \lowervar}] = \nabla_{\lowervar, \lowervar}^2 \lowerobj(\lowervar, \uppervar) &&\expectation[\|\stonabla_{\lowervar, \lowervar}^2 \lowerobj(\lowervar, \uppervar) - \nabla_{\lowervar, \lowervar}^2 \lowerobj(\lowervar, \uppervar)\|^2 \mid \filtration_{t, \lowervar}] \leq \cV_{\lowerobj, 2}^{\lowervar, \lowervar},\\
        &\expectation[\stonabla_{\lowervar, \uppervar}^2 \lowerobj(\lowervar, \uppervar) \mid \filtration_{t, \dualvar}] = \nabla_{\lowervar, \uppervar}^2 \lowerobj(\lowervar, \uppervar) &&\expectation[\|\stonabla_{\lowervar, \uppervar}^2 \lowerobj(\lowervar, \uppervar) - \nabla_{\lowervar, \uppervar}^2 \lowerobj(\lowervar, \uppervar)\|^2 \mid \filtration_{t, \dualvar}] \leq \cV_{\lowerobj, 2}^{\lowervar, \uppervar}\\
        &\expectation[\stonabla_{\lowervar} \upperobj(\lowervar, \uppervar) \mid \filtration_{t, \lowervar}] = \nabla_{\lowervar} \upperobj(\lowervar, \uppervar)
        &&\expectation[\|\stonabla_{\lowervar} \upperobj(\lowervar, \uppervar) - \nabla_{\lowervar} \upperobj(\lowervar, \uppervar)\|^2 \mid \filtration_{t, \lowervar}] \leq \cV_{\upperobj, 1}^{\lowervar},\\
        &\expectation[\stonabla_{\uppervar} \upperobj(\lowervar, \uppervar) \mid \filtration_{t, \dualvar}] = \nabla_{\uppervar} \upperobj(\lowervar, \uppervar)
        &&\expectation[\|\stonabla_{\uppervar} \upperobj(\lowervar, \uppervar) - \nabla_{\uppervar} \upperobj(\lowervar, \uppervar)\|^2 \mid \filtration_{t, \dualvar}] \leq \cV_{\upperobj, 1}^{\uppervar}.
    \end{align*}
    }
    where $\cV_{\lowerobj, 1}, \cV_{\lowerobj, 2}^{\lowervar, \lowervar}, \cV_{\lowerobj, 2}^{\lowervar, \uppervar}, \cV_{\upperobj, 1}^{\lowervar}, \cV_{\upperobj, 1}^{\uppervar} > 0$ are positive constants.
\end{assumption}
We present in the following theorem our convergence result for~\stoalgoname. 
Besides the starting error $\varepsilon_{\lowervar, 0}$ and $\varepsilon_{\dualvar, 0}$, we also introduce $\varepsilon_{\lowervar, \dualvar, 0}^2 \defineq \|\lowervar_0 - \lowersol(\uppervar_0)\|^2 \|\dualvar_0\|^2$ which motivations are stated in the forthcoming remarks.
\begin{theorem}\label{thm:convergence_stochastic}
Consider running the iteration~\eqref{eq:badr_stochastic_updates} for $T$ steps with step sizes $\tau, \rho, \gamma$ satisfying $\tau=1/L_{\lowerobj, 1}^{\lowervar, \lowervar}, \rho=\min(1/L_{\lowerobj,1}^{\lowervar, \lowervar}, \frac{\mu_\lowerobj^2}{4\cV_{\lowerobj, 2}^{\lowervar, \lowervar}})$, and $\gamma=\min(\bar{\bar \gamma}, L_{\varphi}^{-1})$ where 
denotes  the smoothness constant of the implicit loss $\varphi$ derived in Lemma~\ref{lem:lipschitz_implicit_solution} and $\bar{\bar \gamma}$ a constant made explicit in~\eqref{eq:def:barbargamma}. 
Then, under Assumptions~\ref{assumption:upperobj}, ~\ref{assumption:lowerobj}, and~\ref{assumption:stochastic_gradients}, the sequence $(\lowervar_t, \uppervar_t)_{t \geq 0}$ generated by~\eqref{eq:badr_updates} satisfies for all $T \geq 1$,
{\small
\begin{align}\label{eq:thm_convergence_stochastic}
    \frac{1}{T} \sum_{t=0}^{T-1} \expectation\left[\|\pgstep_{\uppervar, t}\| \right] + \tilde{\alpha}^{1/2}\;  \expectation[\|\lowervar_t - \lowersol_t\|]  
        \leq \mathcal{S}\Bigg[\frac{8 \Delta \implicitobj}{\gamma T}
            \!+\! \frac{32 \kappa_{\lowerobj}}{T} \left(\tilde \alpha\varepsilon_{\lowervar, 0}^2 \!+\!  \tilde \beta  \varepsilon_{\dualvar, 0}^2 \!+\! \tilde \delta \varepsilon_{\lowervar, \dualvar, 0}^2 \right) 
            \!+\! \boldsymbol{M}^\top \cV 
            \!+\! r\; C_\gamma^2\Bigg]
\end{align}
}
where $\mathcal{S}[t] \defineq t/C_\gamma + \sqrt{t}$, $\Delta \implicitobj \defineq \implicitobj(\uppervar_0) - \inf_{\uppervar \in \cC} \implicitobj(\uppervar)$, and $\tilde{\alpha}, \tilde{\beta}, \tilde{\delta}$ are problem-dependent constants specified in~\eqref{eq:def_tilde_params_sto}. 
The vector $\boldsymbol{M} \in \RR^{5}$ is defined in~\eqref{eq:def_M_sto}, the scalar $r$ is defined as $r \defineq 18\, L_{\varphi}^{-2} M_\cC^2 \big(L_{\lowerobj, 2}^{\lowervar, \lowervar}\big)^2 \constant_{\dualvar, \star} \mu_{\lowerobj}^{-4}$, and $\cV \defineq [\cV_{\upperobj, 1}^{\uppervar}, \cV_{\upperobj, 1}^{\lowervar}, \cV_{\lowerobj, 1}, \cV_{\lowerobj, 2}^{\lowervar, \uppervar}, \cV_{\lowerobj, 2}^{\lowervar, \lowervar}]^\top$ collects the variances specified in Assumption~\ref{assumption:stochastic_gradients}.
\end{theorem}
%
%
The proof of this theorem is deferred to Section~\ref{sec:proof_convergence_stochastic} of the Appendix.
\begin{remark}
In Theorem~\ref{thm:convergence_stochastic}, the overall convergence metric used also jointly quantifies optimality at the implicit and lower levels.
Regarding optimality at the implicit level, the metric used is the expected \st{norm of the} stochastic generalized gradient $\expectation\left[\|\pgstep_{\uppervar, t}\| \right]$, which slightly differs from the standard metric in stochastic non-convex proximal optimization~\citep{ghadimi2016mini}, which would in place involve $\expectation\left[\|\apgstep_{\uppervar, t}\|^2 \right]$, the expected squared norm of the generalized stochastic gradient.
%
Besides having the expectation outside of the norm -- which we argue allows for a better characterization of stationarity, as randomness only concerns the path taken up to $\uppervar_t$ and not the stochastic gradient at time $t$ -- we note note that our results present convergence in gradient norm instead of squared gradient norm, which is a common feature in analyses involving clipping~\citep{koloskova2023revisiting}.
\end{remark}
\begin{remark}
The right-hand side of our convergence rate involves three components. First two components upper bound respectively the speed at which we forget the initial conditions and the variance of our stochastic estimates. 
We note that the matrix $\boldsymbol{M}$ depends on the problem constants introduced in \cref{assumption:upperobj}, and does not vanish as $\gamma$ goes to zero. Hence, to achieve an $\varepsilon$-stationary point, one may resort to mini-batching to reduce the noise vector $\cV$ to an appropriate norm.
In particular, setting both the minibatch size $m$ and the number of iterations $T$ to scale as $\varepsilon^{-2}$ yields an overall complexity of order $\mathcal{O}(\varepsilon^{-4})$, which matches the known lower bound with respect to $\varepsilon$ for stochastic bilevel problems~\citep{chen2025condition}. We do not, however, investigate optimality with respect to the condition number $\kappa$, as our framework addresses a broader class of problems, and establishing corresponding lower bounds in this regime lies beyond the scope of the present work.
%
\end{remark}
\begin{remark}
We prove Theorem~\ref{thm:convergence_stochastic} based on the derivation of a novel Lyapunov function $\lyap_t$ defined as 
\[
    \lyap_t \defineq \expectation[\implicitobj(\uppervar_t) + \alpha \|\lowervar_t - \lowersol(\uppervar_t)\|^2 + \beta \|\dualvar_t - \dualsol(\uppervar_t)\|^2 + \delta \|\lowervar_t - \lowersol(\uppervar_t)\|^2 \|\dualvar_t\|^2]  
\]
where $\alpha, \beta, \delta > 0$ are carefully tuned parameters. 
The decrease of $\lyap_t$ is made possible thanks to the clipping of the upper-level update, which allows specifically to control the drift on the fourth term $\expectation[\delta \|\lowervar_t - \lowersol(\uppervar_t)\|^2 \|\dualvar_t\|^2]$ of our Lyapunov function.
This term is responsible for the presence of the additional starting error $\varepsilon_{\lowervar, \dualvar, 0}^2$ in our convergence rate.
\end{remark} }

\section{Open-source software}\label{sec:toolbox}
\st{We provide \tbadr, an open-source Python library that implements the adaptive reweighting methodology presented in \cref{sec:preliminaries}.
\tbadr follows the \tsklearn interface, allowing users to train and make predictions with a model through simple function calls directly from \tpandas dataframes.
In the following, we describe the toolbox's interface and give simple examples of a \tbadr workflow.
We refer to the online documentation for more details and advanced usage of the toolbox.
\subsection{Interface}
We show the class structure of the \tbadr toolbox in \cref{fig:architecture} below.
\begin{figure}[ht]
    \centering
    \resizebox{.6\linewidth}{!}{%
    \begin{tikzpicture}[node distance=3mm and 2cm]

\node[
    module={SoftBlue},
    minimum height=3cm,
    ] (datasets) {};
\node at (datasets.north) [font=\sffamily\bfseries,yshift=-13pt] {\textcolor{black}{\faDatabase} Datasets};

\begin{scope}
    \coordinate (mid) at (datasets.center);
    \node[helper={SoftBlue}, anchor=north] 
        at ([yshift=5mm] mid) (o1) {Custom DataFrames};
    \node[helper={SoftBlue}, anchor=north] 
        at ([yshift=-5mm] mid) (o2) {Existing fetchers};
\end{scope}

\node[
    module={SoftRed},
    below= of datasets,
    minimum height=3cm, 
    inner sep=6pt
    ] (metrics) {};
\node at (metrics.north) [font=\sffamily\bfseries,yshift=-13pt] {\textcolor{black}{\faChartBar} Fairness metrics};

\begin{scope}
    \coordinate (mid) at (metrics.center);
    \node[helper={SoftRed}, anchor=north] 
        at ([yshift=7mm] mid) (o1) {Group Variance};
    \node[helper={SoftRed}, anchor=north] 
        at ([yshift=-3mm] mid) (o2) {Demographic Parity};
    \node at ([yshift=-13mm] mid) (o3) {\textcolor{black}{\textbf{...}}};
\end{scope}

\node[
    module={SoftGreen},
    below= of metrics,
    minimum height=4cm, 
    inner sep=6pt
    ] (models) {};
\node at (models.north) [font=\sffamily\bfseries,yshift=-13pt] {\textcolor{black}{\faProjectDiagram} Learning Models};

\begin{scope}
    \coordinate (mid) at (models.center);
    \node[helper={SoftGreen}, anchor=north] 
        at ([yshift=10mm] mid) (o1) {Linear Regression};
    \node[helper={SoftGreen}, anchor=north] 
        at ([yshift=0mm] mid) (o2) {Logistic Regression};
    \node[helper={SoftGreen}, anchor=north] 
        at ([yshift=-10mm] mid) (o3) {Smoothed SVM};
\end{scope}

\node[
    module={SoftOrange},
    minimum height=3cm,
    right=3cm of metrics,
    yshift=0cm
    ] (oracles) {};
\node at (oracles.north) [
    font=\sffamily\bfseries,
    align=center,
    yshift=-13pt  
    ] {\textcolor{black}{\faLayerGroup} Oracles};
\node at (oracles.north) [
    font=\sffamily\bfseries,
    align=center,
    yshift=10pt
    ] {\footnotesize \textcolor{orange}{\faInfoCircle{}} Abstraction layer};

\begin{scope}
    \coordinate (mid) at (oracles.center);
    \node[helper={SoftOrange}, anchor=north] 
        at ([yshift=5mm] mid) (o1) {Implicit Oracle};
    \node[helper={SoftOrange}, anchor=north] 
        at ([yshift=-5mm] mid) (o2) {Stochastic Oracle};
\end{scope}

\node[
    module={SoftGray},
    minimum height=7cm,
    minimum width=4cm,
    right=2cm of oracles,
    yshift=-1.3cm
    ] (algorithms) {};
\node at (algorithms.north) [
    font=\sffamily\bfseries,
    align=center,
    yshift=-13pt  
    ] {\textcolor{black}{\faCogs} Algorithms};

\begin{scope}
    \coordinate (mid) at (algorithms.center);
    \node at (algorithms.center) [
    font=\sffamily\bfseries,
    align=center,
    yshift=25mm
    ] (twoloops) {\footnotesize Two-loop methods};
    \node[helper={SoftGray}, anchor=north] 
        at ([yshift=20mm] mid) (a1) {Frank-Wolfe};
    \node[helper={SoftGray}, anchor=north] 
        at ([yshift=10mm] mid) (a2) {SLSQP};
    \node at (algorithms.center) [
    font=\sffamily\bfseries,
    align=center,
    yshift=-5mm
    ] (singleloops) {\footnotesize Single-loop methods};
    \node[helper={SoftGray}, anchor=north] 
        at ([yshift=-10mm] mid) (a3) {\algoname};
    \node[helper={SoftGray}, anchor=north] 
        at ([yshift=-20mm] mid) (a4) {\stoalgoname};
\end{scope}

\begin{scope}[on background layer]
\node[group, fit=(datasets)(models)(metrics)] (Gtop) {};
\node[grouplabel,anchor=south] at (Gtop.north) {User interface};

\node[group, fit=(oracles)(algorithms)]    (Gmid) {};
\node[grouplabel,anchor=south] at (Gmid.north) {System layer};
\end{scope}

\path[->,draw] (Gtop) edge node[sloped,above]{instantiates}
(oracles);
\path[->]
    (twoloops.west) 
      edge
      node[sloped,above]{uses}
    (o1.east);
\path[->]
    (singleloops.west) 
      edge
      node[sloped,above]{uses}
    (o2.east);

\end{tikzpicture}%
    }
    \caption{\textbf{Overview of the \tbadr package modules.}}
    \label{fig:architecture}
\end{figure}
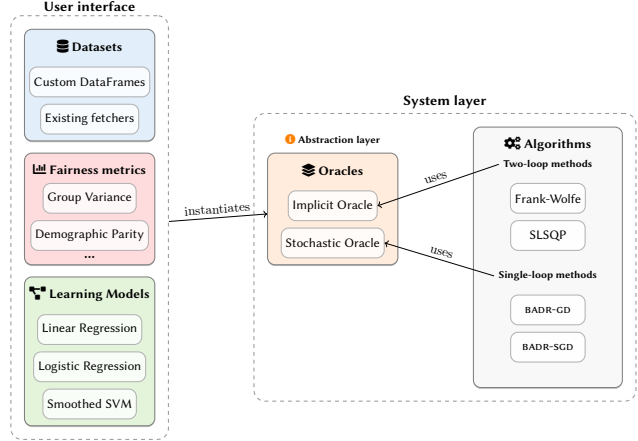
There exists a clear separation between the machine learning task (consisting of a dataset, a fairness metric and a learning model) and the algorithms used to solve the adaptive reweighting problem as stated in \cref{sec:bilevel}.
\subsection{Data}
Users of \tbadr are not expected to prepare custom data loaders or write custom training loops.
They can instead directly use a \tpandas dataframe with the three lines below, which create a Dataset object, perform a train/test split, normalize the feature matrix and partition the data with respect to the specified sensitive attributes.
\begin{python}
from badr.datasets import load_dataframe
df = pd.read_csv(...) # load your dataset
dset = load_dataframe(df = df, target_col = "y", sensitive_cols = ["sex"])
\end{python}
\subsection{Basic usage}
We start by importing the \tbadr package and defining a problem, i.e. a dataset, a learning model and a fairness metric.
\begin{python}
import badr as bdr

# == Problem Definition ==#
dset = bdr.datasets.fetch_adult()
model = bdr.models.LogisticRegression()
metric = bdr.metrics.IndividualFairness()
\end{python}
We can mimic \tsklearn's behavior using the \tbadr estimators. As an example, we show below the fitting and scoring of the \tsklearn LogisticRegression estimator.
\begin{python}
(*@\textcolor{BrickRed}{\#== NO FAIRNESS ENFORCED ==\#}@*)
(*@\textcolor{BrickRed}{model.fit(dset.X\_train, dset.y\_train, dset.groups)}@*)
(*@\textcolor{BrickRed}{test\_score = model.score(dset.X\_test, dset.y\_test)}@*)
(*@\textcolor{BrickRed}{test\_fairness = metric.fun(model.coef\_, dset)}@*)
\end{python}
Building a Pareto-fair counterpart to the \tsklearn model can be done with one additional line, as shown below.
\begin{python}
#== BADR ==#
(*@\textcolor{ForestGreen}{badr = bdr.Badr(dset, model, metric)}@*)
(*@\textcolor{ForestGreen}{badr.run()}@*)
(*@\textcolor{ForestGreen}{badr\_test\_score = badr.model.score(dset.X\_test, dset.y\_test)}@*)
(*@\textcolor{ForestGreen}{badr\_fairness = metric.fun(badr.coef\_, dset)}@*)
\end{python}
To fit the Pareto-fair model, we first instantiate a \tbadr object with the dataset, the learning model and the fairness metric and call \texttt{.run()}.
Depending on the size of the dataset, either an implicit or a stochastic oracle is instantiated, which serves as an abstraction layer between the fairness problem and the proposed algorithms.
The \texttt{.run()} method performs two tasks: (1) it executes a suitable optimization algorithm to compute an adaptive reweighting and (2) its fits the learning model with the computed per-group weights.
%
\subsection{Algorithms}\label{sub:algorithms}
The \tbadr toolbox mainly includes an implementation of \cref{alg:badr,alg:stochastic_badr}, the large-scale algorithms \algoname and \stoalgoname to solve the bilevel problem \eqref{eq:bilevel_formulation}.
We also allow the use of two-loop algorithms to solve small and medium scale instances of \eqref{eq:bilevel_formulation}.

Two-loop methods for bilevel optimization work as follows: at each iteration, the lower-level problem is (approximately) solved using an iterative method (inner loop).
Its solution is used to compute an implicit gradient which is used to update the outer variable, also using an iterative method (outer loop).
This is a standard approach to solving bilevel optimization problems \citep{domke2012generic, ghadimi2018approximation}, but it can be computationally expensive and less scalable than single-loop algorithms, as the lower-level must be solved multiple times, and solving the adjoint equation to compute the gradient of the solution mapping typically requires a matrix inversion.
\subsubsection{Default choices for the optimization method}\label{subsec:exp-optimization}
In the remainder of this section, we present the optimization methods available in the toolbox for two-loop algorithms.
We outline the default parameters and stopping criteria for each method.
We identify the default two-loop method and justify this choice.
We discuss how the wall-clock time of each method scales with an increasing number of groups.
Finally, we compare the scalability of the default two-loop method with \stoalgoname.
\myparagraph{Implementation}
We implement the two-loop algorithms by (i) solving the lower-level problem with the \tsklearn models used in our experiments (ridge regression, the $\ell_2^2$-regularized logistic regression model, and the $\ell_2^2$-regularized smoothed SVM) and (ii) updating the outer variable using an implicit gradient.
For logistic regression and smoothed SVM, we obtain the lower-level gradient and Hessian with \tjax \citep{jax} and solve the resulting linear system with \textsc{numpy}; for ridge regression, we solve it directly with \textsc{cvxpy} \citep{cvxpy_jcd, cvxpy_jmlr} and the \textsc{clarabel} solver \citep{Clarabel_2024} (default settings).
\myparagraph{Available optimization methods}
As outlined in \cref{sub:algorithms}, \eqref{eq:bilevel_formulation} can either be solved using \cref{alg:badr,alg:stochastic_badr} or recast as the minimization of a smooth nonconvex function on the unit probability simplex with available function and gradient evaluations.
We consider three optimization methods to solve this problem: SLSQP \citep{solver_slsqp2}, the trust-region method \citep{solver_trustregion} and the Frank-Wolfe method \citep{solver_fw}.
Both SLSQP and the trust-region method are implemented in \textsc{SciPy}, with SLSQP calling a C backend.
For Frank-Wolfe, we adapt the reference implementation from the book repository \citep{solver_fw} to support \textsc{jax} arrays and just-in-time compilation, and we use the oblivious stepsize $\tfrac{2}{t+2}$ instead of backtracking \citep{fw_line_search}, since repeated implicit-gradient evaluations dominated any gains from line-search stepsizes.
\myparagraph{Stopping criteria} 
Each of the three optimization methods has its own stopping criterion, in addition to a maximum iteration limit of $500$.
SLSQP stops when the absolute difference between consecutive function values falls below $10^{-5}$.
The trust-region method stops when the norm of the Lagrangian gradient falls below $10^{-5}$.
The Frank-Wolfe method stops when the Frank-Wolfe gap \cite[Definition 1.11]{solver_fw} (which lower-bounds the current iterate's suboptimality) falls below $10^{-6}$.
\myparagraph{Default method}
In this section, we justify our selection of SLSQP as the default optimization method for the two-loop approach.
We conducted a systematic experiment on the 10 largest subsamples of the 2014 ACS Employment dataset (the 10 most populated U.S. states).
Each subsample was partitioned into two to six sensitive groups, yielding $50$ (state, number of groups) instances.
We used the Disparate Mistreatment \citep{metric_zafar} fairness metric.
For each instance, we ran all three solvers and recorded the best fairness value $\fairnessmetric_{best}$ attained by any solver.
We then defined the target precision as $\epsilon = \fairnessmetric_{best} + \varepsilon$ and measured the wall-clock time to reach this target across all five partitionings (two to six groups).
Performance profiles generated for three precision values: $\varepsilon = 10^{-3}$, $10^{-5}$, and $10^{-7}$ are provided in \cref{fig:performance_profiles}.

\begin{figure}[htbp]
  \centering
  \subfigure[$\varepsilon = 10^{-3}$]{%
    \includegraphics[width=.3\textwidth]{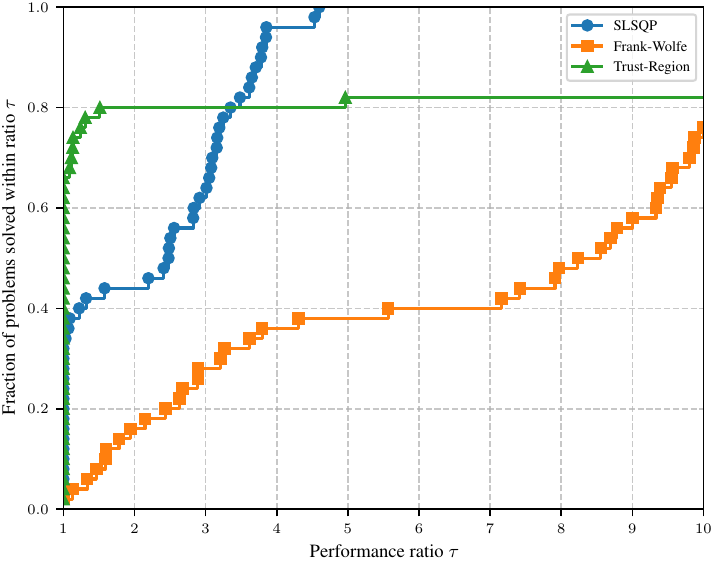}%
    \label{fig:performance_profile_3}%
  }\hfill
  \subfigure[$\varepsilon = 10^{-5}$]{%
    \includegraphics[width=.3\textwidth]{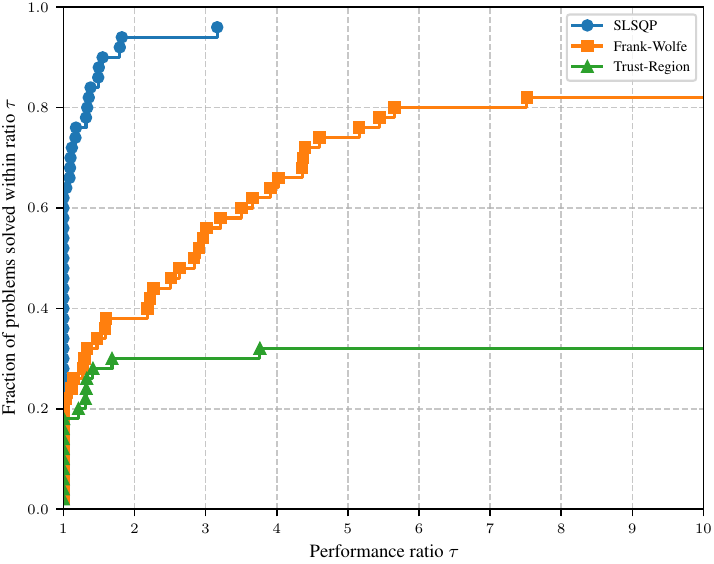}%
    \label{fig:performance_profile_5}%
  }\hfill
  \subfigure[$\varepsilon = 10^{-7}$]{%
    \includegraphics[width=.3\textwidth]{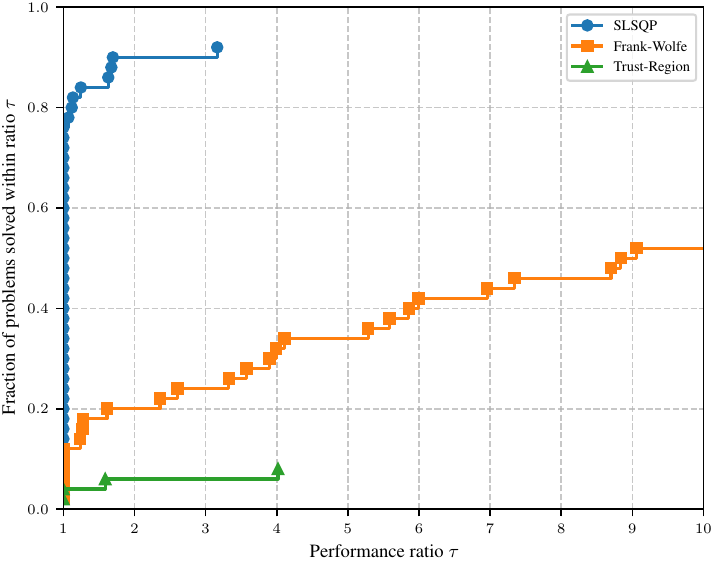}%
    \label{fig:performance_profile_7}%
  }
  \caption{Performance profiles comparing SLSQP, Frank-Wolfe, and Trust-Region methods for achieving ${\fairnessmetric}_{\text{best}} + \varepsilon$ across multiple tolerance levels.
  The vertical axis represents the fraction of instances solved, while the horizontal axis shows the factor $\tau$ of wall-clock time relative to the fastest algorithm.
  Higher curves indicate superior algorithm efficiency.
  Rightward-extending curves identify instances where algorithms demonstrate slower convergence relative to the optimal method.}
  \label{fig:performance_profiles}
\end{figure}
\noindent Our main findings are summarized below:
\begin{itemize}[leftmargin=*, topsep=0.35\baselineskip, itemsep=0.25\baselineskip,
                parsep=0pt, partopsep=0pt]
    \item \myparagraph{SLSQP consistently solves the most instances} SLSQP outperforms both Frank-Wolfe and the trust region methods in terms of fraction of problems solved across all precision values. SLSQP achieves 100\% solution rate for low precision ($\varepsilon = 10^{-3}$) and approximately $94\%$ for moderate and high precision ($\varepsilon \in \{10^{-5},10^{-7}\}$).
    In contrast, although the trust-region method solves around $80\%$ of instances at low precision, its performance degrades considerably when asking for higher precisions.
    \item \myparagraph{SLSQP remains within a factor of three of the fastest method for moderate and high precision} For $\varepsilon \in \{10^{-5},10^{-7}\}$, approximately 90\% of problems are solved by SLSQP with $\tau < 2$, indicating it rarely requires more than twice the computational time of the best solver. When slower, it typically remains within $\tau \approx 3$.
\end{itemize}

\subsubsection{Scalability}
We evaluate how runtime of the proposed methods evolve as problem size increases along two axes: the number of sensitive groups and the sample size.

\textbf{Groups.} We use a dataset derived from the 2018 ACSTravelTime data \citep{dataset_folktables} with up to $36$ groups, obtained by crossing age ($<20$ vs. $\geq 20$), race (nine categories), and gender (male vs. female).
We predict whether an individual's commute exceeds $20$ minutes (the 2018 U.S. median) with an $\ell_2^2$-regularized logistic regression model and the Disparate Mistreatment metric.
For each \tbadr instantiation, we record the wall-clock time to first reach a fairness value of $10^{-4}$ on subsamples from seven randomly selected states (Alaska, Arkansas, Hawaii, Idaho, New Hampshire, New Mexico, and Utah).
Results are shown in \cref{fig:scale_groups}.
\begin{figure}[htbp]
    \centering
    \includegraphics[width=\linewidth]{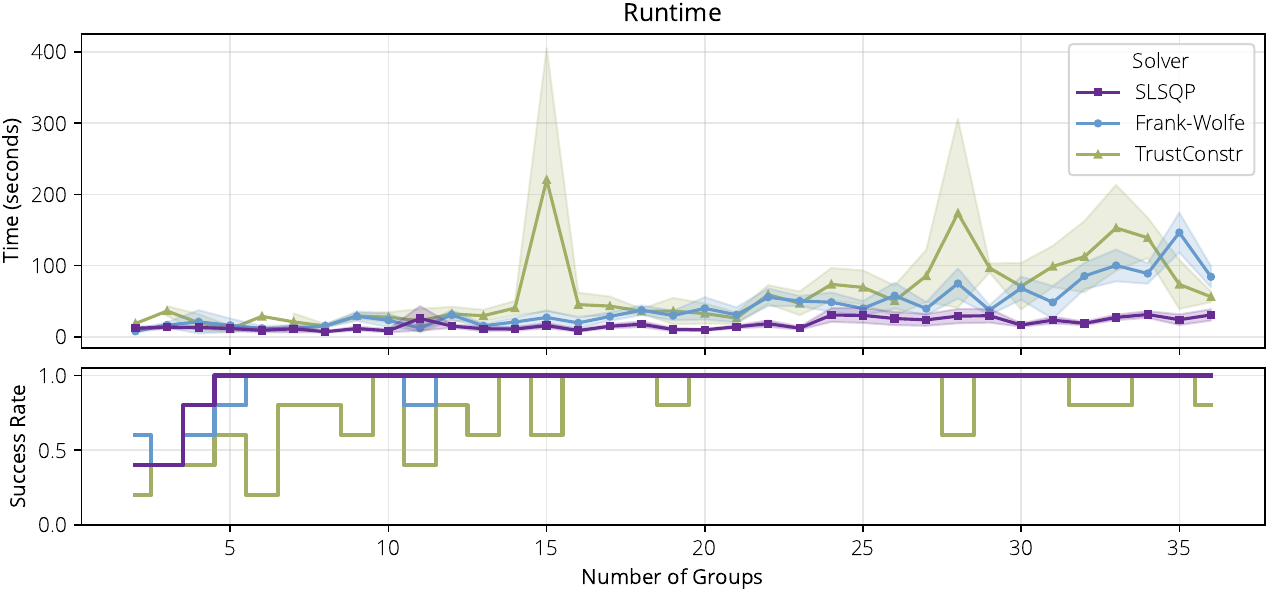}
    \caption{\small{
    (Top): Median wall-clock time (solid) with 25\% and 75\% interquartile ranges (shaded) computed over successful runs across five US states for SLSQP, Frank-Wolfe, and the trust-region method to first reach a fairness value of $10^{-4}$, as a function of the number of sensitive groups.
    (Bottom): Success rate of each method over the seven chosen states as a function of the number of sensitive groups.}}
    \label{fig:scale_groups}
\end{figure}
\noindent For fewer than $10$ groups, SLSQP and Frank-Wolfe have similar mean runtime.
The trust-region method is comparable on successful runs but fails much more often for $2$ to $5$ groups.
Beyond $10$ groups, SLSQP remains nearly constant in mean runtime, with a noticeably narrower interquartile range, whereas Frank--Wolfe and the trust-region method slow down.
Finally, SLSQP's success rate is never below the others except at $12$ groups (6 successful runs for SLSQP vs. 7 for Frank--Wolfe).
These results motivate our choice of SLSQP as the default optimizer for the two-loop bilevel approach.

\textbf{Sample size.} We compare the two-loop method (with SLSQP) and our stochastic algorithm \stoalgoname by scaling the sample size, as reported in \cref{fig:badr-scales}.
\begin{figure}[ht]
\centering
\includegraphics[width=.8\linewidth]{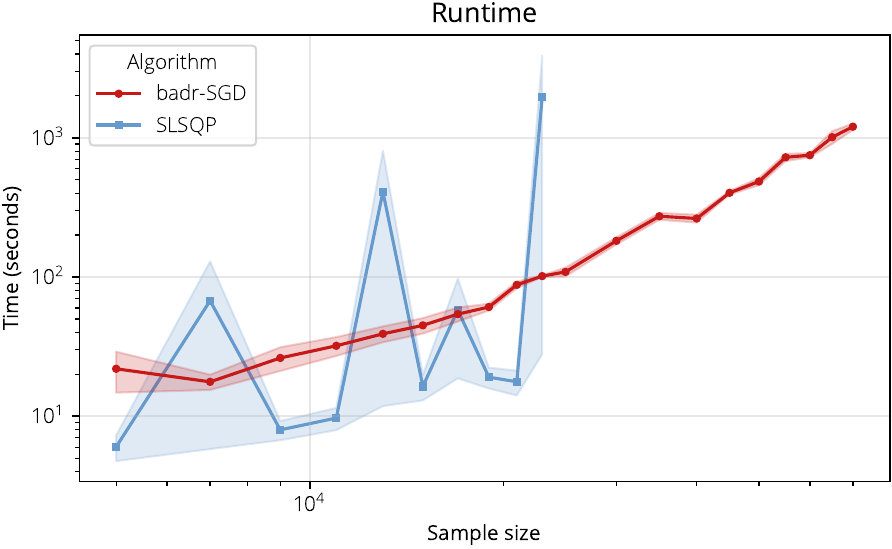}
\caption{Runtime scaling of \stoalgoname and SLSQP with dataset size (mean $\pm$ standard deviation over 5 seeds).}
\label{fig:badr-scales}
\end{figure}
Both methods train a fair logistic regression model with $\ell_2^2$-regularization parameter $10^{-1}$ on gender-partitioned subsamples of the California ACSEmployment dataset.
For each sample size, we draw 5 subsamples (different seeds) and report mean training time and standard deviation.
Experiments were run on a laptop with $32$GB of RAM.
SLSQP ran out of memory at $25{,}000$ samples, indicating that it can be competitive while it fits in memory but becomes impractical as sample size increases.

}

\section{Numerical Experiments}\label{sec:experiments}
\st{In this section, we provide a detailed description of the experiments conducted to analyze the effectiveness of the adaptive reweighting methodology. In \cref{subsec:exp-problem}, we describe the datasets used and the problem formulations considered. We study in \cref{subsec:exp-small,subsec:exp-arbitrary} whether our framework \frameworkname provides fairer solutions than other baselines. In \cref{subsec:exp-tradeoff}, we study how \frameworkname compares with other baselines in terms of test error.

\subsection{Dataset, Tasks and Methods}\label{subsec:exp-problem}
\subsubsection{Datasets}
We consider eleven datasets, summarized in Table~\ref{table:datasets} and described below in detail.
\begin{table*}[!ht]
\setlength{\tabcolsep}{6pt}
\renewcommand{\arraystretch}{1.1}
\small
\centering
\begin{tabular}{ccccrr}
\hline\hline
Task & Dataset & Target & Sensitive Attribute $(S)$ & $n$ & $d$ \\
\hline
\multirow{7}{*}{\rotatebox{90}{Classification}}
    & {\scriptsize ACSEmployment} 
      & {\scriptsize Employment} 
      & {\scriptsize Gender (2), Race (9)}        
      & 3157599 & 17 \\

    & {\scriptsize ACSIncome} 
      & {\scriptsize Income $\ge 50$K} 
      & {\scriptsize Gender (2), Race (9)}        
      & 1664500 & 11 \\ 

    & {\scriptsize ACSTravelTime} 
      & {\scriptsize Commute $\ge 20$min} 
      & {\scriptsize Gender (2), Age (2), Race (9)}        
      & 1376075 & 16 \\ 

    & {\scriptsize Adult}           
      & {\scriptsize Income $\ge 50$K}            
      & {\scriptsize Gender (2), Race (5)}     
      & 32561 & 109 \\ 

    & {\scriptsize COMPAS}          
      & {\scriptsize Recidivism in 2Y}          
      & {\scriptsize Gender (2), Race (6)}        
      & 6172 & 9 \\

    & {\scriptsize German Credit}  
      & {\scriptsize Credit (Good/Bad)}         
      & {\scriptsize Gender (2)}      
      & 1000 & 65 \\ 

    & {\scriptsize Arrhythmia}      
      & {\scriptsize Arrhythmia (Yes/No)}    
      & {\scriptsize Gender (2)}      
      & 418 & 272 \\
\hline

\multirow{4}{*}{\rotatebox{90}{Regression}}
    & {\scriptsize Law School (LSAC)}       
      & {\scriptsize GPA}         
      & {\scriptsize Gender (2), Race (5)}        
      & 22387 & 10 \\ 

    & {\scriptsize Parkinsons Telemonitoring} 
      & {\scriptsize UPDRS Score} 
      & {\scriptsize Gender (2)}      
      & 5875 & 21 \\

    & {\scriptsize Communities and Crime}     
      & {\scriptsize Violent Crimes} 
      & {\scriptsize Gender (2), Race (4)}        
      & 1993 & 97 \\ 

    & {\scriptsize Student Performance}       
      & {\scriptsize Final Grade} 
      & {\scriptsize Gender (2)}      
      & 395 & 54 \\ 
\hline\hline
\end{tabular}
\caption{Summary table of all real-world datasets considered. Reported sample size $(n)$ and number of features $(d)$ are post-processing.}
\label{table:datasets}
\end{table*}

The three ACS datasets (ACSEmployment, ACSIncome, and ACSTravelTime) are preprocessed following the guidelines in \citep{dataset_folktables}, while the remaining datasets use the preprocessing strategy described in \citep{preprocess_fairglm}.
All datasets are standardized to have zero mean and unit variance. 
Additionally, when using a linear model, we append an \emph{intercept} column of ones to the feature matrix.
For the three largest datasets (ACSEmployment, ACSIncome, ACSTravelTime), we also create smaller, subsampled versions by partitioning the original data by state (50 states + Puerto Rico) and year (2014 to 2018) yielding $51 \times 5 = 255$ smaller datasets with sizes ranging between 3,000 and 150,000 samples.
For large-scale experiments, we further partition the ACS datasets by the four statistical regions defined by the US Census Bureau (Northeast, Midwest, West, South), resulting in datasets containing between 300,000 and 600,000 samples.
In total, we obtain 785 datasets. 
Each dataset can be partitioned into sensitive groups, with the number of groups ranging from 2 to 36, depending on the attributes considered.
Moreover, each partitioned dataset can be analyzed using one of many unfairness metrics (we consider $7$ different metrics).
This makes a huge collection of test problems and allows us to thoroughly evaluate both small-scale and large-scale scenarios.
\subsubsection{Learning Models}

We consider three common models for supervised learning, which losses we describe below.

\begin{itemize}[leftmargin=*, topsep=0.35\baselineskip, itemsep=0.25\baselineskip,
                parsep=0pt, partopsep=0pt]
    \item \textbf{$\ell_2^2$-regularized logistic regression:}
    \begin{equation*}
        \ell (f(w,x_i), y_i) = \log \left(1 + \exp (-y_i \langle w, x_i \rangle) \right) + \frac{\rho}{2} \|w\|^2
    \end{equation*}
    This model is used for binary classification tasks.
    \item \textbf{$\ell_2^2$ regularized smoothed SVM:}
    \begin{equation*}
        \ell (f(w,x_i), y_i) = 
            \max\{0,\, 1 - y_i \langle w, x_i \rangle\}^2 + \frac{\rho}{2} \|w\|^2
    \end{equation*}
    This model is also used for binary classification tasks.
    \item \textbf{Ridge regression}
    \begin{equation*}
    \ell (f(w,x_i), y_i) = \left( y_i - \langle w, x_i \rangle \right)^2 + \frac{\rho}{2} \|w\|^2
    \end{equation*}
    This model is used for regression tasks.
\end{itemize}
Note that $\rho$ is a hyperparameter which controls the strength of the regularization parameter.

\subsubsection{Methods} \label{subsubsec:baselines}
\yl{\frameworkname} outputs a set of group weights $(\lambda_a^*)_{a\in\cA} \in \simplex{\attributenumber}$ and the associated fitted model $w^*(\lambda^*)$. For each unfairness metric $\fairnessmetric$, we compare four models obtained through the following strategies:

\begin{enumerate}[label=\alph*),leftmargin=*, topsep=0.35\baselineskip, itemsep=0.25\baselineskip,
                parsep=0pt, partopsep=0pt]
\item \textbf{Adaptive reweighting strategy:} we train the model using \tbadr, our adaptive reweighting framework,
\item \textbf{Uniform sampling strategy:} we train the model without group reweighting that is the group weights vector is the barycenter of the simplex $\simplex{\attributenumber}, \frac{1}{|S|} \mathbf{1}$,
\item \textbf{Balanced sampling strategy:} we train the model using group weights that rebalance the dataset, assigning each group a weight inversely proportional to its sample size and normalizing them to unit sum.
\item \textbf{One-group fitting strategy:} we train separately one model per sensitive group and keep the fairest model, that is the model trained on group $\arg\min_{a \in |\cA|}\cF \left( \theta^*\left(\mathbf{e_a}\right) \right)$,
\item \textbf{Minimax fairness:} we train the model following the minimax fairness approach \citep{martinez2020minimax,hashimoto2018fairness,diana2021minimax,abernethy2020active}, by solving the optimization problem defined in \cref{eq:def_minmax_fairness_pb}.
\end{enumerate}
We remind that these four strategies provide a Pareto-efficient model in terms of the group losses, and that we are mostly interested in whether the model trained 
\yl{with \frameworkname} is fairer than other Pareto-efficient models.
\subsubsection*{Hyperparameters}
Hyperparameters of \cref{alg:badr} are $\tau$ and $\gamma$ (note that we can set $\rho = \tau$, see \cref{thm:convergence_deterministic}).
We select both hyperparameters using a coarse-to-fine grid search.
\subsubsection*{Metric used}
We evaluate the models trained by each method on two aspects: (i) prediction performance, measured by classification error for binary classification and residual mean-squared error (RMSE) for regression, and (ii) the unfairness value of the trained model.
\subsubsection*{Hardware} All experiments were run on a laptop with an Intel i7 processor and 32GB of RAM, using the package \tbadr and its dependencies.
The experiments are conducted on the datasets described in \cref{subsec:exp-problem}.

\subsection{Experimental results}

In this subsection, we provide numerical examples demonstrating the Pareto dominance of solutions produced by \yl{\frameworkname}.
We first analyze the specific cases of two and three groups in \cref{subsec:exp-small}.
For these smaller group configurations, we visualize the implicit fairness function and plot the solutions returned by \yl{\frameworkname} alongside baseline methods.
We then examine the general case in \cref{subsec:exp-arbitrary}, plotting fairness value histograms on train and test sets for each unfairness metric across all five strategies (\yl{\frameworkname}, minimax fairness, one-group fitting, balanced sampling and uniform sampling).
In \cref{subsec:exp-tradeoff}, we turn our attention to how the predictive performance is impacted depending on the reweighting strategy.
\subsubsection{Experiment 1 : Comparison to baselines on two and three groups}\label{subsec:exp-small}
In this experiment, we visualize the value of the unfairness metric $\fairnessmetric$ with varying group weights, for two and three groups and with the $\ell_2^2$-regularized logistic regression model.
For two groups, we consider three different metrics and three different datasets and compare \yl{\frameworkname} with baselines for a partitioning by gender.
\cref{fig:exp1_2groups} illustrates our findings.
\begin{figure}[!htb]
    \centering
    \includegraphics[width=1\linewidth]{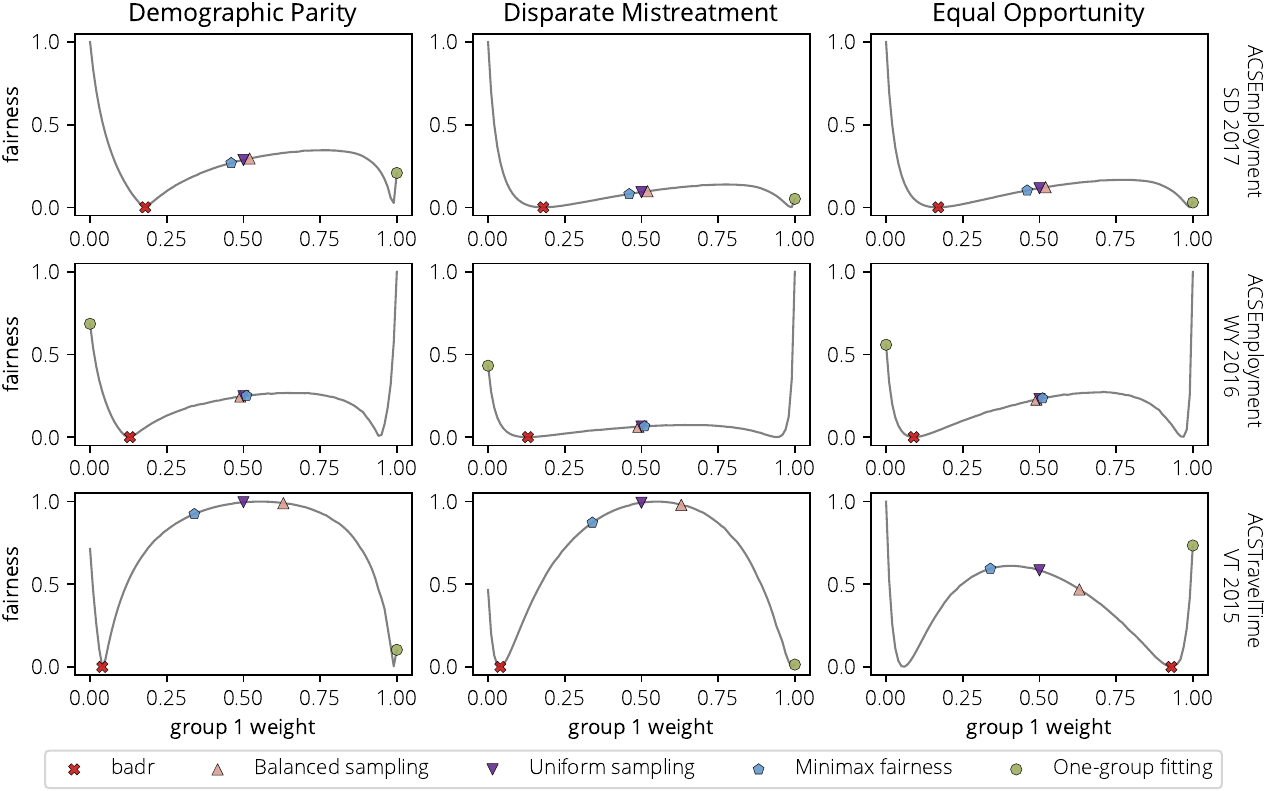}
    \caption{
    Visualization of the implicit fairness function for two groups for different metrics (columns) and different datasets (rows).
    Grey lines map the normalized fairness measure, representing the relative distance from the minimum achievable fairness value (y-axis) for varying group weights (x-axis).
    Markers indicate solutions found by each baseline.}
    \label{fig:exp1_2groups}
\end{figure}
The implicit fairness function represents the value of the unfairness metric $\fairnessmetric$ over the Pareto front of group losses.
\myparagraph{Optimization over the Pareto front is meaningful} We first observe that, for each unfairness metric and dataset, the implicit fairness function does vary, motivating the approach of optimizing an unfairness metric over the Pareto front.
\myparagraph{Mismatch between the sampling provided by the baselines and the unfairness metric}
We observe that the implicit function contains multiple local minima, and that the optimal sampling obtained using \tbadr is non-trivial and never attained by any of the baseline methods.
Moreover, simple and widely used heuristics such as balanced sampling \citep{mcmahan2017communication} do not necessarily improve fairness over standard uniform sampling. 

These observations are consistent on datasets partitioned in three groups, for which we propose a visualization in \cref{fig:exp1_3groups}.
\begin{figure}[!htb]
    \centering
    \includegraphics[width=1\linewidth]{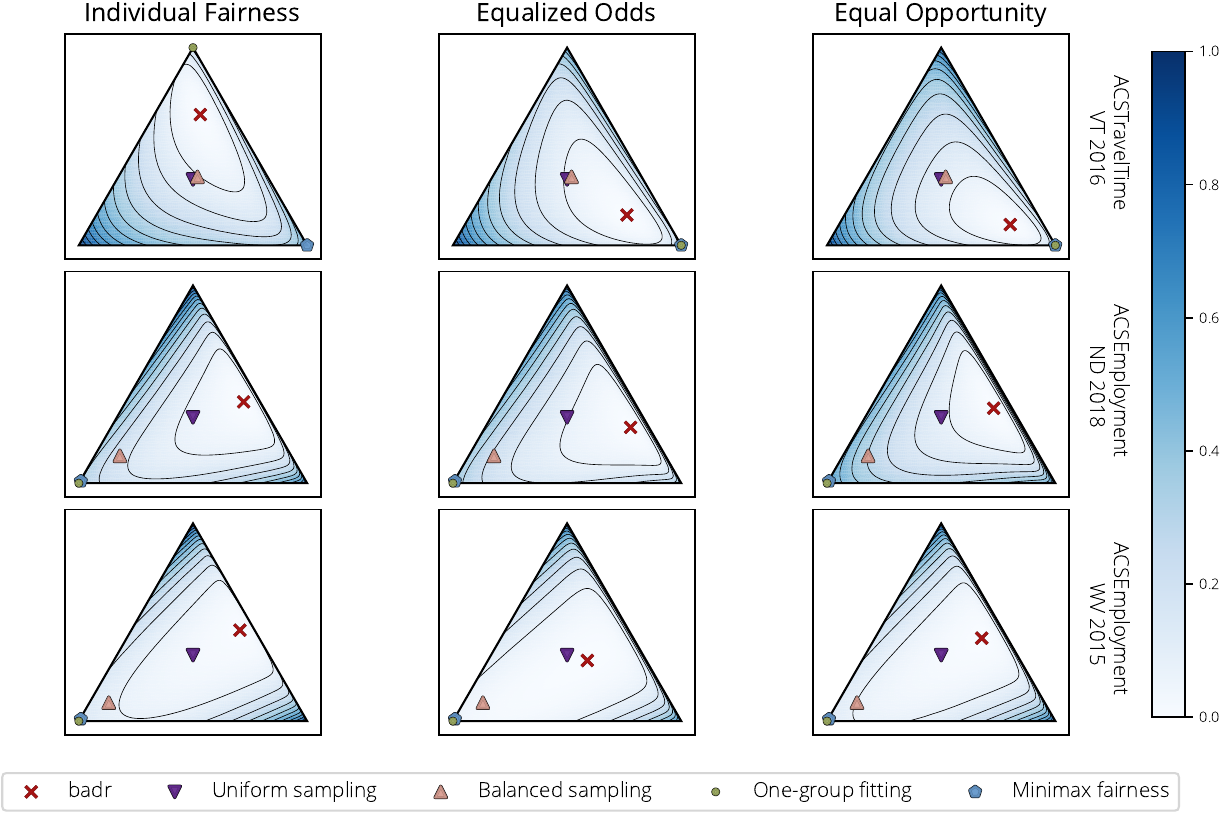}
    \caption{Normalized unfairness metric across three datasets (rows) and three metrics (columns) as a function of the group weights. Markers indicate solutions found by \tbadr and baselines. The three-dimensional unit probability simplex is plotted as an equilateral triangle through barycentric coordinate transformation.}
    \label{fig:exp1_3groups}
\end{figure}

\subsubsection{Experiment 2: Fairness attained on datasets partitioned in an arbitrary number of groups}
\label{subsec:exp-arbitrary}
For an arbitrary number of groups, it is no longer possible to visualize the value of the unfairness metric $\fairnessmetric$ by group weights.
In this experiment, we evaluate the models trained by each strategy (\yl{\frameworkname}, uniform sampling, balanced sampling, one-group fitting and minimax fairness) in terms of value of the unfairness metric $\fairnessmetric$ on both the train and test sets.
Our evaluation uses the ACSEmployment dataset for classification tasks.
We systematically examine all subsamples of the dataset (organized by state and year) and evaluate unfairness metrics across every possible sensitive group partitioning.
For each unfairness metric relevant to classification tasks, we record the fairness values achieved by $\tbadr$ and baseline methods.
We cluster results by unfairness metric type.
\cref{fig:hist-per-metric} shows cdf plots for each of the four strategies across both training and test sets.

\begin{figure}[htbp]
    \centering
    \includegraphics[width=.85\linewidth]{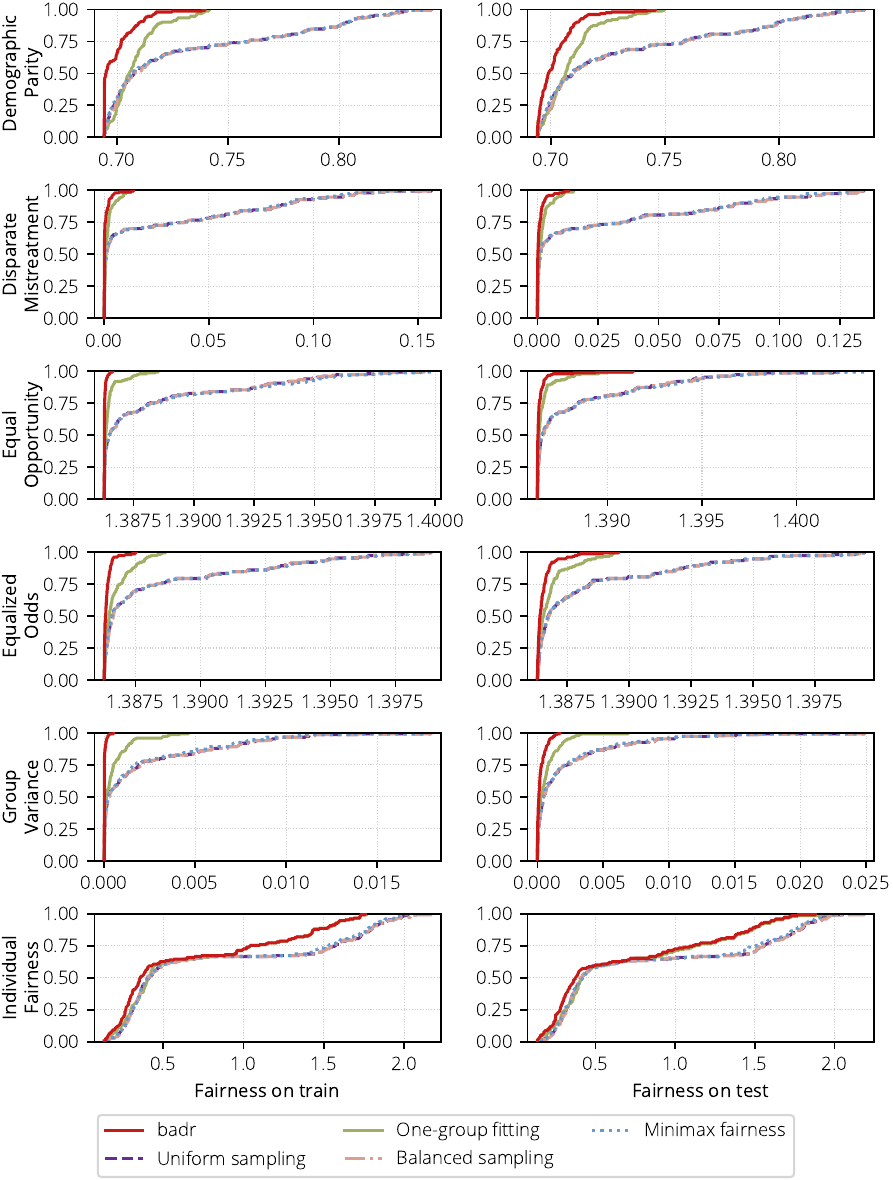}
    \caption{Empirical cumulative distribution functions (ECDFs) of six unfairness metrics (Demographic Parity, Disparate Mistreatment, Equal Opportunity, Equalized Odds, Group Variance, and Individual Fairness), computed using five approaches: \yl{\frameworkname}, Uniform sampling, Balanced sampling, One-group fitting and Minimax fairness.
    Curves are shown separately for the training data (left) and the test data (right).
    The horizontal axes show metric values, and the vertical axes show the empirical cumulative probabilities (fraction of runs with unfairness no larger than a given value).
    For each metric, curves that rise more quickly and lie closer to the top-left indicate better fairness performance.}
    \label{fig:hist-per-metric}
\end{figure}

Our main findings are summarized below.
\begin{itemize}[leftmargin=*, topsep=0.35\baselineskip, itemsep=0.25\baselineskip,
                parsep=0pt, partopsep=0pt]
    \item \myparagraph{\yl{\frameworkname} consistently yields improved fairness across all metrics} \yl{\frameworkname} and Minimax both yield fairness value distributions tightly clustered on the fair end (left), outperforming Uniform Sampling and One-Fit across all metrics.
    \item \myparagraph{\yl{\frameworkname} is the most robust in terms of fairness on the test set} The weights obtained by \yl{\frameworkname} still yield the best fairness on the test set. This contrasts with Minimax, which exhibits a wider right tail on the test set for several metrics: Equal Opportunity, Demographic Parity and Group Variance.
    \item \myparagraph{One-group fitting shows large variation and poor fairness performance} In both training and test sets, One-Fit produces wide ranges of fairness values. Its results differ markedly from the tight clusters formed by \yl{\frameworkname}, Minimax, and Uniform Sampling. 
\item \myparagraph{Uniform Sampling achieves moderate fairness consistency but performs worse than \yl{\frameworkname} and Minimax} Uniform Sampling matches \yl{\frameworkname} and Minimax for individual fairness. For other unfairness metrics, it underperforms compared to these two methods. Overall, it controls fairness better than One-Fit but less effectively than \yl{\frameworkname} and Minimax. 
\item \myparagraph{Fairness performance stability differs between training and testing} All methods show degradation in fairness when moving from training to test data. \yl{\frameworkname} is the least impacted by this degradation compared to other approaches. One-Fit is systematically worse. Minimax degrades notably for the Demographic Parity metric. Uniform Sampling, while robust, performs less consistently than \yl{\frameworkname} and Minimax.
\end{itemize}

\subsubsection{Experiment 3: Accuracy-fairness trade-off}\label{subsec:exp-tradeoff}
In this experiment, we evaluate the accuracy/fairness trade-off of the five Pareto-optimal strategies in \cref{subsubsec:baselines} on classification and regression tasks.
For classification, we use 51 subsamples each from ACSEmployment and ACSTravelTime (102 datasets total), partition each dataset by gender, and apply a 70/30 train-test split; we use logistic regression ($\ell_2^2$ regularization weight $\lambda=10^{-2}$) and individual fairness, and report mean test accuracy and mean test unfairness with standard deviations over the 102 datasets.
For regression, we use Law School, Parkinsons Telemonitoring, Communities and Crime, and Student Performance, with the same gender partitioning and 70/30 split.
We use ridge regression ($\ell_2^2$ regularization weight $\lambda=10^{-1}$), RMSE, and demographic parity, and report mean test RMSE and mean test unfairness with standard deviations averaged over four datasets, each with ten random train/test splits.
The results of the experiment are shown in \cref{tab:accuracy_fairness_tradeoff}.
\begin{table}[t]
\centering
\small
\setlength{\tabcolsep}{4pt}
\begin{tabular}{@{}lccccc@{}}
\toprule
& \multicolumn{2}{c}{Classification (102 runs)} && \multicolumn{2}{c}{Regression (40 runs)} \\
\cmidrule(lr){2-3}\cmidrule(lr){5-6}
Method
& Test accuracy $(\uparrow)$ & Test unfairness $(\downarrow)$
&& Test RMSE $(\downarrow)$ & Test unfairness $(\downarrow)$ \\
\midrule
\yl{\frameworkname}
& $0.7628 \pm 0.1231$ & $\mathbf{0.2947 \pm 0.1031}$
&& $2.5554 \pm 4.1021$ & $\mathbf{0.8666 \pm 0.1188}$ \\
Uniform sampling
& $0.7637 \pm 0.1219$ & $0.3432 \pm 0.1089$
&& $2.5085 \pm 4.0420$ & $1.0974 \pm 0.3672$ \\
Balanced sampling
& $0.7634 \pm 0.1221$ & $0.3470 \pm 0.1102$
&& $2.5168 \pm 4.0550$ & $1.0935 \pm 0.3596$ \\
Minimax fairness
& $0.7633 \pm 0.1223$ & $0.3372 \pm 0.1077$
&& $2.5314 \pm 4.0855$ & $1.0728 \pm 0.3453$ \\
One-group fitting
& $0.7585 \pm 0.1282$ & $0.3180 \pm 0.1152$
&& $2.6800 \pm 4.3114$ & $1.0239 \pm 0.2766$ \\
\bottomrule
\end{tabular}
\caption{Test accuracy and individual unfairness (classification; 102 runs) and test RMSE and demographic parity unfairness (regression; 40 runs), reported as mean $\pm$ std.}
\label{tab:accuracy_fairness_tradeoff}
\end{table}

\myparagraph{Better fairness without loss of accuracy}
All five baselines exhibit comparable mean predictive performance: mean test accuracy is very similar across methods for classification, and mean test RMSE is similarly close for regression, with differences that are minor compared with the reported standard deviations.
In contrast, fairness varies considerably.
\yl{\frameworkname} consistently yields the lowest test unfairness in both settings ($0.2947$ vs.\ $0.3180\text{-}0.3470$ for classification, and $0.8666$ vs.\ $1.02\text{-}1.10$ for regression), indicating improved fairness at comparable predictive performance.}

\section{Concluding remarks}
\st{In this work, we introduce \frameworkname, a bilevel framework for fair learning that selects, among group-wise Pareto-efficient models, the one that optimizes a user-specified fairness metric.
We achieve this by treating group scalarization weights as variables learned through a bilevel formulation: the lower level enforces Pareto efficiency of group losses, while the upper level optimizes the chosen unfairness criterion.
This perspective makes it possible to target a specific fairness metric without sacrificing efficiency, unlike fixed reweighting schemes or robust minimax objectives whose effect on a given metric can be difficult to control.
From an optimization standpoint, we develop scalable single-loop algorithms for solving this bilevel problem and establish convergence guarantees under mild regularity conditions, despite the lack of global smoothness induced by the coupling between weights and group losses.
We complement these theoretical results with an open-source toolbox, \tbadr, that implements our approach for multiple fairness metrics and standard learning models, and we demonstrate on real datasets that \frameworkname improves the targeted metric over Pareto-efficient baselines while preserving Pareto-efficient group performance.}
\yl{This framework opens several research directions at the intersection of multi-objective optimization, bilevel optimization, and statistical fairness. A particularly important direction concerns strengthening the statistical understanding of metric-driven selection over the Pareto front, especially the out-of-sample behavior of the selected fairness metric and the degree to which improvements observed on the empirical Pareto front translate into test performance.}

\bibliographystyle{icml}
\bibliography{references}
\clearpage
\tableofcontents
\clearpage

\appendix
\section{Fairness Metrics}\label{app:fairnessmetrics}

\begin{table}[t]
{\small\centering
\begin{tabular}{l l c c}
\toprule
\textbf{Reference} & \textbf{Fairness metric} & \textbf{Classification} & \textbf{Regression} \\
\midrule
\citep{metric_kearns} & Individual Fairness      & \cmark & \cmark \\
\citep{calders2009building} & Demographic Parity      & \cmark & \cmark \\
\citep{metric_zafar} & Disparate Mistreatment        & \cmark & \xmark \\
\citep{hardt2016equality} & Equal Opportunity       & \cmark & \xmark \\
\citep{hardt2016equality} & Equalized Odds          & \cmark & \xmark \\
\citep{gretton2005measuring} & Hilbert-Schmidt Independence Criterion & \xmark & \cmark \\
$\cdot$ & Group Variance        & \cmark & \cmark \\
\bottomrule
\end{tabular}
\caption{Task type handled by each fairness metric considered}
\label{tab:fairness-metrics}
}\end{table}

Our framework \frameworkname relies on differentiable unfairness metrics to enable its bilevel adaptive procedure.
In this section, \yl{we introduce and describe the unfairness metrics implemented and employed in our numerical experiments. These metrics, summarized in \cref{tab:fairness-metrics}, are defined as follows:}
%
%
\begin{itemize}
    \item[-] \emph{group variance} corresponds to the variance of the group losses, that is
        {\small \begin{displaymath}
        \fairnessmetric_\mathrm{GV}(\model)
        =
        \frac1\cS \sum_{a=1}^{\cS}
        \left(
        F_a(\model)-\frac1\cS\sum_{b=1}^{\cS}F_b(\model)
        \right)^2.
        \end{displaymath}}
    \item[-] \yl{\emph{individual fairness} penalizes models for how differently they treat $x_i$ and $x_j$ when these two samples come from different sensitive groups while having similar outcomes.} 
        We generalizes the definition for two groups proposed in \citep{metric_kearns} to multiple groups as :
        {\small 
        \begin{displaymath}
        \mathrm{IF} (\model)
        \defineq
        \frac{1}{\displaystyle \sum_{1\le a<b\le \cS} n_a n_b}
        \sum_{1\le a<b\le \cS}
        \sum_{i\,:\, a_i=a}
        \sum_{j\,:\, a_j=b}
        e^{-|y_i-y_j|}\bigl(f(\model,x_i)-f(\model,x_j)\bigr)^2
        \end{displaymath}}
    \item[-] \emph{demographic parity} \citep{calders2009building} requires the predictions of a machine learning model to be independent of sensitive attributes. It can be defined as $\fairnessmetric_\mathrm{DP} (\model) = \max_{a\in\attributespace} \left| p_a(\model) - \overline{p}(\model) \right|$, where $p_a(\model) = n_a^{-1} \sum_{i\,:\,a_i=a}\mathbf{1}\!\left\{ f(\model,x_i)\ge 0 \right\}$, and $\overline{p}(\model) = \cS^{-1} \sum_{a\in\attributespace} p_a(\model)$. To alleviate the non-smoothness of this metric, we implement the following differentiable approximation : 
        {\small \begin{displaymath}
            \fairnessmetric_\mathrm{DP}^{\rho} (\model)
            =
            \frac{1}{\rho}\log\!\left(
                \sum_{a\in\attributespace}
                \exp\!\Bigl(
                    \rho \,\sqrt{\bigl(p_a^{(\rho)}(\model)-\overline{p}^{(\rho)}(\model)\bigr)^2}
                \Bigr)
            \right)
        \end{displaymath}}
        where $p_a^{(\rho)}(\model)= n_a^{-1} \sum_{i\,:\,a_i=a}\sigma\!\bigl(\rho f(\model,x_i)\bigr),
        \quad
        \overline{p}^{(\rho)}(\model)= \cS^{-1} \sum_{a\in\attributespace} p_a^{(\rho)}(\model)$, and $\sigma$ denotes the sigmoid function.
    \item[-] \yl{\emph{disparate mistreatment} holds as a convex and differentiable variant of demographic parity introduced} in \citep{metric_zafar}. It is defined as
        {\small \begin{displaymath}
            \fairnessmetric_\mathrm{DM} (\model)
            =
            \Biggl(
                \frac{1}{n}\sum_{i=1}^n
                \bigl(a_i-\bar{a}\bigr)\bigl(f(\model,x_i)-\overline{f}(\model)\bigr)
            \Biggr)^2,
        \end{displaymath}}
    \item[-] \emph{equalized odds} and \emph{equal opportunities} are specific to classification tasks. 
        Equalized odds, defined as $\fairnessmetric_\mathrm{EOd} (\model) = \max_{a\in\attributespace}\mathrm{TPR}_a(\model) - \min_{a\in\attributespace}\mathrm{TPR}_a(\model) + \max_{a\in\attributespace}\mathrm{FPR}_a(\model) - \min_{a\in\attributespace}\mathrm{FPR}_a(\model)$ require both the true positive and false positive rates to be equal across groups \citep{hardt2016equality}. 
        Alternatively, equal opportunity writes $\fairnessmetric_\mathrm{EOp} (\model) = \max_{a\in\attributespace}\mathrm{TPR}_a(\model) - \min_{a\in\attributespace}\mathrm{TPR}_a(\model)$ and only requires true positive rates to be equal across groups.     
        To handle their non-differentiability, we define smoothed counterparts to True positive and False positive rates as
        {\small \begin{displaymath}
        \mathrm{TPR}_a^{\rho}(\model)
        \defineq
        \frac{1}{n_{a,1}}
        \sum_{i\,:\,a_i=a,\ y_i=1}
        \sigma\!\bigl(\rho\,f(\model,x_i)\bigr),
        \qquad
        \mathrm{FPR}_a^{\rho}(\model)
        \defineq
        \frac{1}{n_{a,0}}
        \sum_{i\,:\,a_i=a,\ y_i=0}
        \sigma\!\bigl(\rho\,f(\model,x_i)\bigr),
        \end{displaymath}}
        where $\rho > 0$ is a smoothing parameter, $n_{a,1}=\#\{i:\,a_i=a,\ y_i=1\}$ and $n_{a,0}=\#\{i:\,a_i=a,\ y_i=0\}$. 
        This leads to the differentiable approximations:
        {\small
        \begin{displaymath}
        \begin{split}
            \fairnessmetric_{\mathrm{EOp}}^{\rho}(\model)
                &\defineq
                    \frac{1}{\rho}\log\!\Bigl(
                        \sum_{a\in\attributespace}\exp\bigl(\rho\,\mathrm{TPR}_a^{\rho}(\model)\bigr)
                    \Bigr)
                    -
                    \frac{1}{\rho}\log\!\Bigl(
                        \sum_{a\in\attributespace}\exp\bigl(-\rho\,\mathrm{TPR}_a^{\rho}(\model)\bigr)
                    \Bigr), \\
            \fairnessmetric_{\mathrm{EOd}}^{\rho}(\model)
                &\defineq
                    \fairnessmetric_{\mathrm{EOp}}^{\rho}(\model)
                    +
                    \frac{1}{\rho}\log\!\Bigl(
                        \sum_{a\in\attributespace}\exp\bigl(\rho\,\mathrm{FPR}_a^{\rho}(\model)\bigr)
                    \Bigr)
                    -
                    \frac{1}{\rho}\log\!\Bigl(
                        \sum_{a\in\attributespace}\exp\bigl(-\rho\,\mathrm{FPR}_a^{\rho}(\model)\bigr)
                    \Bigr).
        \end{split}
        \end{displaymath}
        }
    \item[-] \emph{Hilbert-Schmidt Independence Criterion (HSIC)} \citep{gretton2005measuring} measures statistical dependence between two random variables via RKHS embeddings. In our experiments, we use a linear kernel and it reduces to the squared Hilbert-Schmidt norm of the empirical cross-covariance:
        \begin{displaymath}
            \fairnessmetric_\mathrm{HSIC}(\model)\defineq
            \left\|
            \frac{1}{n-1}\sum_{i=1}^n (s_i-\bar s)\bigl(f(\model,x_i)-\overline{f}(\model)\bigr)
            \right\|_2^2.
        \end{displaymath}
        where $\overline{s}\defineq \frac{1}{n}\sum_{i=1}^n s_i, \overline{f}(\model)\defineq \frac{1}{n}\sum_{i=1}^n f(\model,x_i), n\defineq \sum_{a\in\attributespace} n_a.$
\end{itemize}

\section{Convergence analysis}\label{app:convergence}
\yl{This section is dedicated to the proof of our main convergence results, namely \cref{thm:convergence_deterministic} and \cref{thm:convergence_stochastic}. 

\subsection{The deterministic case}\label{sec:proof_convergence}
In this section, we present the proof of Theorem~\ref{thm:convergence_deterministic}. 
%
%
We first derive in Section~\ref{sec:implicit_smoothness} some regularity properties on the implicit objective $\implicitobj : \uppervar \mapsto \upperobj(\lowersol(\uppervar), \uppervar)$.
Sections~\ref{sec:lower_descent} and~\ref{sec:dual_control} are dedicated to the control of the lower and dual updates, respectively.
These properties enable us to control the error on the approximate gradient of $\implicitobj$ in Section~\ref{sec:implicit_gradient_error}.
Finally, we combine all these results to prove Theorem~\ref{thm:convergence_deterministic} in Section~\ref{sec:proof_convergence_deterministic} using a properly parameterized Lyapunov function.


Throughout, $\uppervar_t, \lowervar_t, \dualvar_t$ denote the iterates generated by~\eqref{eq:badr_updates} at iteration $t \ge 0$. 
Optimality on the lower level is measured through the squared distance $\|\lowervar_t - \lowersol(\uppervar_t)\|^2$ to the optimal solution $\lowersol(\uppervar_t) \defineq \argmin_{\lowervar} \lowerobj(\lowervar, \uppervar_t)$, while optimality on the upperlevel is measured through the squared norm $\|\pgstep_{\uppervar, t}\|^2$ of the generalized gradient 

\begin{displaymath}
    \pgstep_{\uppervar, t} \defineq \frac{1}{\gamma}(\uppervar_t - \proj_{\cC}(\uppervar_t - \gamma \nabla \implicitobj(\uppervar_t))).
\end{displaymath}
We further introduce the approximate gradient $\agrad_{\uppervar, t}$ and the approximate generalized gradient $\apgstep_{\uppervar, t}$ as

\begin{displaymath}
\begin{aligned}
    \agrad_{\uppervar, t} &\defineq \nabla_{\uppervar} \upperobj(\lowervar_t, \uppervar_t) +  \nabla_{\lowervar, \uppervar}^2 \lowerobj(\lowervar_t, \uppervar_t)^\top \dualvar_t, \\
    \apgstep_{\uppervar, t} &\defineq \frac{1}{\gamma}(\uppervar_t - \proj_{\cC}(\uppervar_t - \gamma \agrad_{\uppervar, t})). 
\end{aligned}    
\end{displaymath}
Note that the upper level update in~\eqref{eq:badr_updates} can be rewritten as $\uppervar_{t+1} = \uppervar_t - \gamma \apgstep_{\uppervar, t} = \proj_{\cC}(\uppervar_t - \gamma \agrad_{\uppervar, t})$.
Finally, we also introduce the dual solutions $\dualsol(\uppervar_t)$ and $\dualsol(\uppervar_t, \lowervar_t)$ as 

\begin{displaymath}
\begin{aligned}
    \dualsol(\uppervar_t, \lowervar_t) &\defineq \nabla_{\lowervar, \lowervar}^2 \lowerobj(\lowervar_t, \uppervar_t)^{-1} \nabla_{\lowervar} \upperobj(\lowervar_t, \uppervar_t), \\
    \dualsol(\uppervar_t) & \defineq \sv{ \dualsol(\uppervar_t, \lowersol(\uppervar_t)) = \nabla_{\lowervar, \lowervar}^2 \lowerobj(\lowersol(\uppervar_t), \uppervar_t)^{-1} \nabla_{\lowervar} \upperobj(\lowersol(\uppervar_t), \uppervar_t).}\\    
\end{aligned}
\end{displaymath}

\subsubsection{Smoothness of the implicit map $\uppervar \mapsto \upperobj(\lowersol(\uppervar), \uppervar)$}\label{sec:implicit_smoothness}
In this paragraph, we derive regularity results on the solution map $\uppervar \mapsto \lowersol(\uppervar)$ as well as on the implicit objective $\uppervar \mapsto \upperobj(\lowersol(\uppervar), \uppervar)$.
We start by a Lipschitzness result on $\lowersol(\cdot)$. 
\begin{lemma}[\sv{Lipschitz-continuity of the solution map}]
\label{lem:lipschitz_implicit_solution}
    Under \cref{assumption:lowerobj}, the solution map $\uppervar \mapsto \lowersol(\uppervar)$ is $M_{\cC}/\mu_{\lowerobj}$-Lipschitz. 
    That is, for any $\uppervar_1, \uppervar_2 \in \mathbb{R}^{d_{\uppervar}}$, $\|\lowersol(\uppervar_1) - \lowersol(\uppervar_2)\| \le \frac{M_{\cC}}{\mu_{\lowerobj}} \|\uppervar_1 - \uppervar_2\|$.
\end{lemma}
\begin{proof}
    By strong convexity of $\lowerobj$ with respect to $\lowervar$, we have for any $\uppervar_1, \uppervar_2 \in \mathbb{R}^{d_{\uppervar}}$,
    {\small 
    \begin{align*}
        \frac{\mu_{\lowerobj}}{2}\|\lowersol(\uppervar_2)\!-\!\lowersol(\uppervar_1)\|^2
        &\le 
        \lowerobj(\lowersol(\uppervar_2),\uppervar_2)
        -\lowerobj(\lowersol(\uppervar_1),\uppervar_2)\\[-1mm]
        &\quad
        -\!\left\langle
        \nabla_{\lowervar}\lowerobj(\lowersol(\uppervar_1),\uppervar_2),
        \;\lowersol(\uppervar_2)\!-\!\lowersol(\uppervar_1)
        \right\rangle,\\[2pt]
        \frac{\mu_{\lowerobj}}{2}\|\lowersol(\uppervar_2)\!-\!\lowersol(\uppervar_1)\|^2
        &\le 
        \lowerobj(\lowersol(\uppervar_1),\uppervar_2)
        -\lowerobj(\lowersol(\uppervar_2),\uppervar_2)\\[-1mm]
        &\quad
        -\!\left\langle
        \nabla_{\lowervar}\lowerobj(\lowersol(\uppervar_2),\uppervar_2),
        \;\lowersol(\uppervar_1)\!-\!\lowersol(\uppervar_2)
        \right\rangle.
        \end{align*}}
        
    Hence, summing the two inequalities, we obtain
    {\small 
    \begin{align*}
        \mu_{\lowerobj} \left\|\lowersol(\uppervar_{2}) - \lowersol(\uppervar_{1})\right\|^{2}
            &\le \left \langle 
                \nabla_{\lowervar}\lowerobj\big(\lowersol(\uppervar_{2}),\uppervar_{2}\big)
                - \nabla_{\lowervar}\lowerobj\big(\lowersol(\uppervar_{1}),\uppervar_{2}\big) , 
                \lowersol(\uppervar_{2}) - \lowersol(\uppervar_{1})
                \right \rangle. \\
            &= \left \langle 
                \nabla_{\lowervar}\lowerobj\big(\lowersol(\uppervar_{1}),\uppervar_{2}\big) 
                        ,\lowersol(\uppervar_{1}) - \lowersol(\uppervar_{2})
                \right \rangle. 
    \end{align*}   } 
    where the equality follows from the optimality condition $\nabla_{\lowervar}\lowerobj\big(\lowersol(\uppervar_{2}),\uppervar_{2}\big) = 0$.
    Moreover, using \cref{assumption:lowerobj}.4, we have 
    {\small\begin{displaymath}
        \| \nabla_{\lowervar}\lowerobj\big(\lowersol(\uppervar_{1}),\uppervar_{2}\big) \| = \| \nabla_{\lowervar}\lowerobj\big(\lowersol(\uppervar_{1}),\uppervar_{2}\big) - \nabla_{\lowervar}\lowerobj\big(\lowersol(\uppervar_{2}),\uppervar_{2}\big) \| \le M_{\cC} \|\uppervar_1 - \uppervar_2\|.
    \end{displaymath}}
    and by Cauchy-Schwarz inequality, we obtain
    {\small\begin{displaymath}
        \mu_{\lowerobj} \left\|\lowersol(\uppervar_{2}) - \lowersol(\uppervar_{1})\right\|^{2}
            \le M_{\cC} \|\uppervar_1 - \uppervar_2\| \cdot \left\|\lowersol(\uppervar_{2}) - \lowersol(\uppervar_{1})\right\|.
    \end{displaymath}}
    Hence, $\left\|\lowersol(\uppervar_{2}) - \lowersol(\uppervar_{1})\right\| \le \frac{M_{\cC}}{\mu_{\lowerobj}} \|\uppervar_1 - \uppervar_2\|$.
\end{proof}
Let us now show the smoothness of the implicit objective $\implicitobj$.
\begin{lemma}[\sv{Smoothness of the implicit objective}]\label{lem:smoothness_h}
    The implicit objective $\implicitobj : \uppervar \mapsto \upperobj(\lowersol(\uppervar), \uppervar)$ is $L_{\implicitobj}$-smooth with
    {\small 
    \begin{displaymath}
        L_{\implicitobj} \defineq \left(L_{\upperobj, 1} \left(\frac{M_\cC}{\mu_{\lowerobj}} + 1\right) + L_{\upperobj, 0} 
        \left( \frac{L_{\lowerobj, 2}^{\lowervar, \lowervar} M_{\cC}}{\mu_{\lowerobj}^2} + \frac{L_{\lowerobj, 2}^{\lowervar, \uppervar}}{\mu_{\lowerobj}}\right)\right) \left(\frac{M_\cC}{\mu_{\lowerobj}} + 1\right).
    \end{displaymath}}
\end{lemma}
\begin{proof}
    Let us first establish the smoothness of $\lowersol(\cdot)$. 
    \sv{Using the strong convexity of $\lowerobj$ with respect to $\lowervar$, and in turn the invertibility of the Hessian $\nabla_{\lowervar, \uppervar}^2 \lowerobj$, the implicit function theorem guarantees} differentiability of $\lowersol(\cdot)$ with for all $\uppervar \in \cC$,
    {\small\begin{displaymath}
        \Jac \lowersol(\uppervar) = - \nabla_{\lowervar, \lowervar}^2 \lowerobj(\lowersol(\uppervar), \uppervar)^{-1} \nabla_{\lowervar, \uppervar}^2 \lowerobj(\lowersol(\uppervar), \uppervar).
    \end{displaymath}}
    Hence, for any $\uppervar_1, \uppervar_2 \in \cC$, we have :
    {\small \begin{align*}
        \|\Jac\lowersol(\uppervar_1)\!-\!\Jac\lowersol(\uppervar_2)\|
        &\le
        \big\|\nabla^2_{\lowervar\lowervar}\lowerobj(\lowersol(\uppervar_1),\uppervar_1)^{-1}
              -\nabla^2_{\lowervar\lowervar}\lowerobj(\lowersol(\uppervar_2),\uppervar_2)^{-1}\big\| \cdot\,
        \big\|\nabla^2_{\lowervar\uppervar}\lowerobj(\lowersol(\uppervar_1),\uppervar_1)\big\|  \\[2pt]
        &\quad+
        \big\|\nabla^2_{\lowervar\lowervar}\lowerobj(\lowersol(\uppervar_2),\uppervar_2)^{-1}\big\| \cdot\,
        \big\|\nabla^2_{\lowervar\uppervar}\lowerobj(\lowersol(\uppervar_1),\uppervar_1)
              -\nabla^2_{\lowervar\uppervar}\lowerobj(\lowersol(\uppervar_2),\uppervar_2)\big\|.
    \end{align*}}
    By \cref{assumption:lowerobj}.4, we have $\|\nabla_{\lowervar, \uppervar}^2 \lowerobj(\lowersol(\uppervar_1), \uppervar_1)\| \le M_{\cC}$.
    Moreover, using \cref{assumption:lowerobj}.1, we have the upperbound $\|\nabla_{\lowervar, \lowervar}^2 \lowerobj(\lowersol(\uppervar_2), \uppervar_2)^{-1}\| \le 1/\mu_{\lowerobj}$. Thus, combined with the matrix identity $A^{-1} - B^{-1} = A^{-1}(B - A)B^{-1}$, we obtain
    \begin{small}
    \begin{align*}
        \Bigl\|
          \nabla_{\lowervar,\lowervar}^2\lowerobj(\lowersol(\uppervar_1),\uppervar_1)^{-1}
          -\nabla_{\lowervar,\lowervar}^2\lowerobj(\lowersol(\uppervar_2),\uppervar_2)^{-1}
        \Bigr\|
        &\le
        \Bigl\|
          \nabla_{\lowervar,\lowervar}^2\lowerobj(\lowersol(\uppervar_1),\uppervar_1)^{-1}
        \Bigr\|
        \cdot
        \Bigl\|
          \nabla_{\lowervar,\lowervar}^2\lowerobj(\lowersol(\uppervar_2),\uppervar_2)^{-1}
        \Bigr\|
        \\
        &\cdot
        \Bigl\|
          \nabla_{\lowervar,\lowervar}^2\lowerobj(\lowersol(\uppervar_2),\uppervar_2)
          \!-\!\nabla_{\lowervar,\lowervar}^2\lowerobj(\lowersol(\uppervar_1),\uppervar_1)
        \Bigr\|
        \\
        &\le
        \mu_{\lowerobj}^{-2}
        \Bigl\|
          \nabla_{\lowervar,\lowervar}^2\lowerobj(\lowersol(\uppervar_2),\uppervar_2)
          \!-\!\nabla_{\lowervar,\lowervar}^2\lowerobj(\lowersol(\uppervar_1),\uppervar_1)
        \Bigr\|
        \\
        &\le
        L_{\lowerobj,2}^{\lowervar,\lowervar}\mu_{\lowerobj}^{-2}
        \Bigl(
          \|\lowersol(\uppervar_1)-\lowersol(\uppervar_2)\|
          +\|\uppervar_1-\uppervar_2\|
        \Bigr)
    \end{align*}
    \end{small}
    where the last inequality follows from \cref{assumption:lowerobj}.3.
    Finally, using Lemma~\ref{lem:lipschitz_implicit_solution}, we obtain
    {\small
        \begin{equation}\label{eq:bound_jacobian_lowersol_1}
        \Big\| \nabla_{\lowervar, \lowervar}^2 \lowerobj(\lowersol(\uppervar_1), \uppervar_1)^{-1}\!-\!\nabla_{\lowervar, \lowervar}^2 \lowerobj(\lowersol(\uppervar_2), \uppervar_2)^{-1} \Big\| \cdot \Big\| \nabla_{\lowervar, \uppervar}^2 \lowerobj(\lowersol(\uppervar_1), \uppervar_1) \Big\| \le \frac{L_{\lowerobj, 2}^{\lowervar, \lowervar} M_{\cC}}{\mu_{\lowerobj}^2} \Big( \frac{M_{\cC}}{\mu_{\lowerobj}} \!+\! 1 \Big) \|\uppervar_1 \!-\! \uppervar_2\|.            
    \end{equation}
    }
    Furthermore, combining \cref{assumption:lowerobj}.1 and \cref{assumption:lowerobj}.5 yields
    {\small
    \begin{equation*}
    \|\nabla_{\lowervar,\lowervar}^2\lowerobj(\lowersol(\uppervar_2),\uppervar_2)^{-1}\|\,
    \|\nabla_{\lowervar,\uppervar}^2\lowerobj(\lowersol(\uppervar_1),\uppervar_1)
    -\nabla_{\lowervar,\uppervar}^2\lowerobj(\lowersol(\uppervar_2),\uppervar_2)\|
    \le
    L_{\lowerobj,2}^{\lowervar,\uppervar}\mu_{\lowerobj}^{-1}
    \begin{aligned}[t]
    &\bigl(\|\lowersol(\uppervar_1)-\lowersol(\uppervar_2)\|\\
    &\qquad+\|\uppervar_1-\uppervar_2\|\bigr)
    \end{aligned}
    \end{equation*}
    }
    and thus, using again Lemma~\ref{lem:lipschitz_implicit_solution}, we obtain
    {\small
    \begin{equation}\label{eq:bound_jacobian_lowersol_2}
    \|\nabla_{\lowervar,\lowervar}^2\lowerobj(\lowersol(\uppervar_2),\uppervar_2)^{-1}\|\,
    \|\nabla_{\lowervar,\uppervar}^2\lowerobj(\lowersol(\uppervar_1),\uppervar_1)
    -\nabla_{\lowervar,\uppervar}^2\lowerobj(\lowersol(\uppervar_2),\uppervar_2)\|
    \le
    L_{\lowerobj,2}^{\lowervar,\uppervar}\mu_{\lowerobj}^{-1}
    \bigl(M_{\cC}\mu_{\lowerobj}^{-1}+1\bigr)\|\uppervar_1-\uppervar_2\|
    \end{equation}
    }
    Summing Equations~\eqref{eq:bound_jacobian_lowersol_1} and~\eqref{eq:bound_jacobian_lowersol_2}, we obtain that $\Jac \lowersol(\cdot)$ is Lipschitz with constant
    {\small
    \begin{equation}\label{eq:smoothness_lowersol}
        \|\Jac \lowersol(\uppervar_1) - \Jac \lowersol(\uppervar_2)\| \le \Big( \frac{L_{\lowerobj, 2}^{\lowervar, \lowervar} M_{\cC}}{\mu_{\lowerobj}^2} + \frac{L_{\lowerobj, 2}^{\lowervar, \uppervar}}{\mu_{\lowerobj}} \Big) \Big( \frac{M_{\cC}}{\mu_{\lowerobj}} + 1 \Big) \|\uppervar_1 - \uppervar_2\|.
    \end{equation}
    }
    By \cref{assumption:upperobj}, $\implicitobj$ is differentiable, and the chain rule gives for any $\uppervar \in \cC$,
    {\small
    \begin{displaymath}
        \nabla \implicitobj(\uppervar) = \nabla_{\uppervar} \upperobj(\lowersol(\uppervar), \uppervar) + \Jac \lowersol(\uppervar)^\top \nabla_{\lowervar} \upperobj(\lowersol(\uppervar), \uppervar).
    \end{displaymath} }
    Hence, for any $\uppervar_1, \uppervar_2 \in \cC$, we have
    {\small 
        \begin{align*}
        \|\nabla \implicitobj(\uppervar_1) - \nabla \implicitobj(\uppervar_2)\| 
        &\le \|\nabla_{\uppervar} \upperobj(\lowersol(\uppervar_1), \uppervar_1) - \nabla_{\uppervar} \upperobj(\lowersol(\uppervar_2), \uppervar_2)\| \\
        &\quad + \|\Jac \lowersol(\uppervar_1)^\top - \Jac \lowersol(\uppervar_2)^\top\| \cdot \|\nabla_{\lowervar} \upperobj(\lowersol(\uppervar_1), \uppervar_1)\| \\
        &\quad + \|\Jac \lowersol(\uppervar_2)^\top\| \cdot \|\nabla_{\lowervar} \upperobj(\lowersol(\uppervar_1), \uppervar_1) - \nabla_{\lowervar} \upperobj(\lowersol(\uppervar_2), \uppervar_2)\|.
    \end{align*}}
    Combining \cref{assumption:upperobj}.2 and Lemma~\ref{lem:lipschitz_implicit_solution}, we have 
    {\small \begin{align} \label{eq:smoothness_h_part1}
        \|\nabla_{\uppervar} \upperobj(\lowersol(\uppervar_1), \uppervar_1) - \nabla_{\uppervar} \upperobj(\lowersol(\uppervar_2), \uppervar_2)\| 
            &\le L_{\upperobj, 1} \left(\|\uppervar_1 - \uppervar_2\| + \|\lowersol(\uppervar_1) - \lowersol(\uppervar_2)\| \right) \notag\\
            &\le L_{\upperobj, 1} \left(\frac{M_\cC}{\mu_{\lowerobj}} + 1 \right) \|\uppervar_1 - \uppervar_2\| 
    \end{align}}
    Besides, using~\eqref{eq:smoothness_lowersol} and \cref{assumption:upperobj}.1, we have
    {\small \begin{align}\label{eq:smoothness_h_part2}
    \begin{aligned}
    \|\Jac\lowersol(\uppervar_1)^\top-\Jac\lowersol(\uppervar_2)^\top\|\,
    &\|\nabla_{\lowervar}\upperobj(\lowersol(\uppervar_1),\uppervar_1)\|
    \\[-0.2em]
    &\le L_{\upperobj,0}
    \Big(
      \frac{L_{\lowerobj,2}^{\lowervar,\lowervar}M_{\cC}}{\mu_{\lowerobj}^2}
      +\frac{L_{\lowerobj,2}^{\lowervar,\uppervar}}{\mu_{\lowerobj}}
    \Big)\Big(\frac{M_{\cC}}{\mu_{\lowerobj}}+1\Big)\|\uppervar_1-\uppervar_2\|
    \end{aligned}
    \end{align}}
    Finally, using \cref{assumption:upperobj}.2 and Lemma~\ref{lem:lipschitz_implicit_solution}, we have
    {\small \begin{align} \label{eq:smoothness_h_part3}
        \|\Jac \lowersol(\uppervar_2)^\top\| \cdot \|\nabla_{\lowervar} \upperobj(\lowersol(\uppervar_1), \uppervar_1) - \nabla_{\lowervar} \upperobj(\lowersol(\uppervar_2), \uppervar_2)\|   
            &\le \frac{M_{\cC}L_{\upperobj, 1}}{\mu_{\lowerobj}} \left(\frac{M_{\cC}}{\mu_{\lowerobj}} + 1\right) \|\uppervar_1 - \uppervar_2\|. 
    \end{align}}
    Summing Equations~\eqref{eq:smoothness_h_part1},~\eqref{eq:smoothness_h_part2} and~\eqref{eq:smoothness_h_part3}, we obtain the desired result. 
\end{proof}
We finally derive a descent lemma for the implicit objective $\implicitobj$ along the iterates $\uppervar_t$.
This lemma will be instrumental in the derivation of our Lyapunov analysis in Section~\ref{sec:proof_convergence_deterministic}.
\begin{lemma}[\sv{Descent lemma for the implicit objective}]\label{lem:implicit_descent}
    For any $t \ge 0$, we have
    {\small \begin{displaymath}
    \implicitobj(\uppervar_{t+1}) \le \implicitobj(\uppervar_t)
    \!-\!\frac{\gamma}{8}\|\pgstep_{\uppervar,t}\|^2
    \!-\!\frac{\gamma}{4}\|\apgstep_{\uppervar,t}\|^2
    \!+\!\frac{\gamma}{4}\|\pgstep_{\uppervar,t}-\apgstep_{\uppervar,t}\|^2
    \!+\!\frac{\gamma}{2}\|\nabla\implicitobj(\uppervar_t)-\agrad_{\uppervar,t}\|^2
    \!+\!\frac{\gamma^2L_{\implicitobj}}{2}\|\apgstep_{\uppervar,t}\|^2.
    \end{displaymath}}
\end{lemma}
\begin{proof}
By $L_{\implicitobj}$-smoothness of \sv{$\implicitobj$}, we have for any $t \ge 0$,
{\small \begin{align}\label{eq:descent_h_1}
\implicitobj(\uppervar_{t+1}) 
    &\le \implicitobj(\uppervar_t) + \langle \nabla \implicitobj(\uppervar_t), \uppervar_{t+1} - \uppervar_t \rangle + \frac{L_{\implicitobj}}{2} \|\uppervar_{t+1} - \uppervar_t\|^2 \notag\\
    &= \implicitobj(\uppervar_t) - \gamma \langle \nabla \implicitobj(\uppervar_t), \apgstep_{\uppervar, t} \rangle + \frac{\gamma^2 L_{\implicitobj}}{2} \|\apgstep_{\uppervar, t}\|^2 
\end{align}}
Now, by convexity of $\cC$, since $\uppervar_{t+1} = \proj_{\cC}(\uppervar_t - \gamma \agrad_{\uppervar, t})$, we have $\langle \uppervar_t - \gamma \agrad_{\uppervar, t} - \uppervar_{t+1}, \uppervar_t - \uppervar_{t+1}\rangle \le 0$,
i.e.,
{\small \begin{displaymath}
    \langle \agrad_{\uppervar, t}, \uppervar_{t+1} - \uppervar_t \rangle \le -\frac{1}{\gamma} \|\uppervar_{t+1} - \uppervar_t\|^2 = -\gamma \|\apgstep_{\uppervar, t}\|^2.
\end{displaymath}}
Hence, using the inequality $\langle a, b \rangle \le \frac{1}{2}(\|a\|^2 + \|b\|^2)$, we have
{\small \begin{align*}
    - \gamma \langle \nabla \implicitobj(\uppervar_t), \apgstep_{\uppervar, t} \rangle 
    &= - \gamma \langle \agrad_{\uppervar, t}, \apgstep_{\uppervar, t} \rangle + \gamma \langle \agrad_{\uppervar, t} - \nabla \implicitobj(\uppervar_t), \apgstep_{\uppervar, t} \rangle \\    
    &\le - \frac{\gamma}{2} \|\apgstep_{\uppervar, t}\|^2 + \frac{\gamma}{2} \| \agrad_{\uppervar, t} - \nabla \implicitobj(\uppervar_t)\|^2 
\end{align*}}
Plugging this inequality back \sv{into the $L_{\implicitobj}$-smoothness of $\implicitobj$}~\eqref{eq:descent_h_1}, we obtain
{\small \begin{align*}
\implicitobj(\uppervar_{t+1}) 
    &\le \implicitobj(\uppervar_t) - \frac{\gamma}{2} \|\apgstep_{\uppervar, t}\|^2 + \frac{\gamma}{2} \| \agrad_{\uppervar, t} - \nabla \implicitobj(\uppervar_t)\|^2 + \frac{\gamma^2 L_{\implicitobj}}{2} \|\apgstep_{\uppervar, t}\|^2 \\
    &\le \implicitobj(\uppervar_t) 
            - \frac{\gamma}{4} \|\apgstep_{\uppervar, t}\|^2
            - \frac{\gamma}{8} \|\pgstep_{\uppervar, t}\|^2            
            + \frac{\gamma}{4} \|\pgstep_{\uppervar, t} - \apgstep_{\uppervar, t}\|^2 + \frac{\gamma}{2} \|\nabla \implicitobj(\uppervar_t) - \agrad_{\uppervar, t}\|^2
            + \frac{\gamma^2 L_{\implicitobj}}{2} \|\apgstep_{\uppervar, t}\|^2.
\end{align*}}
where the last inequality follows from the inequality $-\|a\|^2 \le -\frac{1}{2}\|b\|^2 + \|a - b\|^2$.
\end{proof}
\subsubsection{Lower level descent}\label{sec:lower_descent}
We move now to descent properties on the lower variable $\lowervar_t$.
We start with a descent lemma property that is shared with the dual variable $\dualvar_t$ in Section~\ref{sec:dual_control}.
This stems from both updates -- for $\lowervar_t$ and $\dualvar_t$ being gradient steps on strongly convex and smooth objectives.
\begin{lemma}[\sv{Linear convergence of the lower and dual variables}]\label{lem:lower-level-descent}    
    \sv{Assume that the lower and dual stepsizes satisfiy} $\tau, \rho \le \frac{1}{L_{\lowerobj,1}^{\lowervar, \lowervar}}$.  
    Then, for any $t \ge 0$, we have
    {\small \begin{align*}
        \|\lowervar_{t+1} - \lowersol_t\|^2 &\le (1 - \tau \mu_{\lowerobj}) \|\lowervar_t - \lowersol_t\|^2. \\
        \|\dualvar_{t+1} - \dualsol(\lowervar_t, \uppervar_t)\|^2 &\le (1 - \rho \mu_{\lowerobj}) \|\dualvar_t - \dualsol(\lowervar_t, \uppervar_t)\|^2
    \end{align*}}
\end{lemma}
\begin{proof}
For any $t \ge 0$,  
{\small \begin{align*}
\|\lowervar_{t+1} - \lowersol_t\|^2 
&= \|\lowervar_t - \tau \nabla_\lowervar \lowerobj(\lowervar_t, \uppervar_t) - \lowersol_t\|^2 \\
&= \|\lowervar_t - \lowersol_t\|^2 - 2\tau \langle \lowervar_t - \lowersol_t, \nabla_\lowervar \lowerobj(\lowervar_t, \uppervar_t)\rangle + \tau^2 \|\nabla_\lowervar \lowerobj(\lowervar_t, \uppervar_t)\|^2 \\
&\quad + \tau^2 \|\nabla_\lowervar \lowerobj(\lowervar_t, \uppervar_t)\|^2. 
\end{align*}}
By strong convexity of $\lowerobj$ in $\lowervar$, $- \tau \langle \lowervar_t - \lowersol_t, \nabla_\lowervar \lowerobj(\lowervar_t, \uppervar_t)\rangle \le -\tau \mu_{\lowerobj} \|\lowervar_t - \lowersol_t\|^2$.
%
Furthermore, by co-coercivity of $\nabla_\lowervar \lowerobj$,  
{\small \begin{displaymath}
- \tau \langle \lowervar_t - \lowersol_t, \nabla_\lowervar \lowerobj(\lowervar_t, \uppervar_t)\rangle + \tau^2 \|\nabla_\lowervar \lowerobj(\lowervar_t, \uppervar_t)\|^2 
\le - \tau \Big(\frac{1}{L_{\lowerobj,1}^{\lowervar, \lowervar}} - \tau\Big)\|\nabla_\lowervar \lowerobj(\lowervar_t, \uppervar_t)\|^2.
\end{displaymath}}
Thus,
\begin{displaymath}
\|\lowervar_{t+1} - \lowersol_t\|^2 \le (1 - \tau \mu_{\lowerobj})\|\lowervar_t - \lowersol_t\|^2 
- \tau \Big(\frac{1}{L_{\lowerobj,1}^{\lowervar, \lowervar}} - \tau\Big)\|\nabla_\lowervar \lowerobj(\lowervar_t, \uppervar_t)\|^2.
\end{displaymath}
Finally, assuming $\tau \le \frac{1}{L_{\lowerobj,1}^{\lowervar, \lowervar}}$, the last term is nonpositive, so
\begin{displaymath}
\|\lowervar_{t+1} - \lowersol_t\|^2 \le (1 - \tau \mu_{\lowerobj})\|\lowervar_t - \lowersol_t\|^2.
\end{displaymath}
Finally, observe that the function $\dualvar \mapsto \frac{1}{2} \dualvar^\top \nabla_{\lowervar, \lowervar}^{2} \lowerobj(\lowervar_t, \uppervar_t) \dualvar + \dualvar^\top \nabla_{\lowervar} \upperobj(\lowervar_t, \uppervar_t)$ is $L_{\lowerobj, 1}^{\lowervar, \lowervar}$-smooth and $\mu_{\lowerobj}$-strongly convex by Assumptions~\ref{assumption:lowerobj}.2 and ~\ref{assumption:lowerobj}.3.
Thus, noting that the update on $v_t$ can be interpreted as a gradient step on this function and using the same rationale as above leads to     
{\small \begin{equation}\label{eq:dual_descent}
    \|\dualvar_{t+1} - \dualsol(\lowervar_t, \uppervar_t)\|^2 \le (1 - \rho \mu_{\lowerobj}) \|\dualvar_t - \dualsol(\lowervar_t, \uppervar_t)\|^2,
\end{equation}}
for all $t \ge 0$.
\end{proof}
We move to a control of the drift of the lower variable $\lowervar_t$ with respect to the implicit solution $\lowersol_t$.
\begin{lemma}[\sv{Drift control of the lower variable}]\label{lem:lower_level_drift}
    Assume $\tau \le \tfrac{1}{L_{\lowerobj,1}^{\lowervar, \lowervar}}$.  
    Then, for any $t \ge 0$, we have  
    {\small 
    \begin{displaymath}
    \|\lowervar_{t+1} - \lowersol_{t+1}\|^2 
    \;\le\; 
    \Bigl(1 - \tfrac{\tau \mu_{\lowerobj}}{2}\Bigr) \|\lowervar_t - \lowersol_t\|^2
    \;+\; \tfrac{2}{\tau \mu_{\lowerobj}} \Bigl(\tfrac{M_{\cC}}{\mu_{\lowerobj}}\Bigr)^2 
    \gamma^2 \,\|\apgstep_{\uppervar, t}\|^2 .
    \end{displaymath}}
\end{lemma}
\begin{proof}
    For any $t \in \mathbb{N}$, using Lemma~\ref{lem:lower-level-descent} and Lemma~\ref{lem:lipschitz_implicit_solution}, we have :  
    {\small \begin{align*}
    \|\lowervar_{t+1} - \lowersol_{t+1}\|^2 
    &= \|\lowervar_{t+1} - \lowersol_t + \lowersol_t - \lowersol_{t+1}\|^2 \\[0.5em]
    &= \|\lowervar_{t+1} - \lowersol_t\|^2 
    + 2\langle \lowervar_{t+1} - \lowersol_t, \lowersol_t - \lowersol_{t+1}\rangle 
    + \|\lowersol_t - \lowersol_{t+1}\|^2 \\[0.5em]
    &\le \Bigl(1 - \tau \mu_{\lowerobj}\Bigr)\|\lowervar_t - \lowersol_t\|^2 
    + \Bigl(\tfrac{M_{\cC}}{\mu_{\lowerobj}}\Bigr)^2 \gamma^2 \|\apgstep_{\uppervar,t}\|^2 
    + 2\langle \lowervar_{t+1} - \lowersol_t, \lowersol_t - \lowersol_{t+1}\rangle .
    \end{align*}}
    Using Young's inequality, and Lemma~\ref{lem:lower-level-descent}, we obtain for any $\eta > 0$ :
    {\small \begin{align*}
        \langle \lowervar_{t+1} - \lowersol_t, \lowersol_t - \lowersol_{t+1}\rangle 
            &\le \frac{\eta}{2} \|\lowervar_{t+1} - \lowersol_t\|^2 + \frac{1}{2\eta} \|\lowersol_t - \lowersol_{t+1}\|^2 \\
            &\le \frac{\eta}{2} (1 - \tau \mu_{\lowerobj}) \|\lowervar_t - \lowersol_t\|^2 + \frac{1}{2\eta} \Bigl(\tfrac{M_{\cC}}{\mu_{\lowerobj}}\Bigr)^2 \gamma^2 \|\apgstep_{\uppervar,t}\|^2.
    \end{align*}}
    Taking $\eta = \frac{\tau \mu_{\lowerobj}}{2(1-\tau\mu_{\lowerobj})}$ finally gives
    {\small \begin{align*}
    \|\lowervar_{t+1} - \lowersol_{t+1}\|^2 
    &\le \Bigl(1 - \frac{\tau \mu_{\lowerobj}}{2}\Bigr)\|\lowervar_t - \lowersol_t \| + \Bigl( \frac{2}{\tau \mu_{\lowerobj}} - 1\Bigr) \Bigl(\tfrac{M_{\cC}}{\mu_{\lowerobj}}\Bigr)^2 \gamma^2 \|\apgstep_{\uppervar, t}\|^2\\
    &\le \Bigl(1 - \frac{\tau \mu_{\lowerobj}}{2}\Bigr)\|\lowervar_t - \lowersol_t \| + \frac{2}{\tau \mu_{\lowerobj}} \Bigl(\tfrac{M_{\cC}}{\mu_{\lowerobj}}\Bigr)^2 \gamma^2 \|\apgstep_{\uppervar, t}\|^2.
    \end{align*}}
\end{proof}
\subsubsection{Dual control}\label{sec:dual_control}
In this paragraph, we derive several bounds on the dual iterates $\dualvar_t$.
We start with a boundedness result on the sequence $(\dualvar_t)_{t \ge 0}$. 
This result is crucial for the control of the error made on the upper-level gradient estimate in Section~\ref{sec:implicit_gradient_error}, as it allows to circumvent the issue of having a potentially unbounded Hessian $\nabla_{\lowervar, \uppervar}^2 \lowerobj$ for the lower level objective.
In particular, prior analyses~\citep{ghadimi2018approximation,dagreou2022framework} do not address these issues.
\begin{lemma}[\sv{Boundness of the dual variable}]\label{lem:dual_bounded}
    \sv{Assume that the dual stepsize satisfies} $\rho \le \tfrac{1}{L_{\lowerobj, 1}^{\lowervar, \lowervar}}$.
    Then, the sequence $(\dualvar_t)_{t \ge 0}$ satisfies for all $t \ge 0$,
    {\small \begin{displaymath}
        \|\dualvar_{t}\| \le \frac{4 L_{\upperobj, 0}}{\rho \mu_{\lowerobj}^2} + \starterrordual,
    \end{displaymath}}
    where $\starterrordual \defineq \|\dualvar_0 - \dualsol(\lowervar_0, \uppervar_0)\|^2$.
\end{lemma}
\begin{proof}    
    Let us now show by induction that for all $t \ge 0$,
    {\small \begin{displaymath}
        \|\dualvar_t\| \le \frac{L_{\upperobj, 0}}{\mu_{\lowerobj}} + \left[ \sum_{s=1}^{t-1} (1-\rho \mu_{\lowerobj})^{s/2} \cdot \frac{2 L_{\upperobj, 0}}{\mu_{\lowerobj}} \right] + (1 - \rho \mu_{\lowerobj})^{t/2} \starterrordual.
    \end{displaymath}}
    First note that for all $t\ge 0$, $\| \dualsol(\lowervar_t, \uppervar_t) \| = \| \nabla_{\lowervar, \lowervar}^2 \lowerobj(\lowervar_t, \uppervar_t)^{-1} \nabla \upperobj(\lowervar_t, \uppervar_t) \| \le \frac{L_{\upperobj, 0}}{\mu_{\lowerobj}}$.
    Thus, for $t = 0$, we obtain,
    \begin{displaymath}
        \|\dualvar_1\| \le \|\dualvar_1 - \dualsol(\lowervar_0, \uppervar_0)\| + \|\dualsol(\lowervar_0, \uppervar_0)\| \le \starterrordual + \frac{L_{\upperobj, 0}}{\mu_{\lowerobj}},
    \end{displaymath}
    which confirms the base case (using the convention that empty sums are null).
    Assume now the result holds for some $t \ge 0$. Then, we have by Lemma~\ref{lem:lower-level-descent}
    {\small \begin{align*}
        \| \dualvar_{t+1} \| &\le \| \dualvar_{t+1} - \dualsol(\lowervar_{t}, \uppervar_{t})\| + \| \dualsol(\lowervar_{t}, \uppervar_{t})\| \\
                         &\le (1-\rho \mu_{\lowerobj})^{1/2} \| \dualvar_{t} - \dualsol(\lowervar_{t}, \uppervar_{t})\| + \| \dualsol(\lowervar_{t}, \uppervar_{t})\| \\
                         &\le (1-\rho \mu_{\lowerobj})^{1/2} \|\dualvar_{t}\| + (1-\rho \mu_{\lowerobj})^{1/2} \frac{L_{\upperobj, 0}}{\mu_{\lowerobj}} + \frac{L_{\upperobj, 0}}{\mu_{\lowerobj}} \\
                         &\le (1-\rho \mu_{\lowerobj})^{1/2} \begin{aligned}[t]
                                                            & \left( \frac{L_{\upperobj, 0}}{\mu_{\lowerobj}} + \sum_{s=1}^{t-1} (1-\rho \mu_{\lowerobj})^{s/2} \cdot \frac{2 L_{\upperobj, 0}}{\mu_{\lowerobj}} + (1 - \rho \mu_{\lowerobj})^{t/2} \starterrordual\right)\\
                                                            & + (1-\rho \mu_{\lowerobj})^{1/2} \frac{L_{\upperobj, 0}}{\mu_{\lowerobj}} + \frac{L_{\upperobj, 0}}{\mu_{\lowerobj}} \\
                                                        \end{aligned} \\
                         &= \frac{L_{\upperobj, 0}}{\mu_{\lowerobj}} + \sum_{s=1}^{t} (1-\rho \mu_{\lowerobj})^{s/2} \cdot \frac{2 L_{\upperobj, 0}}{\mu_{\lowerobj}} + (1 - \rho \mu_{\lowerobj})^{(t+1)/2} \starterrordual,                                                                                                                                           
    \end{align*}}
    This finishes the induction and we can conclude that for all $t \ge 0$,
    {\small \begin{displaymath}
        \|\dualvar_t\| \le \frac{2 L_{\upperobj, 0}}{\mu_{\lowerobj}} \sum_{t=0}^{\infty} (1-\rho \mu_{\lowerobj})^{t/2} + \starterrordual = \frac{4 L_{\upperobj, 0}}{\rho \mu_{\lowerobj}^2} + \starterrordual.
    \end{displaymath}}
    where we used the inequality $\sum_{t=0}^\infty (1-\alpha)^{t/2} \le \tfrac{2}{\alpha}$ for all $\alpha \in (0,1]$.
\end{proof}
We now move to a bound on the drift of the dual solution $\dualsol(\lowervar_t, \uppervar_t)$ along the iterates, which will be useful to control the dual error $\|\dualvar_t - \dualsol(\lowervar_t, \uppervar_t)\|$ in the upcoming Lemma~\ref{lem:dualsol_distance}.
\begin{lemma}[\sv{Drift of the dual solution}]\label{lem:dualsol_update}
    For any $t \ge 0$, we have 
    {\small \begin{displaymath}
        \|\dualsol(\lowervar_{t+1}, \uppervar_{t+1}) -  \dualsol(\lowervar_{t}, \uppervar_{t})\| 
            \le \constantdualdrift
                 \left(
                    \tau L_{\lowerobj, 1}^{\lowervar, \lowervar}  \|\lowervar_t - \lowersol_t\| 
                    + \gamma \|\apgstep_{\uppervar, t}\|
                \right)
    \end{displaymath}}
    where $\constantdualdrift \defineq \Bigl( L_{\upperobj, 0} \frac{L_{\lowerobj, 2}^{\lowervar, \lowervar}}{\mu_{\lowerobj}^2} + \frac{L_{\upperobj, 1}}{\mu_{\lowerobj}} \Bigr)$.
\end{lemma}
\begin{proof}
Observe that, 
\begin{small}\begin{align*}
    \|\dualsol(\lowervar_{t+1}, \uppervar_{t+1}) -  \dualsol(\lowervar_{t}, \uppervar_{t}) \| 
            &\le 
                \|\nabla_{\lowervar, \lowervar}^2 \lowerobj(\lowervar_{t+1}, \uppervar_{t+1})^{-1} \nabla \upperobj(\lowervar_{t+1}, \uppervar_{t+1})\!-\! \nabla_{\lowervar, \lowervar}^2 \lowerobj(\lowervar_{t}, \uppervar_{t}) \nabla \upperobj(\lowervar_{t}, \uppervar_{t})\| \\
            &\le 
                \begin{aligned}[t]
                    &\|\nabla_{\lowervar, \lowervar}^2 \lowerobj(\lowervar_{t+1}, \uppervar_{t+1})^{-1} \!-\! \nabla_{\lowervar, \lowervar}^2 \lowerobj(\lowervar_{t}, \uppervar_{t})^{-1}\| \cdot \|\nabla \upperobj(\lowervar_{t+1}, \uppervar_{t+1})\| \\
                    &\!+\! \|\nabla_{\lowervar, \lowervar}^2 \lowerobj(\lowervar_{t}, \uppervar_{t})^{-1}\| \cdot \|\nabla \upperobj(\lowervar_{t+1}, \uppervar_{t+1}) -  \nabla \upperobj(\lowervar_{t}, \uppervar_{t})\|
                \end{aligned}
\end{align*}\end{small}
Therefore, using similar arguments as in the proof of Lemma~\ref{lem:smoothness_h}, we obtain
{\small \begin{align*}
    \|\nabla_{\lowervar, \lowervar}^2 \lowerobj(\lowervar_{t+1}, \uppervar_{t+1})^{-1} - \nabla_{\lowervar, \lowervar}^2 \lowerobj(\lowervar_{t}, \uppervar_{t})^{-1}\| 
        &\le \frac{L_{\lowerobj, 2}^{\lowervar, \lowervar}}{\mu_{\lowerobj}^2} \left(\|\lowervar_{t+1} - \lowervar_t\| + \|\uppervar_{t+1} - \uppervar_t\| \right) \\
        &= \frac{L_{\lowerobj, 2}^{\lowervar, \lowervar}}{\mu_{\lowerobj}^2} \left(\tau \|\nabla_\lowervar \lowerobj(\lowervar_t, \uppervar_t)\| +  \gamma \|\apgstep_{\uppervar, t}\| \right)\\
        &\le \frac{L_{\lowerobj, 2}^{\lowervar, \lowervar}}{\mu_{\lowerobj}^2} \left(\tau L_{\lowerobj, 1}^{\lowervar, \lowervar} \|\lowervar_t - \lowersol_t\| + \gamma \|\apgstep_{\uppervar, t}\| \right)
\end{align*}}
where last line follows from \cref{assumption:lowerobj}.2.
Besides, using Assumptions~\ref{assumption:upperobj}.2 and~\ref{assumption:lowerobj}.1, we have
{\small \begin{align*}
    \|\nabla_{\lowervar, \lowervar}^2 \lowerobj(\lowervar_{t}, \uppervar_{t})^{-1}\| \cdot \|\nabla \upperobj(\lowervar_{t+1}, \uppervar_{t+1}) -  \nabla \upperobj(\lowervar_{t}, \uppervar_{t})\| 
        &\le \frac{L_{\upperobj,1}}{\mu_{\lowerobj}} \left( \|\lowervar_{t+1} - \lowervar_t\| + \|\uppervar_{t+1} - \uppervar_t\|\right) \\
        &\le \frac{L_{\upperobj,1}}{\mu_{\lowerobj}} \left( \tau L_{\lowerobj, 1}^{\lowervar, \lowervar} \|\lowervar_t - \lowersol_t\| + \gamma \|\apgstep_{\uppervar, t}\|\right) 
\end{align*}}
Hence, using one last time \cref{assumption:upperobj}.1, we get overall
{\small 
\begin{align*}
    \|\dualsol(\lowervar_{t+1}, \uppervar_{t+1}) -  \dualsol(\lowervar_{t}, \uppervar_{t})\| 
        &\le L_{\upperobj,0} \frac{L_{\lowerobj, 2}^{\lowervar, \lowervar}}{\mu_{\lowerobj}^2} \left(\tau L_{\lowerobj, 1}^{\lowervar, \lowervar} \|\lowervar_t - \lowersol_t\| + \gamma \|\apgstep_{\uppervar, t}\| \right) \\
        &+ \frac{L_{\upperobj,1}}{\mu_{\lowerobj}} \left( \tau L_{\lowerobj, 1}^{\lowervar, \lowervar} \|\lowervar_t - \lowersol_t\| + \gamma \|\apgstep_{\uppervar, t}\|\right) \\
        &= \constantdualdrift
                 \left(
                    \tau L_{\lowerobj, 1}^{\lowervar, \lowervar}  \|\lowervar_t - \lowersol_t\| 
                    + \gamma \|\apgstep_{\uppervar, t}\|
                \right)
\end{align*}}
\end{proof}
To get a bound on the dual error $\|\dualvar_t - \dualsol(\lowervar_t, \uppervar_t)\|$, we need finally to control the partial error stemming from the drift of the lower variable $\lowervar_t$ with respect to the dual solution $\|\dualsol(\lowervar_t, \uppervar_t) - \dualsol(\uppervar_t)\|$.
\begin{lemma}[\sv{Lower drift impact on the dual solution}]\label{lem:dualsol_sensitivity}
    For any $t \ge 0$, we have
    {\small \begin{displaymath}
        \|\dualsol(\lowervar_t, \uppervar_t) - \dualsol(\uppervar_t)\| \le \constantdualdrift \|\lowervar_t - \lowersol_t\|.
    \end{displaymath}}
\end{lemma}
\begin{proof}
    Observe that
    {\small \begin{align*}
        \|\dualsol(\lowervar_t, \uppervar_t) - \dualsol(\uppervar_t)\| 
            &= \|\nabla_{\lowervar, \lowervar}^2 \lowerobj(\lowervar_t, \uppervar_t)^{-1} \nabla \upperobj(\lowervar_t, \uppervar_t) - \nabla_{\lowervar, \lowervar}^2 \lowerobj(\lowersol_t, \uppervar_t)^{-1} \nabla \upperobj(\lowersol_t, \uppervar_t)\| \\
            &\le 
                \begin{aligned}[t]
                    &\|\nabla_{\lowervar, \lowervar}^2 \lowerobj(\lowervar_t, \uppervar_t)^{-1} - \nabla_{\lowervar, \lowervar}^2 \lowerobj(\lowersol_t, \uppervar_t)^{-1}\| \cdot \|\nabla \upperobj(\lowervar_t, \uppervar_t)\| \\
                    &+ \|\nabla_{\lowervar, \lowervar}^2 \lowerobj(\lowersol_t, \uppervar_t)^{-1}\| \cdot \|\nabla \upperobj(\lowervar_t, \uppervar_t) -  \nabla \upperobj(\lowersol_t, \uppervar_t)\|
                \end{aligned}
    \end{align*}}
    and using the same arguments as in the proof of Lemma~\ref{lem:smoothness_h}, we obtain:
    {\small \begin{align*}
        \|\nabla_{\lowervar, \lowervar}^2 \lowerobj(\lowervar_t, \uppervar_t)^{-1} - \nabla_{\lowervar, \lowervar}^2 \lowerobj(\lowersol_t, \uppervar_t)^{-1}\| \cdot \|\nabla \upperobj(\lowervar_t, \uppervar_t)\|
            &\le \frac{L_{\lowerobj, 2}^{\lowervar, \lowervar}}{\mu_{\lowerobj}^2} \cdot L_{\upperobj, 0} \|\lowervar_t - \lowersol_t\|, \\
        \|\nabla_{\lowervar, \lowervar}^2 \lowerobj(\lowersol_t, \uppervar_t)^{-1}\| \cdot \|\nabla \upperobj(\lowervar_t, \uppervar_t) -  \nabla \upperobj(\lowersol_t, \uppervar_t)\| 
            &\le \frac{L_{\upperobj,1}}{\mu_{\lowerobj}} \|\lowervar_t - \lowersol_t\|.
    \end{align*}}
    Summing the two above inequalities gives the result.
\end{proof}
Our recursive bound on the dual error $\|\dualvar_t - \dualsol(\uppervar_t)\|$ finally writes as follows.
\begin{lemma}[\sv{Control of the dual error}]\label{lem:dualsol_distance}
    For any $t \ge 0$, we have 
    {\small \begin{align*}
        \|\dualvar_{t+1} - \dualsol(\uppervar_{t+1})\|^2
            &\le \begin{aligned}[t]
                \;& \left(1-\frac{\rho \mu_{\lowerobj}}{2}\right) \; \|\dualvar_{t} - \dualsol(\uppervar_{t})\|^2 + \frac{107}{\rho \mu_{\lowerobj}} \cdot \constantdualdrift^2 \|\lowervar_t - \lowersol_t)\|^2 \\ 
                 &+ \frac{25}{\rho \mu_{\lowerobj}} \; \constantdualdrift^2
                        \begin{aligned}[t]
                            \bigg[& \frac{2}{\tau \mu_{\lowerobj}} \left(\frac{M_\cC}{\mu_{\lowerobj}}\right)^2 + 3 \bigg] \gamma^2 \|\apgstep_{\uppervar, t}\|^2.
                        \end{aligned}                                 
                \end{aligned} 
    \end{align*}}
\end{lemma}
\begin{proof}
    For any $\alpha, \beta > 0$, we have by Young's inequality,
    {\small \begin{align*}
        \|\dualvar_{t+1} - \dualsol(\uppervar_{t+1})\|^2 
            &= (1+\alpha) \; \|\dualvar_{t+1} - \dualsol(\lowervar_{t+1}, \uppervar_{t+1})\|^2 + (1+\tfrac{1}{\alpha}) \; \|\dualsol(\lowervar_{t+1}, \uppervar_{t+1}) - \dualsol(\uppervar_{t+1})\|^2 \\
            &\le \begin{aligned}[t]
                        &(1+\alpha) \; (1+\beta) \; \|\dualvar_{t+1} - \dualsol(\lowervar_{t}, \uppervar_{t})\|^2 \\
                        & + (1+\alpha) \; (1+1/\beta) \; \|\dualsol(\lowervar_{t}, \uppervar_{t}) - \dualsol(\lowervar_{t+1}, \uppervar_{t+1})\|^2 \\
                        & + (1+1/\alpha) \; \|\dualsol(\lowervar_{t+1}, \uppervar_{t+1}) - \dualsol(\uppervar_{t+1})\|^2                      
                  \end{aligned}   
    \end{align*}}
    Thus, plugging Lemmas~\ref{lem:lower-level-descent},~\ref{lem:dualsol_update}, and~\ref{lem:dualsol_sensitivity} into the above inequality gives
    {\small \begin{equation}\label{eq:dual_descent_eq_a}
    \begin{aligned}
        \|\dualvar_{t+1} - \dualsol(\uppervar_{t+1})\|^2 
        &\le 
            (1+\alpha)(1+\beta)(1-\rho \mu_{\lowerobj}) 
            \|\dualvar_{t} - \dualsol(\lowervar_{t}, \uppervar_{t})\|^2 \\
        &\quad + (1+\alpha)(1+1/\beta)\, 2 \tau^2 (L_{\lowerobj, 1}^{\lowervar, \lowervar})^2 
            \Bigl( L_{\upperobj, 0} \frac{L_{\lowerobj, 2}^{\lowervar, \lowervar}}{\mu_{\lowerobj}^2} 
                 + \frac{L_{\upperobj, 1}}{\mu_{\lowerobj}} \Bigr)^2 
            \|\lowervar_t - \lowersol_t\|^2 \\
        &\quad + (1+1/\alpha) 
            \Bigl( L_{\upperobj, 0} \frac{L_{\lowerobj, 2}^{\lowervar, \lowervar}}{\mu_{\lowerobj}^2} 
                 + \frac{L_{\upperobj, 1}}{\mu_{\lowerobj}} \Bigr)^2  
            \|\lowervar_{t+1} - \lowersol_{t+1}\|^2 \\
        &\quad + (1+\alpha)(1+1/\beta)\, 2 \gamma^2 
            \Bigl( L_{\upperobj, 0} \frac{L_{\lowerobj, 2}^{\lowervar, \lowervar}}{\mu_{\lowerobj}^2} 
                 + \frac{L_{\upperobj, 1}}{\mu_{\lowerobj}} \Bigr)^2 
            \|\apgstep_{\uppervar, t}\|^2.
\end{aligned}
\end{equation} }
    Leveraging one more time Lemma~\ref{lem:dualsol_sensitivity} together with Young's inequality, we derive for any $\delta>0$  
    {\small \begin{equation*}
        \|\dualvar_{t} - \dualsol(\lowervar_{t}, \uppervar_{t})\|^2 \le (1+\delta) \|\dualvar_{t} - \dualsol(\uppervar_{t})\|^2 + (1+1/\delta) \; \constantdualdrift^2 \|\lowervar_t - \lowersol_t\|^2
    \end{equation*}}
    Plugging the above inequality together with Lemma~\eqref{lem:lower-level-descent} into~\eqref{eq:dual_descent_eq_a} gives after some rearrangement:
    {\small \begin{align}\label{eq:dual_descent_eq_b}
        \|\dualvar_{t+1} - \dualsol(\uppervar_{t+1})\|^2             
            &\le 
                \begin{aligned}[t] 
                \;& (1-\rho \mu_{\lowerobj}) \; (1+\alpha) \; (1+\beta) \; (1+\delta) \; \|\dualvar_{t} - \dualsol(\uppervar_{t})\|^2 \\
                 &+ \constantdualdrift^2
                        \begin{aligned}[t]
                            \bigg[ & (1-\rho \mu_{\lowerobj}) \; (1+\alpha) \; (1+\beta) \; (1+1/\delta)\\
                             & + (1+\alpha) \; (1+1/\beta) \cdot 2 \tau^2 (L_{\lowerobj, 1}^{\lowervar, \lowervar})^2 \\
                             & + (1+1/\alpha) \; (1-\frac{\tau \mu_{\lowerobj}}{2}) \bigg] \|\lowervar_t - \lowersol_t)\|^2 
                        \end{aligned} \;   \\ 
                 &+ \constantdualdrift^2
                        \begin{aligned}[t]
                            \bigg[&(1+1/\alpha) \;  \frac{2}{\tau \mu_{\lowerobj}} \left(\frac{M_\cC}{\mu_{\lowerobj}}\right)^2 \\
                            &+ 2 (1+\alpha) \; (1+1/\beta) \Bigg] \gamma^2 \|\apgstep_{\uppervar, t}\|^2.
                        \end{aligned}                                 
            \end{aligned} 
    \end{align}}
    Finally, setting $\alpha=\beta=\delta=\frac{\rho \mu_{\lowerobj}}{24}$, we observe that $\alpha \le \frac{\mu_\lowerobj}{24 L_{\lowerobj, 1}^{\lowervar,\lowervar}} \le 1$. This first ensures that $(1+\alpha)^3 \le 1 + \alpha \sup_{1\le s \le 2} \frac{\mathrm{d}}{\mathrm{d}t} [t^3] \le 1 + 12 \alpha$, 
    which in turn implies
    {\small \begin{align*}
        (1+\alpha)^3 \cdot (1-\rho \mu_{\lowerobj})  
            &\le (1+12\alpha) (1-\rho \mu_{\lowerobj}) = (1+\frac{\rho \mu_{\lowerobj}}{2}) (1-\rho \mu_{\lowerobj}) \le (1-\frac{\rho\mu_{\lowerobj}}{2}).
    \end{align*}}
    And since $\rho \mu_{\lowerobj} \le 1$, we have
    {\small \begin{align*}
        (1+1/\alpha) &= 1 + 24/(\rho \mu_{\lowerobj}) \le \frac{25}{\rho \mu_{\lowerobj}} \\
        (1+\alpha)\cdot (1+1/\alpha) &\le (25/24) \cdot (1+1/\alpha) \le \frac{27}{\rho \mu_{\lowerobj}} \\
        (1+\alpha)^2\cdot (1+1/\alpha) &\le (25/24)^2 \cdot (1+1/\alpha) \le \frac{28}{\rho \mu_{\lowerobj}}
    \end{align*}}
    This allows to simplify~\eqref{eq:dual_descent_eq_b} into
    {\small \begin{align*}
        \|\dualvar_{t+1} - \dualsol(\uppervar_{t+1})\|^2             
            &\le
                \begin{aligned}[t]
                \;& \left(1-\frac{\rho \mu_{\lowerobj}}{2}\right) \; \|\dualvar_{t} - \dualsol(\uppervar_{t})\|^2 \\
                 &+ \frac{107}{\rho \mu_{\lowerobj}} \cdot \constantdualdrift^2 \|\lowervar_t - \lowersol_t)\|^2 \\ 
                 &+ \frac{25}{\rho \mu_{\lowerobj}} \; \constantdualdrift^2
                        \begin{aligned}[t]
                            \bigg[& \frac{2}{\tau \mu_{\lowerobj}} \left(\frac{M_\cC}{\mu_{\lowerobj}}\right)^2 + 3 \bigg] \gamma^2 \|\apgstep_{\uppervar, t}\|^2.
                        \end{aligned}                                 
                \end{aligned} 
    \end{align*}}
    where we also used the upper bounds $\tau \le 1 / L_{\lowerobj, 1}^{\lowervar, \lowervar}$ and $(1-\tau \mu_{\lowerobj}/2) \le 1$.    
\end{proof}
\subsubsection{Control of the error on implicit gradient estimation}\label{sec:implicit_gradient_error}
We now turn to the control of the error on the implicit gradient estimate $\|\nabla \implicitobj(\uppervar_t) - \agrad_{\uppervar, t}\|$.
This proposition will be combined with the descent Lemma~\ref{lem:implicit_descent} on the implicit objective to derive our Lyapunov analysis in the upcoming Section~\ref{sec:proof_convergence_deterministic}.
\begin{lemma}[\sv{Error bound on the implicit gradient}]\label{lem:implicit_gradient_error}
    For any $t \ge 0$, we have
    {\small \begin{align*}
        \|\nabla \implicitobj(\uppervar_t) - \agrad_{\uppervar, t}\|
            &\le \left(L_{\upperobj, 1} + L_{\lowerobj, 2}^{\lowervar, \uppervar} \left(\frac{4 L_{\upperobj, 0}}{\rho \mu_{\lowerobj}^2} + \varepsilon_{\dualvar, 0}\right)\right) \|\lowervar_t - \lowersol_t\| + M_\cC \; \|\dualsol(\uppervar_t) - \dualvar_t\|.
    \end{align*}}
\end{lemma} 
\begin{proof}
    For any $t \ge 0$, we have by the triangular inequality:
    \begin{small}\begin{align*}
        \|\nabla \implicitobj(\uppervar_t) \!-\! \agrad_{\uppervar, t}\| 
            &\!=\! \|\nabla_{\uppervar} \upperobj(\lowersol_t, \uppervar_t) \!+\! \nabla_{\lowervar, \uppervar}^2 \lowerobj(\lowersol_t, \uppervar_t)^\top \dualsol(\uppervar_t) \!-\! (\nabla_{\uppervar} \upperobj(\lowervar_t, \uppervar_t) \!+\! \nabla_{\lowervar, \uppervar}^2 \lowerobj(\lowervar_t, \uppervar_t)^\top \dualvar_t)\| \\
            &\le \|\nabla_{\uppervar} \upperobj(\lowersol_t, \uppervar_t) \!-\! \nabla_{\uppervar} \upperobj(\lowervar_t, \uppervar_t)\| 
                    \!+\! \|\nabla_{\lowervar, \uppervar}^2 \lowerobj(\lowersol_t, \uppervar_t)^\top \dualsol(\uppervar_t) \!-\!  \nabla_{\lowervar, \uppervar}^2 \lowerobj(\lowervar_t, \uppervar_t)^\top \dualvar_t\| 
    \end{align*}\end{small}
    The first term can be upper bounded using \cref{assumption:upperobj}.2: $\|\nabla_{\uppervar} \upperobj(\lowersol_t, \uppervar_t) - \nabla_{\uppervar} \upperobj(\lowervar_t, \uppervar_t)\| \le L_{\upperobj, 1} \|\lowervar_t - \lowersol_t\|$.
    We now focus on the second term. First observe that
    {\small \begin{align*}
        &\|\nabla_{\lowervar, \uppervar}^2 \lowerobj(\lowersol_t, \uppervar_t) \dualsol(\uppervar_t) -  \nabla_{\lowervar, \uppervar}^2 \lowerobj(\lowervar_t, \uppervar_t) \dualvar_t\| \\
            &\quad \quad \quad \le
                \begin{aligned}[t]
                    &\|\nabla_{\lowervar, \uppervar}^2 \lowerobj(\lowersol_t, \uppervar_t) \| \cdot \|\dualsol(\uppervar_t) - \dualvar_t\| 
                        + \|\nabla_{\lowervar, \uppervar}^2 \lowerobj(\lowersol_t, \uppervar_t)  - \nabla_{\lowervar, \uppervar}^2 \lowerobj(\lowervar_t, \uppervar_t)\| \cdot \|\dualvar_t\|
                \end{aligned}
    \end{align*}}
    By Assumptions~\ref{assumption:lowerobj}.4 and \ref{assumption:lowerobj}.5, we note that $\|\nabla_{\lowervar, \uppervar}^2 \lowerobj(\lowersol_t, \uppervar_t) \| \le M_\cC$, and $\|\nabla_{\lowervar, \uppervar}^2 \lowerobj(\lowersol_t, \uppervar_t)  - \nabla_{\lowervar, \uppervar}^2 \lowerobj(\lowervar_t, \uppervar_t)\| \le L_{\lowerobj, 2}^{\lowervar, \uppervar} \|\lowervar_t - \lowersol_t\|$. 
    Furthermore, Lemma~\ref{lem:dual_bounded} gives $\|\dualvar_t\| \le \frac{4 L_{\upperobj, 0}}{\rho \mu_{\lowerobj}^2} + \varepsilon_{\dualvar, 0}$, so that 
    {\small \begin{align*}
        &\|\nabla_{\lowervar, \uppervar}^2 \lowerobj(\lowersol_t, \uppervar_t) \dualsol(\uppervar_t) -  \nabla_{\lowervar, \uppervar}^2 \lowerobj(\lowervar_t, \uppervar_t) \dualvar_t\| \le M_\cC \; \|\dualsol(\uppervar_t) - \dualvar_t\| + L_{\lowerobj, 2}^{\lowervar, \uppervar} \left(\frac{4 L_{\upperobj, 0}}{\rho \mu_{\lowerobj}^2} + \varepsilon_{\dualvar, 0}\right) \|\lowervar_t - \lowersol_t\|.
    \end{align*}}
    Hence, we have overall
    {\small \begin{align*}
        \|\nabla \implicitobj(\uppervar_t) - \agrad_{\uppervar, t}\| 
            &\le \left(L_{\upperobj, 1} + L_{\lowerobj, 2}^{\lowervar, \uppervar} \left(\frac{4 L_{\upperobj, 0}}{\rho \mu_{\lowerobj}^2} + \varepsilon_{\dualvar, 0}\right)\right) \|\lowervar_t - \lowersol_t\| + M_\cC \; \|\dualsol(\uppervar_t) - \dualvar_t\|. \\
    \end{align*}}
\end{proof}
As a direct consequence of the above lemma and the non-expansiveness of the projection operator on the convex set $\cC$, we immediately have the following corollary. 
\begin{corollary}[\sv{Error bound on the generalized gradient}]\label{cor:implicit_gradient_error}
    For any $t \ge 0$, we have
    {\small \begin{align*}
        \|\apgstep_{\uppervar, t} - \pgstep_{\uppervar, t}\|
            &\le \left(L_{\upperobj, 1} + L_{\lowerobj, 2}^{\lowervar, \uppervar} \left(\frac{4 L_{\upperobj, 0}}{\rho \mu_{\lowerobj}^2} + \varepsilon_{\dualvar, 0}\right)\right) \|\lowervar_t - \lowersol_t\| + M_\cC \; \|\dualsol(\uppervar_t) - \dualvar_t\|.
    \end{align*}}
\end{corollary}

\subsubsection{Proof of Theorem~\ref{thm:convergence_deterministic}}\label{sec:proof_convergence_deterministic}

\begin{proof}
    Let us introduce the Lyapunov function $\lyap_t = \implicitobj(\uppervar_t) + \alpha \|\lowervar_t - \lowersol_t\|^2 + \beta \|\dualvar_t - \dualsol(\uppervar_t)\|^2$ for some $\alpha, \beta > 0$.
    By Lemma~\ref{lem:deterministic_lyap_parameters_for_descent}, setting $\alpha$ and $\beta$ as 
    {\small \begin{align*}
        \alpha &= 6 \gamma \kappa_{\lowerobj, \lowervar} \left(L_{\upperobj, 1} + L_{\lowerobj, 2}^{\lowervar, \uppervar} \left(4 \kappa_{\lowerobj, \lowervar} \frac{L_{\upperobj, 0}}{\mu_{\lowerobj}} + \varepsilon_{\dualvar, 0}\right)\right)^2 
                    + 1284 \gamma \kappa_{\lowerobj, \lowervar}^3 M_\cC^2 \constantdualdrift^2 \\
        \beta &= 3 \gamma \kappa_{\lowerobj, \lowervar}  M_\cC^2,
    \end{align*}}
    together with the parameterization $\tau = \rho = \frac{1}{L_{\lowerobj, 1}^{\lowervar, \lowervar}}$, and $\gamma \le \min(\frac{1}{\kappa_{\lowerobj, \lowervar}}, \bar \gamma)$ where 
    {\small \begin{align}\label{eq:def_bar_gamma}            
        \bar \gamma^{-1} = & 2{L_{\implicitobj}} + 48 \; \kappa_{\lowerobj, \lowervar} \left(\tfrac{M_{\cC}}{\mu_{\lowerobj}} \right)^2 \left(L_{\upperobj, 1} + L_{\lowerobj, 2}^{\lowervar, \uppervar} \left(4 \kappa_{\lowerobj, \lowervar} \frac{L_{\upperobj, 0}}{\mu_{\lowerobj}} + \varepsilon_{\dualvar, 0}\right)\right)^2 \notag \\
                    & + 10272 \; \kappa_{\lowerobj, \lowervar}^3 \frac{M_\cC^4}{\mu_{\lowerobj^2}} \constantdualdrift^2 + 300 \; \kappa_{\lowerobj, \lowervar} M_\cC^2  \; \constantdualdrift^2 
                                            \bigg[2\kappa_{\lowerobj, \lowervar} \left(\frac{M_\cC}{\mu_{\lowerobj}}\right)^2 + 3 \bigg]. 
    \end{align}}
    ensures that $\lyap_{t+1} \le \lyap_t - \frac{\gamma}{8} \|\pgstep_{\uppervar, t}\|^2 - \tfrac{\alpha}{4 \kappa_{\lowerobj,\lowervar}} \|\lowervar_t - \lowersol_t\|^2$ for all $t \ge 0$.
    Summing this inequality for $t=0, \ldots, T-1$ gives
    {\small \begin{displaymath}
        \frac{1}{T} \sum_{t=0}^{T-1} \|\pgstep_{\uppervar, t}\|^2 + \frac{2 \alpha}{\gamma \kappa_{\lowerobj,\lowervar}} \|\lowervar_t - \lowersol_t\|^2 
            \le \frac{8}{\gamma T} (\lyap_0 - \lyap_T) \le \frac{8}{\gamma T} (\implicitobj(\uppervar_0) - \inf \implicitobj + \alpha \varepsilon_{\lowervar, 0}^2 + \beta \varepsilon_{\dualvar, 0}^2).
    \end{displaymath}}
    Setting $\bar \alpha = 2 \alpha / (\kappa_{\lowerobj, \lowervar}\;\gamma)$ and $\bar \beta = 2 \beta / (\kappa_{\lowerobj, \lowervar}\;\gamma)$ concludes the proof.

\end{proof}

\subsection{The stochastic case}
\label{sec:proof_convergence_stochastic}
We now turn to the proof of Theorem~\ref{thm:convergence_stochastic}.
To this end, we introduce stochastic errors on gradients estimates $\stoerror_{\upperobj, 1}^{\lowervar}$, $\stoerror_{\upperobj, 1}^{\uppervar}$, and $\stoerror_{\lowerobj, 1}, \stoerror_{\lowerobj, 2}^{\lowervar, \lowervar}, \stoerror_{\lowerobj, 2}^{\lowervar, \uppervar}$, as follows :
{\small \begin{align*}
    \stoerror_{\upperobj, 1, t}^{\lowervar} &= \stonabla_{\lowervar} \upperobj(\lowervar_t, \uppervar_t) - \nabla_{\lowervar} \upperobj(\lowervar_t, \uppervar_t) \quad  &\stoerror_{\lowerobj, 1, t} &= \stonabla_{\lowervar} \lowerobj(\lowervar_t, \uppervar_t) - \nabla_{\lowervar} \lowerobj(\lowervar_t, \uppervar_t)\\
    \stoerror_{\upperobj, 1, t}^{\uppervar} &= \stonabla_{\uppervar} \upperobj(\lowervar_t, \uppervar_t) - \nabla_{\uppervar} \upperobj(\lowervar_t, \uppervar_t) \quad  &\stoerror_{\lowerobj, 2, t}^{\lowervar, \lowervar} &= \stonabla_{\lowervar, \lowervar}^2 \lowerobj(\lowervar_t, \uppervar_t) - \nabla_{\lowervar, \lowervar}^2 \lowerobj(\lowervar_t, \uppervar_t) \\ 
     & &\stoerror_{\lowerobj, 2, t}^{\lowervar, \uppervar} &= \stonabla_{\lowervar, \uppervar}^2 \lowerobj(\lowervar_t, \uppervar_t) - \nabla_{\lowervar, \uppervar}^2 \lowerobj(\lowervar_t, \uppervar_t), 
\end{align*}}
and use now the notations $\apgstep_{\uppervar, t}, \agrad_{\uppervar, t},$ and $\pgstep_{\uppervar, t}$ as
{\small \begin{align*}
    \apgstep_{\uppervar, t} &\defineq \frac{1}{\todo{\gamma_t}} \left(\uppervar_t - \proj_{\cC}(\uppervar_t - \todo{\gamma_t} \agrad_{\uppervar, t}) \right) \\
    \agrad_{\uppervar, t} &\defineq \stonabla_{\uppervar} \upperobj(\lowervar_t, \uppervar_t) +  \stonabla_{\lowervar, \uppervar}^2 \lowerobj(\lowervar_t, \uppervar_t)^\top \dualvar_t, \\
    \pgstep_{\uppervar, t} &\defineq \frac{1}{\todo{\gamma_t}} \left(\uppervar_t - \proj_{\cC}(\uppervar_t - \todo{\gamma_t} \nabla \implicitobj(\uppervar_t))\right)
\end{align*}}
for the stochastic counterparts of the approximate gradient, the approximate projected gradient, and the exact projected gradient introduced in the last section.
\subsubsection{Expected lower-level descent}
We first provide stochastic counterparts to the descent properties derived in Section~\ref{sec:lower_descent}. 
The proofs, which follow closely those of the deterministic case -- with additional handling of the stochastic errors -- are deferred to the online supplementary material to this paper.
\begin{lemma}[\sv{Stochastic descent of the lower, dual and upper variables}]\label{lem:lower-level-descent-stochastic}
    \sv{Assume that the lower and dual stepsizes satisfy} $\tau, \rho \le 1/L_{\lowerobj, 1}^{\lowervar, \lowervar}$. 
    Then, for any $t \ge 0$, we have
    {\small \begin{align*}
        \expectation \left[\|\lowervar_{t+1} - \lowersol_{t}\|^2 \mid \mathcal{F}_t \right]
            &\le (1 - \tau \mu_{\lowerobj}) \|\lowervar_t - \lowersol_t\|^2 + \tau^2\; \cV_{\lowerobj, 1} \\
        \expectation \left[\|\dualvar_{t+1} - \dualsol(\lowervar_t, \uppervar_t)\|^2 \mid \mathcal{F}_t \right]
            &\le (1 - \rho \mu_{\lowerobj}) \|\dualvar_t - \dualsol(\lowervar_t, \uppervar_t)\|^2 + \rho^2\; \cV_{\lowerobj, 2}^{\lowervar, \lowervar} \|\dualvar_t\|^2 + \rho^2\; \cV_{\upperobj, 1}^{\lowervar} \\
        \expectation \left[\|\lowervar_{t+1} - \lowersol_{t+1}\|^2 \mid \mathcal{F}_t \right] 
            &\le \left(1 - \frac{\tau \mu_{\lowerobj}}{2}\right) \|\lowervar_t - \lowersol_t\|^2 
            + \frac{3}{\tau \mu_{\lowerobj}}  \left(\frac{M_{\cC}}{\mu_{\lowerobj}} \right)^2 \expectation[\|\gamma_t \apgstep_{\uppervar, t}\|^2 \mid \mathcal{F}_t]  
            + \frac{\tau^2 \kappa_{\lowerobj, \lowervar}}{\kappa_{\lowerobj, \lowervar} - 1} \cV_{\lowerobj, 1}.
    \end{align*}}
\end{lemma}
The proof is deferred to section~\ref{app:proof_lower_level_descent_stochastic} of the online supplementary material.

\subsubsection{Stochastic dual control}
We now establish stochastic variants of Lemmas~\ref{lem:dual_bounded} to~\ref{lem:dualsol_distance}, which proofs are all deferred to Section~\ref{app:proof_dual_control_stochastic} of the online supplementary material.
\begin{lemma}[\sv{Expected boundness of the dual variable}]\label{lem:dual_bounded_stochastic}
    \sv{Assume that the dual stepsize satisfies} $\rho \le \min(1/L_{\lowerobj, 1}^{\lowervar, \lowervar}, \mu_\lowerobj/4 \cV_{\lowerobj, 2}^{\lowervar, \lowervar})$. 
    Then, for any $t \ge 0$, we have
    {\small \begin{align*}
        \expectation\left[\|\dualvar_{t+1}\|^2 \mid \mathcal{F}_t \right] 
             &\le (1 - \rho \mu_{\lowerobj}/8) \|\dualvar_t\|^2 
                    + \frac{15}{\rho \mu_{\lowerobj}} \left(\frac{L_{\upperobj, 0}}{\mu_{\lowerobj}}\right)^2
                    + 2 \rho^2 \cV_{\upperobj, 1}^{\lowervar} \\
        \expectation\left[\|\dualvar_t\|^2\right] 
        &\le \constant_{\dualvar, \star} 
    \end{align*}}
where $\constant_{\dualvar, \star} \defineq 2 \varepsilon_{\dualvar, 0} + \frac{121}{\rho^2 \mu_{\lowerobj}^2} \left(\frac{L_{\upperobj, 0}}{\mu_{\lowerobj}}\right)^2 + \frac{16\rho}{\mu_{\lowerobj}} \cV_{\upperobj, 1}^{\lowervar}$.
\end{lemma}
\sv{We now turn our attention to the drift of the dual solution by bounding $\|\dualsol(\lowervar_{t}, \uppervar_{t}) - \dualsol(\uppervar_t)\|^2$, as a stochastic counterparts of Lemmas~\ref{lem:dualsol_update} and~\ref{lem:dualsol_sensitivity}.}
\begin{lemma}[\sv{Stochastic drift of the dual solution}]\label{lem:dualsol_update_stochastic}
    For any $t \ge 0$, we have
    {\small 
    \begin{align*}
        & \expectation\left[\|\dualsol(\lowervar_{t+1}, \uppervar_{t+1}) - \dualsol(\lowervar_t, \uppervar_t)\|^2 \mid \mathcal{F}_t \right] \le L_{\dualvar, \star}  \left(\tau^2 \left(L_{\lowerobj, 1}^{\lowervar, \lowervar}\right)^2\|\lowervar_t - \lowersol_t\|^2 + \tau^2 \cV_{\lowerobj, 1} + \expectation\left[ \|\gamma_t\apgstep_{\uppervar, t}\|^2 \mid \mathcal{F}_t\right)\right], \\
        & \|\dualsol(\lowervar_{t}, \uppervar_{t}) - \dualsol(\uppervar_t)\|^2 \le L_{\dualvar, \star} \|\lowervar_t - \lowersol_t \|^2 
    \end{align*}}
    where $L_{\dualvar, \star} \defineq 2 \mu_{\lowerobj}^{-2}\; \left(\left(L_{\upperobj, 0} \cdot L_{\lowerobj,2}^{\lowervar, \lowervar}/\mu_{\lowerobj}\right)^2 + \left(L_{\upperobj, 1}\right)^2 \right)$. 
\end{lemma}
\yl{Finally the lemma below provides a stochastic counterpart of Lemma~\ref{lem:dualsol_distance}}.
\begin{lemma}[\sv{Stochastic control of the dual error}]\label{lem:dualsol_distance_stochastic}
    Assume \sv{that the lower stepsize satisfies} $\tau \le 1/L_{\lowerobj, 1}^{\lowervar, \lowervar}$ and \sv{that the dual stepsize satisfies} $\rho \le \min(1/L_{\lowerobj, 1}^{\lowervar, \lowervar}, \mu_\lowerobj/4 \cV_{\lowerobj, 2}^{\lowervar, \lowervar})$.
    For any $t \ge 0$, we have
    {\small \begin{align*}
    \expectation\left[\|\dualvar_{t+1} - \dualsol(\uppervar_{t+1})\|^2 \right]
      &\le \begin{aligned}[t]
            &(1-\rho \mu_{\lowerobj}/2) \; \|\dualvar_{t} - \dualsol(\uppervar_{t})\|^2 \\
            &+ 28\; \rho^{-1} \mu_{\lowerobj}^{-1} \; L_{\dualvar, \star} \left(2 + \tau^2 \left(L_{\lowerobj, 1}^{\lowervar, \lowervar}\right)^2 \right) \|\lowervar_t - \lowersol_t \|^2 \\
            &+ 27\; \rho^{-1} \mu_{\lowerobj}^{-1} \; L_{\dualvar, \star} \; \left(1 + 3 \tau^{-1} \mu_\lowerobj^{-3} M_{\cC}^2\right) \expectation[\|\gamma_t \apgstep_{\uppervar, t}\|^2 \mid \mathcal{F}_t] \\
            &+ 27\; \tau^2 \rho^{-1} \mu_{\lowerobj}^{-1} \; L_{\dualvar, \star}\frac{\kappa_{\lowerobj, \lowervar}}{\kappa_{\lowerobj, \lowervar} - 1}\; \cV_{\lowerobj, 1} \\
            &+ 4\; \rho^2 \left(\cV_{\upperobj, 1}^{\lowervar} + \cV_{\lowerobj, 2}^{\lowervar, \lowervar} \constant_{\dualvar, \star} \right)
      \end{aligned} \\          
\end{align*}}
where $L_{\dualvar, \star}$ is the constant defined in Lemma~\ref{lem:dualsol_update_stochastic}.
\end{lemma}

\subsubsection{Stochastic control of the error on the implicit gradient estimate}
We now extend Lemma~\ref{lem:implicit_gradient_error} to the stochastic setting.
\begin{lemma}[\sv{Stochastic error bound on the implicit gradient}]\label{lem:implicit_gradient_error_stochastic}
    For any $t \ge 0$, we have
    {\small \begin{align*}
        &\expectation\left[\|\nabla \implicitobj(\uppervar_t) - \agrad_{\uppervar, t}\|^2 \right] 
            \le \begin{aligned}[t]
                & 2 (L_{\upperobj, 1})^2 \expectation[\|\lowervar_t - \lowersol_t\|^2] + 4 M_\cC^2 \; \expectation[\|\dualvar_t - \dualsol(\uppervar_t)\|^2] + 4 (L_{\lowerobj, 2}^{\lowervar, \uppervar})^2 \; \expectation[\|\lowervar_t - \lowersol_t\|^2 \cdot \|v_t\|^2] \\
                & + \cV_{\upperobj, 1}^{\uppervar} + \cV_{\lowerobj, 2}^{\lowervar, \uppervar} \; \constant_{\dualvar, \star}.                
                \end{aligned} 
    \end{align*}}
and $\expectation\left[\|\apgstep_{\uppervar, t} - \pgstep_{\uppervar, t}\|^2 \right]$ satisfies the same upper bound.
\end{lemma}
\begin{proof}
First note that, by \cref{assumption:stochastic_gradients}, we have $\expectation[\agrad_{\uppervar, t} \mid \mathcal{F}_{t-1}] = \nabla_{\uppervar} \upperobj(\lowervar_t, \uppervar_t) +  \nabla_{\lowervar, \uppervar}^2 \lowerobj(\lowervar_t, \uppervar_t)^\top \dualvar_t$.
Hence, for any $t \ge 0$, 
{\small \begin{align*}
    & \expectation\left[\|\nabla \implicitobj(\uppervar_t) - \expectation[\agrad_{\uppervar, t} \mid \mathcal{F}_{t-1}]\|^2 \mid \mathcal{F}_{t-1} \right] \\
    & \hspace{3em} \le \begin{aligned}[t] 
                    &  2 \|\nabla_\uppervar \upperobj(\lowersol_t, \uppervar_t) - \nabla_\uppervar \upperobj(\lowervar_t, \uppervar_t)\|^2 + 2 \|\nabla_{\lowervar, \uppervar}^2 \lowerobj(\lowersol_t, \uppervar_t)^\top \dualsol(\uppervar_t) - \nabla_{\lowervar, \uppervar}^2 \lowerobj(\lowervar_t, \uppervar_t)^\top \dualvar_t\|^2
                \end{aligned} \\
    & \hspace{3em} \le \begin{aligned}[t]
                    &  2 (L_{\upperobj, 1})^2 \|\lowervar_t - \lowersol_t\|^2 + 4 M_\cC^2 \; \|\dualvar_t - \dualsol(\uppervar_t)\|^2 + 4 (L_{\lowerobj, 2}^{\lowervar, \uppervar})^2 \; \|\lowervar_t - \lowersol_t\|^2 \cdot \|v_t\|^2. 
              \end{aligned} 
\end{align*}}
Furthermore, we have by \cref{assumption:stochastic_gradients}
{\small \begin{align*}
    \expectation\left[\|\expectation[\agrad_{\uppervar, t}] - \agrad_{\uppervar, t}\|^2 | \mathcal{F}_{t, \dualvar} \right] 
        &\le \expectation\left[\|\nabla_{\uppervar} \upperobj(\lowervar_t, \uppervar_t) - \stonabla_{\uppervar} \upperobj(\lowervar_t, \uppervar_t)\|^2 | \mathcal{F}_{t, \dualvar} \right] \\
        & \hspace{1em} + \expectation \left[\|\nabla_{\lowervar, \uppervar}^2 \lowerobj(\lowervar_t, \uppervar_t) - \stonabla_{\lowervar, \uppervar}^2 \lowerobj(\lowervar_t, \uppervar_t) \|^2 \cdot \|\dualvar_t\|^2  | \mathcal{F}_{t, \dualvar} \right] \\
        &\le \cV_{\upperobj, 1}^{\uppervar} + \cV_{\lowerobj, 2}^{\lowervar, \uppervar} \; \|\dualvar_t\|^2.
\end{align*}}
Hence, by standard properties of the conditional expectation, we obtain 
{\small \begin{align*}
    & \expectation\left[\|\nabla \implicitobj(\uppervar_t) - \agrad_{\uppervar, t}\|^2 \right] \\
    &\hspace{5em}\le \begin{aligned}[t]
        & 2 (L_{\upperobj, 1})^2 \expectation[\|\lowervar_t - \lowersol_t\|^2] + 4 M_\cC^2 \; \expectation[\|\dualvar_t - \dualsol(\uppervar_t)\|^2] + 4 (L_{\lowerobj, 2}^{\lowervar, \uppervar})^2 \; \expectation[\|\lowervar_t - \lowersol_t\|^2 \cdot \|v_t\|^2] \\
        & + \cV_{\upperobj, 1}^{\uppervar} + \cV_{\lowerobj, 2}^{\lowervar, \uppervar} \; \constant_{\dualvar, \star}.
        \end{aligned}         
\end{align*}}
where last line follows again from Lemma~\ref{lem:dual_bounded_stochastic}.
The same upper bound is satisfied by $\expectation\left[\|\apgstep_{\uppervar, t} - \pgstep_{\uppervar, t}\| \right]$ thanks to the contraction properties of the projection operator on $\cC$.
\end{proof}
In contrast with Lemma~\ref{lem:implicit_gradient_error}, we have now an additional bound on $\expectation[\|\lowervar_t - \lowersol_t\|^2 \cdot \|v_t\|^2]$. 
The following lemma enables to control it recursively. 
\begin{lemma}[\sv{Stochastic control of the joint dynamic}]\label{lem:expected_product}
    For any $t \ge 0$, we have
    {\small \begin{align*}
        &\expectation[\|\lowervar_{t+1} - \lowersol_{t+1}\|^2 \cdot \|v_{t+1}\|^2] \\
            &\le \begin{aligned}[t]
                & \left(1 - \frac{\tau \mu_{\lowerobj}}{2}\right) \left(1-\rho \mu_\lowerobj/8\right) \expectation \left[ \|\lowervar_t - \lowersol_t\|^2 \cdot \|v_t\|^2 \right]\\
                &+ \left(1-\rho \mu_\lowerobj/8\right) \cdot  3 \tau^{-1} \mu_{\lowerobj}^{-3} M_{\cC}^2 \expectation \left[ \expectation[\|\gamma_t \apgstep_{\uppervar, t}\|^2 \mid \mathcal{F}_t] \cdot \|v_t\|^2 \right]\\
                &+ \left(1-\rho \mu_\lowerobj/8\right) \tau^2 \frac{\kappa_{\lowerobj, \lowervar}}{\kappa_{\lowerobj, \lowervar} - 1} \cV_{\lowerobj, 1} \expectation \left[ \|v_t\|^2 \right]\\
                &+ \left(1 - \frac{\tau \mu_{\lowerobj}}{2}\right) \left(\frac{15}{\rho \mu_{\lowerobj}} \left(\frac{L_{\upperobj, 0}}{\mu_{\lowerobj}}\right)^2
                    + 2 \rho^2 \cV_{\upperobj, 1}^{\lowervar}\right)  \expectation \left[ \|\lowervar_t - \lowersol_t\|^2 \right]\\
                &+ 3 \tau^{-1} \mu_{\lowerobj}^{-3} M_{\cC}^2  \left(\frac{15}{\rho \mu_{\lowerobj}} \left(\frac{L_{\upperobj, 0}}{\mu_{\lowerobj}}\right)^2
                    + 2 \rho^2 \cV_{\upperobj, 1}^{\lowervar}\right)  \expectation \left[ \expectation[\|\gamma_t \apgstep_{\uppervar, t}\|^2 \mid \mathcal{F}_t] \right] \\
                &+ \tau^2 \left(\frac{15}{\rho \mu_{\lowerobj}} \left(\frac{L_{\upperobj, 0}}{\mu_{\lowerobj}}\right)^2
                    + 2 \rho^2 \cV_{\upperobj, 1}^{\lowervar}\right) \frac{\kappa_{\lowerobj, \lowervar}}{\kappa_{\lowerobj, \lowervar} - 1} \cV_{\lowerobj, 1}
            \end{aligned}       
    \end{align*}}
\end{lemma}
\begin{proof}
First observe that, in the first upper bound given in Lemma~\ref{lem:dual_bounded_stochastic}, one can condition on $\mathcal{F}_{t, \lowervar}$ instead of $\mathcal{F}_t$ without affecting the bound, so that,
{\small 
\begin{align*}
    \expectation[\|\lowervar_{t+1} - \lowersol_{t+1}\|^2 \cdot \|v_{t+1}\|^2] 
        &= \expectation\left[\|\lowervar_{t+1} - \lowersol_{t+1}\|^2 \cdot \expectation[\|v_{t+1}\|^2 \mid \mathcal{F}_{t, \lowervar}]\right] \\
        &\le \expectation\left[\|\lowervar_{t+1} - \lowersol_{t+1}\|^2 \left(\| \left(1-\rho \mu_\lowerobj/8\right) \|v_t\|^2 + \frac{15}{\rho \mu_{\lowerobj}} \left(\frac{L_{\upperobj, 0}}{\mu_{\lowerobj}}\right)^2
            + 2 \rho^2 \cV_{\upperobj, 1}^{\lowervar} \right)\right]. 
\end{align*}}
Thus, conditioning now on $\mathcal{F}_t$ and using Lemma~\ref{lem:lower-level-descent-stochastic}, we obtain
{\small \begin{align*}
    &\expectation[\|\lowervar_{t+1} - \lowersol_{t+1}\|^2 \cdot \|v_{t+1}\|^2] \\         
        &\hspace{3em} \le \expectation 
                \begin{aligned}[t]
                    &\Bigg[ \begin{aligned}[t]
                                   \Bigg( &\left(1 - \frac{\tau \mu_{\lowerobj}}{2}\right) \|\lowervar_t - \lowersol_t\|^2 + 3 \tau^{-1} \mu_{\lowerobj}^{-3} M_{\cC}^2 \expectation[\|\gamma_t \apgstep_{\uppervar, t}\|^2 \mid \mathcal{F}_t] + \tau^2 \frac{\kappa_{\lowerobj, \lowervar}}{\kappa_{\lowerobj, \lowervar} - 1} \cV_{\lowerobj, 1} \Bigg)
                            \end{aligned} \\
                    &\hspace{1em} \cdot \left(\left(1-\rho \mu_\lowerobj/8\right) \|v_t\|^2 + \frac{15}{\rho \mu_{\lowerobj}} \left(\frac{L_{\upperobj, 0}}{\mu_{\lowerobj}}\right)^2
                        + 2 \rho^2 \cV_{\upperobj, 1}^{\lowervar} 
                    \right)                    \Bigg] \\
                \end{aligned} \\        
        &\hspace{3em} \le 
                \begin{aligned}[t]
                    & \left(1 - \frac{\tau \mu_{\lowerobj}}{2}\right) \left(1-\rho \mu_\lowerobj/8\right) \expectation \left[ \|\lowervar_t - \lowersol_t\|^2 \cdot \|v_t\|^2 \right]\\
                    &+ \left(1-\rho \mu_\lowerobj/8\right) \cdot  3 \tau^{-1} \mu_{\lowerobj}^{-3} M_{\cC}^2 \expectation \left[ \expectation[\|\gamma_t \apgstep_{\uppervar, t}\|^2 \mid \mathcal{F}_t] \cdot \|v_t\|^2 \right]\\
                    &+ \left(1-\rho \mu_\lowerobj/8\right) \tau^2 \frac{\kappa_{\lowerobj, \lowervar}}{\kappa_{\lowerobj, \lowervar} - 1} \cV_{\lowerobj, 1} \expectation \left[ \|v_t\|^2 \right]\\
                    &+ \left(1 - \frac{\tau \mu_{\lowerobj}}{2}\right) \left(\frac{15}{\rho \mu_{\lowerobj}} \left(\frac{L_{\upperobj, 0}}{\mu_{\lowerobj}}\right)^2
                        + 2 \rho^2 \cV_{\upperobj, 1}^{\lowervar}\right)  \expectation \left[ \|\lowervar_t - \lowersol_t\|^2 \right]\\
                    &+ 3 \tau^{-1} \mu_{\lowerobj}^{-3} M_{\cC}^2  \left(\frac{15}{\rho \mu_{\lowerobj}} \left(\frac{L_{\upperobj, 0}}{\mu_{\lowerobj}}\right)^2
                        + 2 \rho^2 \cV_{\upperobj, 1}^{\lowervar}\right)  \expectation \left[ \expectation[\|\gamma_t \apgstep_{\uppervar, t}\|^2 \mid \mathcal{F}_t] \right] \\
                    &+ \tau^2 \left(\frac{15}{\rho \mu_{\lowerobj}} \left(\frac{L_{\upperobj, 0}}{\mu_{\lowerobj}}\right)^2
                        + 2 \rho^2 \cV_{\upperobj, 1}^{\lowervar}\right) \frac{\kappa_{\lowerobj, \lowervar}}{\kappa_{\lowerobj, \lowervar} - 1} \cV_{\lowerobj, 1}\\
                \end{aligned}              
\end{align*}  }      
\end{proof}

\subsubsection{Proof of Theorem~\ref{thm:convergence_stochastic}}
We can now present the proof of Theorem~\ref{thm:convergence_stochastic}.
\begin{proof}
For the stochastic setting, we consider the Lyapunov function 
{\small \begin{displaymath}
    \lyap_t \defineq \expectation[\implicitobj(\uppervar_t)] + \alpha\; \expectation[\|\lowervar_t - \lowersol_t\|^2] + \beta\; \expectation[\|\dualvar_t - \dualsol(\uppervar_t)\|^2] + \delta\; \expectation[\|\lowervar_t - \lowersol_t\|^2 \cdot \|v_t\|^2]
\end{displaymath}}
where $\alpha, \beta, \delta$ are non-negative parameteres to be fixed subsequently in the analysis.
By Lemma~\ref{sup:lem:descent_lyap_stochastic}, we have for all $t \ge 0$,
{\small \begin{align}\label{eq:lyap_descent_stochastic}
    \lyap_{t+1} \le 
        \begin{aligned}[t]
            \lyap_t &- \expectation\left[\frac{\gamma_t}{8} \|\apgstep_{\uppervar, t}\|^2\right] - \expectation\left[\frac{\gamma_t}{8} \|\pgstep_{\uppervar, t}\|^2\right] - \frac{\alpha \tau \mu_{\lowerobj}}{4} \expectation[\|\lowervar_t - \lowersol_t\|^2] + \expectation[\constant_{\lyap_t}] \\
                    &- \expectation\left[\constant_{D, t} \|\apgstep_{\uppervar, t}\|^2 + \constant_{\lowervar, t} \|\lowervar_t - \lowersol_t\|^2 + \constant_{\dualvar, t} \|\dualvar_t - \dualsol(\uppervar_t)\|^2 + \constant_{\lowervar, \dualvar, t} \|\dualvar_t - \dualsol(\uppervar_t)\|^2 \cdot \|\dualvar_t\|^2 \right] \\
        \end{aligned}
\end{align}}
where the five quantities $\constant_{D,t}, \constant_{\lowervar, t}, \constant_{\dualvar, t}, \constant_{\lowervar, \dualvar, t}$ and $\constant_{\lyap, t}$ are specified in Lemma~\ref{sup:lem:descent_lyap_stochastic}.
To ensure the non-negativity of the first four ones, we set $\alpha, \beta, \delta$ as specified in Lemma~\ref{sup:lem:parameter_choices_stochastic}, together with the parameterization $\tau=1/L_{\lowerobj, 1}^{\lowervar, \lowervar}, \rho=\min(1/L_{\lowerobj, 1}^{\lowervar, \lowervar}, \mu_{\lowerobj}/(4 \cV_{\lowerobj, 2}^{\lowervar, \lowervar}))$ and $\gamma = \min\left(\bar{\bar\gamma}, L_{\implicitobj}^{-1}\right)$ where 
{\small 
\begin{equation}\label{eq:def:barbargamma}
    {\bar{\bar\gamma}}^{-1} \defineq \begin{aligned}[t]
                        2 \; L_{\implicitobj} 
                            &+ 72\; M_\cC^2 
                                 \kappa_{\lowerobj, \lowervar}^2 (L_{\upperobj, 1})^2 L_{\implicitobj}^{-1} \mu_{\lowerobj}^{-2} \\
                            &+ 32256 \;  L_{\dualvar, \star} M_\cC^4 
                                \kappa_{\lowerobj, \lowervar}^2 L_{\implicitobj}^{-1} \mu_{\lowerobj}^{-2} \left(\kappa_{\lowerobj, \lowervar} + 4 \mu_\lowerobj^{-2} \cV_{\upperobj, 1}^{\lowervar}\right)^2   \\     
                            &+ 288\; \kappa_{\lowerobj, \lowervar}^2  M_\cC^2 (L_{\lowerobj, 2}^{\lowervar, \lowervar})^2 L_{\implicitobj}^{-1} \mu_{\lowerobj}^{-2} \left(15 \left(\kappa_{\lowerobj, \lowervar} + 4 \mu_\lowerobj^{-2} \cV_{\upperobj, 1}^{\lowervar}\right) \left(\frac{L_{\upperobj, 0}}{\mu_{\lowerobj}}\right)^2 + 2  \frac{\cV_{\upperobj, 1}^{\lowervar}}{(L_{\lowerobj, 1}^{\lowervar, \lowervar})^2}\right) \Bigg) \\
                            &+ 648 \; L_{\dualvar, \star} M_{\cC}^2 L_{\implicitobj}^{-1}  \left(1 + 3\; \kappa_{\lowerobj, \lowervar} \mu_{\lowerobj}^{-2} M_{\cC}^2\right) \left(\kappa_{\lowerobj, \lowervar} + 4 \mu_\lowerobj^{-2} \cV_{\upperobj, 1}^{\lowervar}\right)^2 \\
                            &+ 72 \; \kappa_{\lowerobj, \lowervar}^2 (L_{\lowerobj, 2}^{\lowervar, \lowervar})^2 M_{\cC}^2 L_{\implicitobj}^{-1} \mu_{\lowerobj}^{-2}  \left(15 \left(\kappa_{\lowerobj, \lowervar} + 4 \mu_\lowerobj^{-2} \cV_{\upperobj, 1}^{\lowervar}\right) \left(\frac{L_{\upperobj, 0}}{\mu_{\lowerobj}}\right)^2 + 2\; \frac{\cV_{\upperobj, 1}^{\lowervar}}{(L_{\lowerobj, 1}^{\lowervar, \lowervar})^2}\right). 
            \end{aligned}    
\end{equation}
}
Under this parameterization, the remaining term $\constant_{\lyap, t}$ can be upper bounded by a constant $\constant_{\lyap}$ we specify in Lemma~\ref{sup:lem:constant_lyap_stochastic}.
This allows us to unroll \eqref{eq:lyap_descent_stochastic} to get
{\small \begin{align*}
    \frac{1}{T} \sum_{t=0}^{T-1} \expectation\left[\frac{\gamma_t}{8} \left(\|\pgstep_{\uppervar, t}\|^2 + \|\apgstep_{\uppervar, t}\|^2 \right) \right] + \frac{\alpha \tau \mu_{\lowerobj}}{4} \expectation[\|\lowervar_t - \lowersol_t\|^2] \le \frac{1}{T} \left(\lyap_0 - \inf_{\cC} \implicitobj \right) + \constant_{\lyap}.         
\end{align*}}
This implies 
{\small \begin{align}
    \frac{1}{T} \sum_{t=0}^{T-1} \expectation\left[\indicator_{\|\apgstep_{\uppervar, t}\| \le C_\gamma} \left(\|\pgstep_{\uppervar, t}\|^2 + \|\apgstep_{\uppervar, t}\|^2 \right) \right] + \frac{\alpha \tau \mu_{\lowerobj}}{4\gamma} \expectation[\|\lowervar_t - \lowersol_t\|^2]   
        &\le \frac{1}{\gamma T} \left(\lyap_0 - \inf_{\cC} \implicitobj \right) + \frac{\constant_{\lyap}}{\gamma} \label{eq:ergodic_bound_a}\\
    \frac{1}{T} \sum_{t=0}^{T-1} \expectation\left[\frac{C_\gamma}{\|\apgstep_{\uppervar, t}\|} \indicator_{\|\apgstep_{\uppervar, t}\| > C_\gamma} \left(\|\pgstep_{\uppervar, t}\|^2 + \|\apgstep_{\uppervar, t}\|^2 \right) \right]
        &\le \frac{1}{\gamma T} \left(\lyap_0 - \inf_{\cC} \implicitobj \right) + \frac{\constant_{\lyap}}{\gamma}. \label{eq:ergodic_bound_b}                  
\end{align}}
%
Using the standard inequality $a^2 \ge 2 \varepsilon a - \varepsilon^2$ for $a, \varepsilon \ge 0$, we derive from~\eqref{eq:ergodic_bound_a} that
{\small \begin{align*}
    \frac{1}{T} \sum_{t=0}^{T-1} \expectation\left[\indicator_{\|\apgstep_{\uppervar, t}\| \le C_\gamma} \|\pgstep_{\uppervar, t}\| \right] + \frac{1}{2} \left(\frac{\alpha \tau \mu_{\lowerobj}}{\gamma}\right)^{1/2} \expectation[\|\lowervar_t - \lowersol_t\|] 
        &\le \left(\frac{1}{\gamma T} \left(\lyap_0 - \inf_{\cC} \implicitobj \right) + \frac{\constant_{\lyap}}{\gamma}\right)^{1/2},   
\end{align*}}
where we set $\varepsilon = \left(\frac{1}{T} \left(\lyap_0 - \inf_{\cC} \implicitobj \right) + \constant_{\lyap}\right)^{1/2}$.
Furthermore, noting that $\frac{1}{\|\apgstep_{\uppervar, t}\|} (\|\pgstep_{\uppervar, t}\|^2 + \|\apgstep_{\uppervar, t}\|^2) \ge \frac{2}{\|\apgstep_{\uppervar, t}\|} \left(\|\pgstep_{\uppervar, t}\| \cdot \|\apgstep_{\uppervar, t}\|\right) = 2 \|\pgstep_{\uppervar, t}\|$, we derive from~\eqref{eq:ergodic_bound_b} that
{\small \begin{displaymath}
    \frac{1}{T} \sum_{t=0}^{T-1} \expectation\left[ \indicator_{\|\apgstep_{\uppervar, t}\| > C_\gamma} \|\pgstep_{\uppervar, t}\| \right] 
        \le \frac{1}{C_\gamma} \left(\frac{1}{\gamma T} \left(\lyap_0 - \inf_{\cC} \implicitobj \right) + \frac{\constant_{\lyap}}{\gamma}\right).
\end{displaymath}}
Combining these two inequalities allows us to conclude that:
{\small \begin{displaymath}
    \frac{1}{T} \sum_{t=0}^{T-1} \expectation\left[\|\pgstep_{\uppervar, t}\| \right] + \frac{1}{2} \left(\frac{\alpha \tau \mu_{\lowerobj}}{\gamma}\right)^{1/2} \expectation[\|\lowervar_t - \lowersol_t\|]  
        \le \left(\frac{1}{\gamma T} \left(\lyap_0 - \inf_{\cC} \implicitobj \right) + \frac{\constant_{\lyap}}{\gamma}\right)^{1/2} 
               +  \frac{1}{C_\gamma} \left(\frac{1}{\gamma T} \left(\lyap_0 - \inf_{\cC} \implicitobj \right) + \frac{\constant_{\lyap}}{\gamma}\right),
\end{displaymath}}
which writes as~\eqref{eq:thm_convergence_stochastic} for $[\tilde{\alpha}, \tilde{\beta}, \tilde{\delta}] = \frac{1}{4 \gamma \kappa_\lowerobj} [\alpha, \beta, \delta]$, i.e.
{\small \begin{align}
    \tilde \alpha &= \begin{aligned}[t]
            \Bigg( \frac{3}{2}\; (L_{\upperobj, 1})^2  
                &+ 672 \;  \left(\kappa_{\lowerobj, \lowervar} + 4 \mu_\lowerobj^{-2} \cV_{\upperobj, 1}^{\lowervar}\right)^2 L_{\dualvar, \star} M_{\cC}^2   \\
                &+ 6 (L_{\lowerobj, 2}^{\lowervar, \lowervar})^2 \left(15 \left(\kappa_{\lowerobj, \lowervar} + 4 \mu_\lowerobj^{-2} \cV_{\upperobj, 1}^{\lowervar}\right) \left(\frac{L_{\upperobj, 0}}{\mu_{\lowerobj}}\right)^2 + 2  \frac{\cV_{\upperobj, 1}^{\lowervar}}{(L_{\lowerobj, 1}^{\lowervar, \lowervar})^2}\right) \Bigg)
        \end{aligned} \notag \\
    \tilde \beta &= \frac{3}{2} \left(1 + 4 (L_{\lowerobj, 1}^{\lowervar, \lowervar})^{-1} \mu_\lowerobj^{-1} \cV_{\upperobj, 1}^{\lowervar}\right) M_{\cC}^2, \quad \tilde \delta = \frac{3}{2} (L_{\lowerobj, 2}^{\lowervar, \lowervar})^2. \label{eq:def_tilde_params_sto} 
\end{align}}
and $M = [M_1, M_2, M_3, M_4, M_5]^\top$ with the entries given by $M_1 = \tfrac{3}{4}$, $M_2 = 24 \; M_{\cC}^2 (L_{\lowerobj,1}^{\lowervar,\lowervar})^{-1}\,\mu_{\lowerobj}^{-1}$, $M_{4} = \tfrac{3}{4}\; \cC_{\dualvar,\star}$, $M_5 = 24\; \cC_{\dualvar,\star} \; M_{\cC}^2 \;(L_{\lowerobj,1}^{\lowervar,\lowervar})^{-1}\,\mu_{\lowerobj}^{-1}$, and  
{\small \begin{equation}\label{eq:def_M_sto}
\begin{aligned}
 M_3 &= \Bigg[
    6\,L_{\upperobj,1}^2\,\frac{1}{\kappa_{\lowerobj,\lowervar}-1}\,\mu_{\lowerobj}^{-2} +\; 2688 \;L_{\dualvar,\star}\,M_{\cC}^2\;
            \frac{\big(\kappa_{\lowerobj,\lowervar}+4\mu_{\lowerobj}^{-2}\,\cV_{\upperobj,1}^{\lowervar}\big)^2}{(\kappa_{\lowerobj,\lowervar}-1)\,\mu_{\lowerobj}^{2}} \\[0.5ex]
    &\qquad\; +\; 24\;\frac{\kappa_{\lowerobj,\lowervar}^2}{\kappa_{\lowerobj,\lowervar}-1}\;\Bigg(
            15\big(\kappa_{\lowerobj,\lowervar}+4\mu_{\lowerobj}^{-2}\,\cV_{\upperobj,1}^{\lowervar}\big)\Big(\frac{L_{\upperobj,0}}{\mu_{\lowerobj}}\Big)^2
            + 2\frac{\cV_{\upperobj,1}^{\lowervar}}{(L_{\lowerobj,1}^{\lowervar,\lowervar})^2}
        \Bigg) \\[0.5ex]
    &\qquad\; +\; 324\; L_{\dualvar,\star}\,M_{\cC}^2\;
            \frac{\big(\kappa_{\lowerobj,\lowervar}+4\mu_{\lowerobj}^{-2}\,\cV_{\upperobj,1}^{\lowervar}\big)^2}{(\kappa_{\lowerobj,\lowervar}-1)\,\mu_{\lowerobj}}\;(L_{\lowerobj,1}^{\lowervar,\lowervar})^{-1} +\; 6\;\cC_{\dualvar,\star}\,(L_{\lowerobj,2}^{\lowervar,\lowervar})^2\;\frac{1}{(\kappa_{\lowerobj,\lowervar}-1)\,\mu_{\lowerobj}^{2}} \\[0.5ex]
    &\qquad\; +\; 6\,(L_{\lowerobj,2}^{\lowervar,\lowervar})^2\;\frac{1}{(\kappa_{\lowerobj,\lowervar}-1)\,\mu_{\lowerobj}^{2}}\;
            \Bigg(
                15\big(\kappa_{\lowerobj,\lowervar}+4\mu_{\lowerobj}^{-2}\,\cV_{\upperobj,1}^{\lowervar}\big)\Big(\frac{L_{\upperobj,0}}{\mu_{\lowerobj}}\Big)^2
                + 2\frac{\cV_{\upperobj,1}^{\lowervar}}{(L_{\lowerobj,1}^{\lowervar,\lowervar})^2}
            \Bigg)
    \Bigg],
\end{aligned} 
\end{equation}
}
\end{proof}}

\clearpage
\yl{\thispagestyle{empty} 
\begin{center} 
  \Large 
  Online supplementary material for the paper titled "Fairness-informed Pareto Optimization : An Efficient Bilevel Framework"
\end{center}

\section{Missing proofs from Section~\ref{sec:proof_convergence_stochastic}}

\subsection{Proof for expected lower-level descent}\label{app:proof_lower_level_descent_stochastic}

\setcounter{theorem}{21}
\begin{lemma}[\sv{Stochastic descent of the lower, dual and upper variables}]
    \sv{Assume that the lower and dual stepsizes satisfy} $\tau, \rho \le 1/L_{\lowerobj, 1}^{\lowervar, \lowervar}$. 
    Then, for any $t \ge 0$, we have
    {\small \begin{align*}
        \expectation \left[\|\lowervar_{t+1} - \lowersol_{t}\|^2 \mid \mathcal{F}_t \right]
            &\le (1 - \tau \mu_{\lowerobj}) \|\lowervar_t - \lowersol_t\|^2 + \tau^2\; \cV_{\lowerobj, 1} \\
        \expectation \left[\|\dualvar_{t+1} - \dualsol(\lowervar_t, \uppervar_t)\|^2 \mid \mathcal{F}_t \right]
            &\le (1 - \rho \mu_{\lowerobj}) \|\dualvar_t - \dualsol(\lowervar_t, \uppervar_t)\|^2 + \rho^2\; \cV_{\lowerobj, 2}^{\lowervar, \lowervar} \|\dualvar_t\|^2 + \rho^2\; \cV_{\upperobj, 1}^{\lowervar} \\
        \expectation \left[\|\lowervar_{t+1} - \lowersol_{t+1}\|^2 \mid \mathcal{F}_t \right] 
            &\le \left(1 - \frac{\tau \mu_{\lowerobj}}{2}\right) \|\lowervar_t - \lowersol_t\|^2 
            + \frac{3}{\tau \mu_{\lowerobj}}  \left(\frac{M_{\cC}}{\mu_{\lowerobj}} \right)^2 \expectation[\|\gamma_t \apgstep_{\uppervar, t}\|^2 \mid \mathcal{F}_t]  
            + \frac{\tau^2 \kappa_{\lowerobj, \lowervar}}{\kappa_{\lowerobj, \lowervar} - 1} \cV_{\lowerobj, 1}.
    \end{align*}}
\end{lemma}
\begin{proof}
For any $t \ge 0$, we have
{\small \begin{align*}
    \|\lowervar_{t+1} - \lowersol_{t}\|^2 
        &= \|\lowervar_t - \tau \stonabla_{\lowervar} \lowerobj(\lowervar_t, \uppervar_t) - \lowersol_t\|^2 \\
        &= \|\lowervar_t - \tau \nabla_{\lowervar} \lowerobj(\lowervar_t, \uppervar_t) - \lowersol_t - \tau \stoerror_{\lowerobj, 1, t}\|^2 \\
        &= \|\lowervar_t - \tau \nabla_{\lowervar} \lowerobj(\lowervar_t, \uppervar_t) - \lowersol_t\|^2 - 2\tau \langle \lowervar_t - \tau \nabla_{\lowervar} \lowerobj(\lowervar_t, \uppervar_t) - \lowersol_t, \stoerror_{\lowerobj, 1, t}\rangle + \tau^2 \|\stoerror_{\lowerobj, 1, t}\|^2.
\end{align*}
}
And taking the expectation w.r.t. the stochasticity at iteration $t$ gives 
{
\small \begin{align*}
    \expectation\left[ \|\lowervar_{t+1} - \lowersol_{t}\|^2 \mid \mathcal{F}_t \right]
    &\le \|\lowervar_t - \tau \nabla_{\lowervar} \lowerobj(\lowervar_t, \uppervar_t) - \lowersol_t\|^2 + \tau^2\; \cV_{\lowerobj, 1} \\
    &\le (1 - \tau \mu_{\lowerobj}) \|\lowervar_t - \lowersol_t\|^2 + \tau^2\; \cV_{\lowerobj, 1} 
\end{align*}
}
where last inequality follows from the analysis of the deterministic case (Lemma~\ref{lem:lower-level-descent}), assuming $\tau \le 1/L_{\lowerobj, 1}^{\lowervar, \lowervar}$.
Similarly, we have for all $t \ge 0$,
{\small \begin{align*}
    \|\dualvar_{t+1} - \dualsol(\lowervar_{t}, \uppervar_{t})\|^2 
        &=  \|\dualvar_t  - \rho (\stonabla_{\lowervar}^2 \lowerobj(\lowervar_t, \uppervar_t) \dualvar_t + \stonabla_{\lowervar} \upperobj(\lowervar_t, \uppervar_t)) - \dualsol(\lowervar_{t}, \uppervar_{t})\|^2 \\
        &= \|\dualvar_t  - \rho (\nabla_{\lowervar}^2 \lowerobj(\lowervar_t, \uppervar_t) \dualvar_t + \nabla_{\lowervar} \upperobj(\lowervar_t, \uppervar_t)) - \dualsol(\lowervar_{t}, \uppervar_{t})\|^2 \\
        &\quad \quad \quad + 2\; \rho\; \langle \dualvar_t  - \rho (\nabla_{\lowervar}^2 \lowerobj(\lowervar_t, \uppervar_t) \dualvar_t + \nabla_{\lowervar} \upperobj(\lowervar_t, \uppervar_t)) - \dualsol(\lowervar_{t}, \uppervar_{t}),\; \rho (\stoerror_{\lowerobj, 2, t}^{\lowervar, \lowervar} \dualvar_t + \stoerror_{\upperobj, 1, t}) \rangle \\
        &\quad \quad \quad + \rho^2 \|\stoerror_{\lowerobj, 2, t}^{\lowervar, \lowervar} \dualvar_t + \stoerror_{\upperobj, 1, t}\|^2.
\end{align*}
}
%
and using \sv{the linearity of the} conditional expectation, and independance between \sv{the two random estimates} $\stoerror_{\lowerobj, 2, t}^{\lowervar, \lowervar}$ and $\stoerror_{\upperobj, 1, t}$, we get
{\small \begin{align*}
    \expectation\left[ \|\dualvar_{t+1} - \dualsol(\lowervar_{t}, \uppervar_{t})\|^2 \mid \mathcal{F}_t \right] 
        &\le \|\dualvar_t  - \rho (\nabla_{\lowervar}^2 \lowerobj(\lowervar_t, \uppervar_t) \dualvar_t + \nabla_{\lowervar} \upperobj(\lowervar_t, \uppervar_t)) - \dualsol(\lowervar_{t}, \uppervar_{t})\|^2 \\
        &\quad \quad \quad + \rho^2 \left(\cV_{\lowerobj, 2}^{\lowervar, \lowervar}\; \|\dualvar_t\|^2 + \cV_{\upperobj, 1}^{\lowervar}\right) \\
        &\le \left(1- \rho \mu_{\lowerobj}\right) \; \|\dualvar_{t} - \dualsol(\lowervar_{t}, \uppervar_{t})\|^2 + \rho^2 \left(\cV_{\lowerobj, 2}^{\lowervar, \lowervar}\; \|\dualvar_t\|^2 + \cV_{\upperobj, 1}^{\lowervar}\right) 
\end{align*}}
where last inequality follows also from Lemma~\ref{lem:lower-level-descent} and having $\rho \le 1/L_{\lowerobj, 1}^{\lowervar, \lowervar}$.
Using now the same rationale as in the proof of Lemma~\ref{lem:lower_level_drift}, we have for any $\eta >0$,
{\small \begin{align*}
    \expectation[\|\lowervar_{t+1} - \lowersol_t\|^2 \mid \mathcal{F}_t]
        &\le (1+\eta)\; \expectation[\|\lowervar_{t+1} - \lowersol_t\|^2 \mid \mathcal{F}_t] + (1+\eta^{-1})\; \expectation[\|\lowersol_t - \lowersol_{t+1}\|^2 \mid \mathcal{F}_t] \\
        &\le (1+\eta) \left((1 - \tau \mu_{\lowerobj}) \|\lowervar_t - \lowersol_t\|^2 + \tau^2\; \cV_{\lowerobj, 1} \right) + (1+\eta^{-1})\; \left(\frac{M_{\cC}}{\mu_{\lowerobj}}\right)^2 \expectation[\|\gamma_t \apgstep_{\uppervar, t}\|^2 \mid \mathcal{F}_t] 
\end{align*}}
and setting again $\eta = \frac{\tau \mu_{\lowerobj}}{2(1-\tau \mu_{\lowerobj})}$ gives  
{\small \begin{align*}
    \|\lowervar_{t+1} - \lowersol_t\|^2 
        &\le \left(1 - \frac{\tau \mu_{\lowerobj}}{2}\right) \|\lowervar_t - \lowersol_t\|^2 
                + \left(\frac{M_{\cC}}{\mu_{\lowerobj}} \right)^2 \left(1 + \frac{2(1-\tau \mu_{\lowerobj})}{\tau \mu_{\lowerobj}} \right) \expectation[\|\gamma_t \apgstep_{\uppervar, t}\|^2 \mid \mathcal{F}_t]  \\
        &\quad + \tau^2 \left(1+ \frac{\tau \mu_{\lowerobj}}{2(1-\tau \mu_{\lowerobj})} \right) \cV_{\lowerobj, 1} \\
        &\le \left(1 - \frac{\tau \mu_{\lowerobj}}{2}\right) \|\lowervar_t - \lowersol_t\|^2 
                + \frac{3}{\tau \mu_{\lowerobj}} \left(\frac{M_{\cC}}{\mu_{\lowerobj}} \right)^2 \expectation[\|\gamma_t \apgstep_{\uppervar, t}\|^2 \mid \mathcal{F}_t]  
                + \tau^2 \frac{\kappa_{\lowerobj, \lowervar}}{\kappa_{\lowerobj, \lowervar} - 1} \cV_{\lowerobj, 1} 
\end{align*}}
where the last line follows from having $\tau \le 1/L_{\lowerobj, 1}^{\lowervar, \lowervar} \le 1/\mu_\lowerobj$ and $\kappa_{\lowerobj, \lowervar} \defineq L_{\lowerobj, 1}^{\lowervar, \lowervar}/\mu_{\lowerobj} \ge 1$.
\end{proof}

\subsection{Proofs of the stochastic dual control lemmas}\label{app:proof_dual_control_stochastic}

\begin{lemma}[\sv{Expected boundness of the dual variable}]
    \sv{Assume that the dual stepsize satisfies} $\rho \le \min(1/L_{\lowerobj, 1}^{\lowervar, \lowervar}, \mu_\lowerobj/4 \cV_{\lowerobj, 2}^{\lowervar, \lowervar})$. 
    Then, for any $t \ge 0$, we have
    {\small \begin{align*}
        \expectation\left[\|\dualvar_{t+1}\|^2 \mid \mathcal{F}_t \right] 
             &\le (1 - \rho \mu_{\lowerobj}/8) \|\dualvar_t\|^2 
                    + \frac{15}{\rho \mu_{\lowerobj}} \left(\frac{L_{\upperobj, 0}}{\mu_{\lowerobj}}\right)^2
                    + 2 \rho^2 \cV_{\upperobj, 1}^{\lowervar} \\
        \expectation\left[\|\dualvar_t\|^2\right] 
        &\le \constant_{\dualvar, \star} 
    \end{align*}}
where $\constant_{\dualvar, \star} \defineq 2 \varepsilon_{\dualvar, 0} + \frac{121}{\rho^2 \mu_{\lowerobj}^2} \left(\frac{L_{\upperobj, 0}}{\mu_{\lowerobj}}\right)^2 + \frac{16\rho}{\mu_{\lowerobj}} \cV_{\upperobj, 1}^{\lowervar}$.
\end{lemma}
\begin{proof}
By Lemma~\ref{lem:lower-level-descent-stochastic} and Young's inequality, we have for any $\rho \le 1/L_{\lowerobj, 1}^{\lowervar, \lowervar}$ and $\eta > 0$,
{\small 
\begin{align*}
    \expectation\left[\|\dualvar_{t+1} - \dualsol(\lowervar_{t}, \uppervar_{t})\|^2 \mid \mathcal{F}_t \right]                
        &\le (1 - \rho \mu_{\lowerobj}) \|\dualvar_t - \dualsol(\lowervar_{t}, \uppervar_{t})\|^2 + \rho^2 \left(\cV_{\lowerobj, 2}^{\lowervar, \lowervar}\; \|\dualvar_t\|^2 + \cV_{\upperobj, 1}^{\lowervar}\right) \\        
        &\le \left(((1+\eta) - \rho (\mu_{\lowerobj} (1+\eta) - \rho \cV_{\lowerobj, 2}^{\lowervar, \lowervar})\right)  \|\dualvar_t\|^2 \\ &\quad \quad + (1 - \rho \mu_{\lowerobj})(1 + 1/\eta) \|\dualsol(\lowervar_{t}, \uppervar_{t})\|^2 + \rho^2 \cV_{\upperobj, 1}^{\lowervar}.
\end{align*}
}
Setting $\eta = \rho \mu_{\lowerobj}/2$ and having $\rho \le \mu_{\lowerobj} / (4 \cV_{\lowerobj, 2}^{\lowervar, \lowervar})$, ensures that 
{\small
\[
    \left(((1+\eta) - \rho (\mu_{\lowerobj} (1+\eta) - \rho \cV_{\lowerobj, 2}^{\lowervar, \lowervar})\right) \le 1 - \rho \mu_{\lowerobj}/4.
\]
}  
This in turn gives
{\small \begin{align*}
    \expectation\left[\|\dualvar_{t+1} - \dualsol(\lowervar_{t}, \uppervar_{t})\|^2 \mid \mathcal{F}_t \right]                        
        &\le (1 - \rho \mu_{\lowerobj} / 4) \|\dualvar_t\|^2 + \frac{3}{\rho \mu_{\lowerobj}} \left(\frac{L_{\upperobj, 0}}{\mu_{\lowerobj}}\right)^2 + \rho^2 \cV_{\upperobj, 1}^{\lowervar}.
\end{align*}}
where we used the upper bound on $\|\dualsol(\lowervar_{0}, \uppervar_{0})\|^2$ given in Lemma~\ref{lem:dual_bounded} and that $0 \le \rho \mu_{\lowerobj} \le 1$.
Hence, using one more time Young's inequality, we obtain for any $\nu > 0$,
{\small 
\begin{align*}
    \expectation\left[\|\dualvar_{t+1}\|^2 \mid \mathcal{F}_t \right]                       
        &\le (1 + \nu) \expectation\left[\|\dualvar_{t+1} - \dualsol(\lowervar_{t}, \uppervar_{t})\|^2 \mid \mathcal{F}_t \right] + (1 + 1/\nu) \|\dualsol(\lowervar_{t}, \uppervar_{t})\|^2 \\
        &\le (1 + \nu)(1 - \rho \mu_{\lowerobj} / 4) \|\dualvar_t\|^2 
            \begin{aligned}[t]
                &+ (1 + \nu) \left(\frac{3}{\rho \mu_{\lowerobj}} \left(\frac{L_{\upperobj, 0}}{\mu_{\lowerobj}}\right)^2 + \rho^2 \cV_{\upperobj, 1}^{\lowervar}\right) \\ 
                &+ (1 + 1/\nu) \left(\frac{L_{\upperobj, 0}}{\mu_{\lowerobj}}\right)^2.    
            \end{aligned} 
\end{align*}}
Setting now $\nu = \rho \mu_{\lowerobj}/8$ gives $(1 + \nu)(1 - \rho \mu_{\lowerobj} / 4) \le 1 - \rho \mu_{\lowerobj}/8$ and we obtain 
{\small \begin{equation}\label{eq:recurive_bound_dualvar_stochastic}
    \expectation\left[\|\dualvar_{t+1}\|^2 \mid \mathcal{F}_t \right]                       
        \le (1 - \rho \mu_{\lowerobj}/8) \|\dualvar_t\|^2 
            + \frac{15}{\rho \mu_{\lowerobj}} \left(\frac{L_{\upperobj, 0}}{\mu_{\lowerobj}}\right)^2
            + 2 \rho^2 \cV_{\upperobj, 1}^{\lowervar}
\end{equation}}
where we used the upper bounds $1+\nu \le 2$, $1+1/\nu \le 9 /(\rho \mu_{\lowerobj})$.
By the tower property of conditional expectation, we can unroll the above recursion down to $t=0$ and obtain
{\small \begin{align*}
     \expectation\left[\|\dualvar_{t}\|^2 \right] 
        &\le (1 - \rho \mu_{\lowerobj}/8)^t \|\dualvar_0\|^2 + \left(\frac{15}{\rho \mu_{\lowerobj}} \left(\frac{L_{\upperobj, 0}}{\mu_{\lowerobj}}\right)^2 + 2 \rho^2 \cV_{\upperobj, 1}^{\lowervar}\right) \sum_{s=0}^{t-1} (1 - \rho \mu_{\lowerobj}/8)^s \\
        &\le (1 - \rho \mu_{\lowerobj}/8)^t \|\dualvar_0\|^2 + \frac{120}{\rho^2 \mu_{\lowerobj}^2} \left(\frac{L_{\upperobj, 0}}{\mu_{\lowerobj}}\right)^2 + \frac{16\rho}{\mu_{\lowerobj}} \cV_{\upperobj, 1}^{\lowervar} \\
        &\le 2 \varepsilon_{\dualvar, 0} + \frac{121}{\rho^2 \mu_{\lowerobj}^2} \left(\frac{L_{\upperobj, 0}}{\mu_{\lowerobj}}\right)^2 + \frac{16\rho}{\mu_{\lowerobj}} \cV_{\upperobj, 1}^{\lowervar} 
\end{align*}}
where last last makes one more time use of Young's inequality and the fact that $0 \le \rho \mu_{\lowerobj} \le 1$.
\end{proof}
\begin{lemma}[\sv{Stochastic drift of the dual solution}]
    For any $t \ge 0$, we have
    {\small 
    \begin{align*}
        & \expectation\left[\|\dualsol(\lowervar_{t+1}, \uppervar_{t+1}) - \dualsol(\lowervar_t, \uppervar_t)\|^2 \mid \mathcal{F}_t \right] \le L_{\dualvar, \star}  \left(\tau^2 \left(L_{\lowerobj, 1}^{\lowervar, \lowervar}\right)^2\|\lowervar_t - \lowersol_t\|^2 + \tau^2 \cV_{\lowerobj, 1} + \expectation\left[ \|\gamma_t\apgstep_{\uppervar, t}\|^2 \mid \mathcal{F}_t\right)\right], \\
        & \|\dualsol(\lowervar_{t}, \uppervar_{t}) - \dualsol(\uppervar_t)\|^2 \le L_{\dualvar, \star} \|\lowervar_t - \lowersol_t \|^2 
    \end{align*}}
    where $L_{\dualvar, \star} \defineq 2 \mu_{\lowerobj}^{-2}\; \left(\left(L_{\upperobj, 0} \cdot L_{\lowerobj,2}^{\lowervar, \lowervar}/\mu_{\lowerobj}\right)^2 + \left(L_{\upperobj, 1}\right)^2 \right)$. 
\end{lemma}
\begin{proof}
For any $t\ge 0$, using the same arguments as in Lemma~\ref{lem:dualsol_update}, we have
{\small 
\begin{align*}
    \|\nabla_{\lowervar, \lowervar}^2 \lowerobj(\lowervar_{t+1}, \uppervar_{t+1})^{-1} - \nabla_{\lowervar, \lowervar}^2 \lowerobj(\lowervar_t, \uppervar_t)^{-1}\|^2 
        &\le \left(\frac{L_{\lowerobj,2}^{\lowervar, \lowervar}}{\mu_{\lowerobj}^2}\right)^2 \left(\|\lowervar_{t+1} - \lowervar_{t}\|^2 + \|\uppervar_{t+1} - \uppervar_{t}\|^2\right) \\
        &= \left(\frac{L_{\lowerobj,2}^{\lowervar, \lowervar}}{\mu_{\lowerobj}^2}\right)^2 \left(\tau^2 \|\stonabla_\lowervar \lowerobj(\lowervar_t, \uppervar_t)\|^2 + \|\gamma_t\apgstep_{\uppervar, t}\|^2\right). 
\end{align*}}
Hence, using again \cref{assumption:upperobj}.1 we have, and taking the expectation w.r.t. $\mathcal{F}_t$ gives
{\small \begin{align*}
    &\expectation \left[\|\nabla_{\lowervar, \lowervar}^2 \lowerobj(\lowervar_{t+1}, \uppervar_{t+1})^{-1} - \nabla_{\lowervar, \lowervar}^2 \lowerobj(\lowervar_t, \uppervar_t)^{-1}\|^2 \right. \cdot \left.\|\nabla_{\lowervar} \upperobj(\lowervar_{t+1}, \uppervar_{t+1})\|^2 \mid \mathcal{F}_t \right] \\
        & \hspace{7em}\le \left(L_{\upperobj, 0}\right)^2 \left(\frac{L_{\lowerobj,2}^{\lowervar, \lowervar}}{\mu_{\lowerobj}^2}\right)^2 \left(\tau^2 \|\nabla_\lowervar \lowerobj(\lowervar_t, \uppervar_t)\|^2 + \tau^2 \cV_{\lowerobj, 1} + \expectation[  \|\gamma_t\apgstep_{\uppervar, t}\|^2 \mid \mathcal{F}_t] \right). \\
        & \hspace{7em}\le \left(L_{\upperobj, 0}\right)^2 \left(\frac{L_{\lowerobj,2}^{\lowervar, \lowervar}}{\mu_{\lowerobj}^2}\right)^2 \left(\tau^2 \left(L_{\lowerobj, 1}^{\lowervar, \lowervar}\right)^2\|\lowervar_t - \lowersol_t\|^2 + \tau^2 \cV_{\lowerobj, 1} + \expectation[ \|\gamma_t\apgstep_{\uppervar, t}\|^2 \mid \mathcal{F}_t] \right). 
\end{align*}}
Similarly, for any $t \ge 0$, we obtain
{\small \begin{align*}
    \expectation \left[\|\nabla_{\lowervar, \uppervar}^2 \lowerobj(\lowervar_t, \uppervar_t)^{-1}\|^2\right. &\cdot \left. \|\nabla_\lowerobj \upperobj(\lowervar_{t+1}, \uppervar_{t+1}) - \nabla_\lowerobj \upperobj(\lowervar_{t}, \uppervar_{t})\|^2 \mid \mathcal{F}_t \right] \\
        &\le \left(\frac{L_{\upperobj, 1}}{\mu_{\lowerobj}}\right)^2 \left(\tau^2 \left(L_{\lowerobj, 1}^{\lowervar, \lowervar}\right)^2\|\lowervar_t - \lowersol_t\|^2 + \tau^2 \cV_{\lowerobj, 1} + \expectation[ \|\gamma_t\apgstep_{\uppervar, t}\|^2 \mid \mathcal{F}_t]\right). 
\end{align*}}
Hence, using the convexity of $\|\cdot\|^2$ and combining the two previous bounds, we obtain
{\small \begin{align*}
    & \expectation\left[\|\dualsol(\lowervar_{t+1}, \uppervar_{t+1}) - \dualsol(\lowervar_t, \uppervar_t)\|^2 \mid \mathcal{F}_t \right] \\
        &\quad \quad \quad \le 2 \; \expectation \left[\|\nabla_{\lowervar, \lowervar}^2 \lowerobj(\lowervar_{t+1}, \uppervar_{t+1})^{-1} - \nabla_{\lowervar, \lowervar}^2 \lowerobj(\lowervar_t, \uppervar_t)^{-1}\|^2 \cdot \|\nabla_{\lowervar} \upperobj(\lowervar_{t+1}, \uppervar_{t+1})\|^2 \mid \mathcal{F}_t \right] \\
        &\quad \quad \quad \quad + 2\; \expectation \left[\|\nabla_{\lowervar, \uppervar}^2 \lowerobj(\lowervar_t, \uppervar_t)^{-1}\|^2 \cdot \|\nabla_\lowerobj \upperobj(\lowervar_{t+1}, \uppervar_{t+1}) - \nabla_\lowerobj \upperobj(\lowervar_{t}, \uppervar_{t})\|^2 \mid \mathcal{F}_t \right] \\        
        &\quad \quad \quad \le 2 \mu_{\lowerobj}^{-2}\; \left(\left(L_{\upperobj, 0} \cdot L_{\lowerobj,2}^{\lowervar, \lowervar}/\mu_{\lowerobj}\right)^2 + \left(L_{\upperobj, 1}\right)^2 \right)  \left(\tau^2 \left(L_{\lowerobj, 1}^{\lowervar, \lowervar}\right)^2\|\lowervar_t - \lowersol_t\|^2 + \tau^2 \cV_{\lowerobj, 1} + \expectation[ \|\gamma_t\apgstep_{\uppervar, t}\|^2 \mid \mathcal{F}_t] \right)          
\end{align*}}
The proof of the second inequality follows from the same arguments as in Lemma~\ref{lem:dualsol_update}. We omit the details for brevity.
\end{proof}
\begin{lemma}[\sv{Stochastic control of the dual error}]
    Assume \sv{that the lower stepsize satisfies} $\tau \le 1/L_{\lowerobj, 1}^{\lowervar, \lowervar}$ and \sv{that the dual stepsize satisfies} $\rho \le \min(1/L_{\lowerobj, 1}^{\lowervar, \lowervar}, \mu_\lowerobj/4 \cV_{\lowerobj, 2}^{\lowervar, \lowervar})$.
    For any $t \ge 0$, we have
    {\small \begin{align*}
    \expectation\left[\|\dualvar_{t+1} - \dualsol(\uppervar_{t+1})\|^2 \right]
      &\le \begin{aligned}[t]
            &(1-\rho \mu_{\lowerobj}/2) \; \|\dualvar_{t} - \dualsol(\uppervar_{t})\|^2 \\
            &+ 28\; \rho^{-1} \mu_{\lowerobj}^{-1} \; L_{\dualvar, \star} \left(2 + \tau^2 \left(L_{\lowerobj, 1}^{\lowervar, \lowervar}\right)^2 \right) \|\lowervar_t - \lowersol_t \|^2 \\
            &+ 27\; \rho^{-1} \mu_{\lowerobj}^{-1} \; L_{\dualvar, \star} \; \left(1 + 3 \tau^{-1} \mu_\lowerobj^{-3} M_{\cC}^2\right) \expectation[\|\gamma_t \apgstep_{\uppervar, t}\|^2 \mid \mathcal{F}_t] \\
            &+ 27\; \tau^2 \rho^{-1} \mu_{\lowerobj}^{-1} \; L_{\dualvar, \star}\frac{\kappa_{\lowerobj, \lowervar}}{\kappa_{\lowerobj, \lowervar} - 1}\; \cV_{\lowerobj, 1} \\
            &+ 4\; \rho^2 \left(\cV_{\upperobj, 1}^{\lowervar} + \cV_{\lowerobj, 2}^{\lowervar, \lowervar} \constant_{\dualvar, \star} \right)
      \end{aligned} \\          
\end{align*}}
where $L_{\dualvar, \star}$ is the constant defined in Lemma~\ref{lem:dualsol_update_stochastic}.
\end{lemma}
\begin{proof}
We follow similar steps to Lemma~\ref{lem:dualsol_distance}.
Using Lemma~\ref{lem:lower-level-descent-stochastic} and Young's inequality, we have for any $\alpha, \beta, \delta > 0$,      
{\small \begin{align*}
    \expectation\left[\|\dualvar_{t+1} - \dualsol(\uppervar_{t+1})\|^2 \mid \mathcal{F}_t \right]
      & \le \begin{aligned}[t]
            &(1+\alpha) \; (1+\beta) (1+\delta) (1-\rho \mu_{\lowerobj}) \; \|\dualvar_{t} - \dualsol(\uppervar_{t})\|^2 \\
            &+ (1+\alpha) \; (1+\beta) (1+\delta^{-1}) (1-\rho \mu_{\lowerobj}) \; \|\dualsol(\uppervar_{t}) - \dualsol(\lowervar_t, \uppervar_{t})\|^2 \\
            &+ (1+\alpha) \; (1+\beta) \rho^2 \left(\cV_{\lowerobj, 2}^{\lowervar, \lowervar} \|\dualvar_t\|^2 + \cV_{\upperobj, 1}^{\lowervar} \right) \\
            &+ (1+\alpha) \; (1+\beta^{-1}) \; \expectation\left[\|\dualsol(\lowervar_{t}, \uppervar_{t}) - \dualsol(\lowervar_{t+1}, \uppervar_{t+1})\|^2 \mid \mathcal{F}_t \right] \\
            &+ (1+\alpha^{-1}) \; \expectation\left[\|\dualsol(\lowervar_{t+1}, \uppervar_{t+1}) - \dualsol(\uppervar_{t+1})\|^2 \mid \mathcal{F}_t \right].                        
      \end{aligned} 
\end{align*}}
Using Lemma~\ref{lem:dualsol_update_stochastic}, we deduce that, for any $\rho \le 1/L_{\lowerobj, 1}^{\lowervar, \lowervar}$,
{\small \begin{align*}
    \expectation\left[\|\dualvar_{t+1} - \dualsol(\uppervar_{t+1})\|^2 \mid \mathcal{F}_t \right]
      & \le \begin{aligned}[t]
            &(1+\alpha) \; (1+\beta) (1+\delta) (1-\rho \mu_{\lowerobj}) \; \|\dualvar_{t} - \dualsol(\uppervar_{t})\|^2 \\
            &+ (1+\alpha) \; (1+\beta) (1+\delta^{-1}) (1-\rho \mu_{\lowerobj}) \; L_{\dualvar, \star} \|\lowervar_t - \lowersol_t \|^2 \\
            &+ (1+\alpha) \; (1+\beta)\; \rho^2 \left(\cV_{\lowerobj, 2}^{\lowervar, \lowervar} \|\dualvar_t\|^2 + \cV_{\upperobj, 1}^{\lowervar} \right) \\
            &+ (1+\alpha) \; (1+\beta^{-1}) \; \begin{aligned}[t]
                & L_{\dualvar, \star} \left(\tau^2 \left(L_{\lowerobj, 1}^{\lowervar, \lowervar}\right)^2\|\lowervar_t - \lowersol_t\|^2 + \tau^2 \cV_{\lowerobj, 1} + \expectation[\|\gamma_t \apgstep_{\uppervar, t}\|^2 \mid \mathcal{F}_t] \right) \\
            \end{aligned}   \\
            &+ (1+\alpha^{-1}) \; L_{\dualvar, \star} \expectation\left[\|\lowervar_{t+1} - \lowersol_{t+1} \|^2 \mid \mathcal{F}_t \right].
      \end{aligned} 
\end{align*}}
Using once more Lemma~\ref{lem:lower-level-descent-stochastic}, we obtain for any $\tau \le 1/L_{\lowerobj, 1}^{\lowervar, \lowervar}$,
{\small \begin{align*}
    \expectation\left[\|\dualvar_{t+1} - \dualsol(\uppervar_{t+1})\|^2 \mid \mathcal{F}_t \right]
      & \le \begin{aligned}[t]
            &(1+\alpha) \; (1+\beta) (1+\delta) (1-\rho \mu_{\lowerobj}) \; \|\dualvar_{t} - \dualsol(\uppervar_{t})\|^2 \\
            &+ (1+\alpha) \; (1+\beta) (1+\delta^{-1}) (1-\rho \mu_{\lowerobj}) \; L_{\dualvar, \star} \|\lowervar_t - \lowersol_t \|^2 \\
            &+ (1+\alpha) \; (1+\beta)\; \rho^2 \left(\cV_{\lowerobj, 2}^{\lowervar, \lowervar} \|\dualvar_t\|^2 + \cV_{\upperobj, 1}^{\lowervar} \right) \\
            &+ (1+\alpha) \; (1+\beta^{-1}) \; \begin{aligned}[t]
                & L_{\dualvar, \star} \left(\tau^2 \left(L_{\lowerobj, 1}^{\lowervar, \lowervar}\right)^2\|\lowervar_t - \lowersol_t\|^2 + \tau^2 \cV_{\lowerobj, 1} + \expectation[\|\gamma_t \apgstep_{\uppervar, t}\|^2 \mid \mathcal{F}_t]\right) \\
            \end{aligned}   \\
            &+ (1+\alpha^{-1}) \; 
                \begin{aligned}[t]
                    &L_{\dualvar, \star} \begin{aligned}[t]
                    & \bigg[\left(1 - \tau \mu_{\lowerobj}/2\right) \|\lowervar_t - \lowersol_t\|^2 \\
                        &\hspace{3em} + 3 \tau^{-1} \mu_{\lowerobj}^{-3} M_{\cC}^2 \expectation[\|\gamma_t \apgstep_{\uppervar, t}\|^2 \mid \mathcal{F}_t] \\
                        &\hspace{3em} + \tau^2 \frac{\kappa_{\lowerobj, \lowervar}}{\kappa_{\lowerobj, \lowervar} - 1} \cV_{\lowerobj, 1} \bigg] \\                    
                    \end{aligned}    
                \end{aligned} 
      \end{aligned} \\      
\end{align*}}
Setting again $\alpha = \beta = \delta = \rho \mu_{\lowerobj}/24$ and using the bounds derived in the proof of Lemma~\ref{lem:dualsol_distance}, we obtain:
{\small \begin{align*}
    \expectation\left[\|\dualvar_{t+1} - \dualsol(\uppervar_{t+1})\|^2 \mid \mathcal{F}_t \right]
      & \le \begin{aligned}[t]
            &(1-\rho \mu_{\lowerobj}/2) \; \|\dualvar_{t} - \dualsol(\uppervar_{t})\|^2 \\
            &+ 28\; \rho^{-1} \mu_{\lowerobj}^{-1} (1-\rho \mu_{\lowerobj}) \; L_{\dualvar, \star} \|\lowervar_t - \lowersol_t \|^2 \\
            &+ 4\; \rho^2 \left(\cV_{\lowerobj, 2}^{\lowervar, \lowervar} \|\dualvar_t\|^2 + \cV_{\upperobj, 1}^{\lowervar} \right) \\
            &+ 27\; \rho^{-1} \mu_{\lowerobj}^{-1} \; \begin{aligned}[t]
                & L_{\dualvar, \star} \left(\tau^2 \left(L_{\lowerobj, 1}^{\lowervar, \lowervar}\right)^2\|\lowervar_t - \lowersol_t\|^2 + \tau^2 \cV_{\lowerobj, 1} + \expectation[\|\gamma_t \apgstep_{\uppervar, t}\|^2 \mid \mathcal{F}_t]\right) \\
            \end{aligned}   \\
            &+ 25\; \rho^{-1} \mu_{\lowerobj}^{-1} \; 
                \begin{aligned}[t]
                    &L_{\dualvar, \star} 
                        \begin{aligned}[t]
                        \bigg[\left(1 - \tau \mu_{\lowerobj}/2\right) \|\lowervar_t - \lowersol_t\|^2 
                            & + 3 \tau^{-1} \mu_{\lowerobj}^{-3} M_{\cC}^2 \expectation[\|\gamma_t \apgstep_{\uppervar, t}\|^2 \mid \mathcal{F}_t] \\
                            & + \tau^2 \frac{\kappa_{\lowerobj, \lowervar}}{\kappa_{\lowerobj, \lowervar} - 1} \cV_{\lowerobj, 1} \bigg] \\                    
                    \end{aligned}    
                \end{aligned} 
      \end{aligned} \\      
      &\le \begin{aligned}[t]
            &(1-\rho \mu_{\lowerobj}/2) \; \|\dualvar_{t} - \dualsol(\uppervar_{t})\|^2 \\
            &+ 28\; \rho^{-1} \mu_{\lowerobj}^{-1} \; L_{\dualvar, \star} \left(2 + \tau^2 \left(L_{\lowerobj, 1}^{\lowervar, \lowervar}\right)^2 \right) \|\lowervar_t - \lowersol_t \|^2 \\
            &+ 27\; \rho^{-1} \mu_{\lowerobj}^{-1} \; L_{\dualvar, \star} \; \left(1 + 3 \tau^{-1} \mu_{\lowerobj}^{-3} M_{\cC}^2 \right) \expectation[\|\gamma_t \apgstep_{\uppervar, t}\|^2] \\
            &+ 
                \begin{aligned}[t]
                    & 4\; \rho^2 \cV_{\lowerobj, 2}^{\lowervar, \lowervar}  \|\dualvar_t\|^2 + 4\; \rho^2 \cV_{\upperobj, 1}^{\lowervar} + 27\; \rho^{-1} \mu_{\lowerobj}^{-1} \; L_{\dualvar, \star} \tau^2 \frac{2\kappa_{\lowerobj, \lowervar}}{\kappa_{\lowerobj, \lowervar} - 1} \cV_{\lowerobj, 1} \\
                \end{aligned}            
      \end{aligned} \\      
\end{align*}}
where we additionally used that $(1+\alpha) (1+\beta) = (1+\alpha)^2 \le 4$.
The desired result directly follows from the upper bound from Lemma~\ref{lem:dual_bounded_stochastic}: 
{\small \begin{align*}
    \expectation\left[\|\dualvar_{t+1} - \dualsol(\uppervar_{t+1})\|^2 \right]
      &\le \begin{aligned}[t]
            &(1-\rho \mu_{\lowerobj}/2) \; \|\dualvar_{t} - \dualsol(\uppervar_{t})\|^2 \\
            &+ 28\; \rho^{-1} \mu_{\lowerobj}^{-1} \; L_{\dualvar, \star} \left(2 + \tau^2 \left(L_{\lowerobj, 1}^{\lowervar, \lowervar}\right)^2 \right) \|\lowervar_t - \lowersol_t \|^2 \\
            &+ 27\; \rho^{-1} \mu_{\lowerobj}^{-1} \; L_{\dualvar, \star} \; \left(1 + 3 \tau^{-1} \mu_{\lowerobj}^{-3} M_{\cC}^2 \right) \expectation[\|\gamma_t \apgstep_{\uppervar, t}\|^2] \\
            &+ 27\; \tau^2 \rho^{-1} \mu_{\lowerobj}^{-1} \; L_{\dualvar, \star}\frac{\kappa_{\lowerobj, \lowervar}}{\kappa_{\lowerobj, \lowervar} - 1}\; \cV_{\lowerobj, 1} \\
            &+ 4\; \rho^2 \left(\cV_{\upperobj, 1}^{\lowervar} + \cV_{\lowerobj, 2}^{\lowervar, \lowervar} \constant_{\dualvar, \star} \right)
      \end{aligned} \\      
\end{align*}}
\end{proof}

\section{Supplementary derivations for our Lyapunov arguments}

In the next two sections, we provide the missing derivation steps for the Lyapunov decrease arguments presented in Appendix~\ref{app:convergence}.

\subsection{Parameter policy for the deterministic setting}
\setcounter{theorem}{27}
\begin{lemma}\label{lem:deterministic_lyap_parameters_for_descent}
    Assume $\tau = \rho = \frac{1}{L_{\lowerobj, 1}^{\lowervar, \lowervar}}$ and $\gamma \le \min(\frac{1}{\kappa_{\lowerobj, \lowervar}}, \bar \gamma)$ where 
    {\small \begin{align}\label{eq:def_bar_gamma_1}            
        \bar \gamma^{-1} = & 2{L_{\implicitobj}} + 48 \; \kappa_{\lowerobj, \lowervar} \left(\tfrac{M_{\cC}}{\mu_{\lowerobj}} \right)^2 \left(L_{\upperobj, 1} + L_{\lowerobj, 2}^{\lowervar, \uppervar} \left(4 \kappa_{\lowerobj, \lowervar} \frac{L_{\upperobj, 0}}{\mu_{\lowerobj}} + \varepsilon_{\dualvar, 0}\right)\right)^2 \notag \\
                    & + 10272 \; \kappa_{\lowerobj, \lowervar}^3 \frac{M_\cC^4}{\mu_{\lowerobj^2}} \constantdualdrift^2 + 300 \; \kappa_{\lowerobj, \lowervar} M_\cC^2  \; \constantdualdrift^2 
                                            \bigg[2\kappa_{\lowerobj, \lowervar} \left(\frac{M_\cC}{\mu_{\lowerobj}}\right)^2 + 3 \bigg]. 
    \end{align}}
    Assume further that $\alpha, \beta$ are set as 
    {\small \begin{align*}
        \alpha &= 6 \gamma \kappa_{\lowerobj, \lowervar} \left(L_{\upperobj, 1} + L_{\lowerobj, 2}^{\lowervar, \uppervar} \left(4 \kappa_{\lowerobj, \lowervar} \frac{L_{\upperobj, 0}}{\mu_{\lowerobj}} + \varepsilon_{\dualvar, 0}\right)\right)^2 
                    + 1284 \gamma \kappa_{\lowerobj, \lowervar}^3 M_\cC^2 \constantdualdrift^2 \\
        \beta &= 3 \gamma \kappa_{\lowerobj, \lowervar}  M_\cC^2.
    \end{align*}}
    Then, for any $\lyap_{t+1} \le \lyap_t - \frac{\gamma}{8} \|\pgstep_{\uppervar, t}\|^2 - \tfrac{\alpha}{4 \kappa_{\lowerobj,\lowervar}} \|\lowervar_t - \lowersol_t\|^2$ for all $t \ge 0$.    
\end{lemma}

\begin{proof}
    By Lemmas~\ref{lem:implicit_descent},~\ref{lem:implicit_gradient_error}, and Corollary~\ref{cor:implicit_gradient_error}, we have
    {\small \begin{align*}
        \implicitobj(\uppervar_{t+1}) 
            &\le \implicitobj(\uppervar_t) - \frac{\gamma}{8} \|\pgstep_{\uppervar, t}\|^2 
                                            - \frac{\gamma}{4} \|\apgstep_{\uppervar, t}\|^2
                                            + \frac{\gamma}{4} \|\apgstep_{\uppervar, t} - \pgstep_{\uppervar, t}\|^2
                                            + \frac{\gamma}{2} \|\nabla \implicitobj(\uppervar_t) - \agrad_{\uppervar, t}\|^2 
                                            + \frac{\gamma^2 L_{\implicitobj}}{2} \|\apgstep_{\uppervar, t}\|^2\\
            &\le \implicitobj(\uppervar_t)
                - \frac{\gamma}{8} \|\pgstep_{\uppervar, t}\|^2 
                - \frac{\gamma}{4} \left(1 - 2\gamma L_{\implicitobj}\right) \|\apgstep_{\uppervar, t}\|^2 \\ 
                &\quad +\frac{3\gamma}{2} \left(L_{\upperobj, 1} + L_{\lowerobj, 2}^{\lowervar, \uppervar} \left(\frac{4 L_{\upperobj, 0}}{\rho \mu_{\lowerobj}^2} + \varepsilon_{\dualvar, 0}\right)\right)^2 \|\lowervar_t - \lowersol_t\|^2 + \frac{3\gamma}{2} M_\cC^2 \; \|\dualsol(\uppervar_t) - \dualvar_t\|^2.
    \end{align*}}
    Hence, using Lemmas~\ref{lem:lower_level_drift} and~\ref{lem:dualsol_distance}, we obtain,
    {\small
    \begin{align*}
        \lyap_{t+1}
            &\le \implicitobj(\uppervar_t) - \frac{\gamma}{8} \|\pgstep_{\uppervar, t}\|^2 
            - \frac{\gamma}{4} \left(1 - 2 \gamma L_{\implicitobj}\right) \|\apgstep_{\uppervar, t}\|^2 + \frac{3\gamma}{2} \left(L_{\upperobj, 1} + L_{\lowerobj, 2}^{\lowervar, \uppervar} \left(\frac{4 L_{\upperobj, 0}}{\rho \mu_{\lowerobj}^2} + \varepsilon_{\dualvar, 0}\right)\right)^2 \|\lowervar_t - \lowersol_t\|^2 \\
            & + \frac{3\gamma}{2} M_\cC^2 \; \|\dualsol(\uppervar_t) - \dualvar_t\|^2 + \alpha \left(\Bigl(1 - \tfrac{\tau \mu_{\lowerobj}}{2}\Bigr) \|\lowervar_t - \lowersol_t\|^2
            \;+\; \tfrac{2}{\tau \mu_{\lowerobj}} \Bigl(\tfrac{M_{\cC}}{\mu_{\lowerobj}}\Bigr)^2 
            \gamma^2 \,\|\apgstep_{\uppervar, t}\|^2\right) \\
            & + \beta  \Bigg( \left(1-\frac{\rho \mu_{\lowerobj}}{2}\right) \; \|\dualvar_{t} - \dualsol(\uppervar_{t})\|^2  + \frac{107}{\rho \mu_{\lowerobj}} \cdot \constantdualdrift^2 \|\lowervar_t - \lowersol_t)\|^2 \\ 
            &\qquad + \frac{25}{\rho \mu_{\lowerobj}} \; \constantdualdrift^2 
            \bigg[\frac{2}{\tau \mu_{\lowerobj}} \left(\frac{M_\cC}{\mu_{\lowerobj}}\right)^2 + 3 \bigg] \gamma^2 \|\apgstep_{\uppervar, t}\|^2 \Bigg) \\
            &= \lyap_t - \frac{\gamma}{8} \|\pgstep_{\uppervar, t}\|^2 - \alpha \tfrac{\tau \mu_{\lowerobj}}{4} \|\lowervar_t - \lowersol_t\|^2 - \frac{\gamma}{4}
            \Bigg( 1 - 2 \gamma L_{\implicitobj} - 8 \gamma \tfrac{\alpha}{\tau \mu_{\lowerobj}} \Bigl(\tfrac{M_{\cC}}{\mu_{\lowerobj}} \;\Bigr)^2  \\
            &\qquad  + \gamma \beta \frac{100}{\rho \mu_{\lowerobj}} \; \constantdualdrift^2 
            \bigg[\frac{2}{\tau \mu_{\lowerobj}} \left(\frac{M_\cC}{\mu_{\lowerobj}}\right)^2 + 3 \bigg]  \Bigg) \|\apgstep_{\uppervar, t}\|^2 \\
            &\qquad - \left( \alpha \tfrac{\tau \mu_{\lowerobj}}{4} - \frac{3\gamma}{2} \left(L_{\upperobj, 1} + L_{\lowerobj, 2}^{\lowervar, \uppervar} \left(\frac{4 L_{\upperobj, 0}}{\rho \mu_{\lowerobj}^2} + \varepsilon_{\dualvar, 0}\right)\right)^2 
            - \beta \frac{107}{\rho \mu_{\lowerobj}} \cdot \constantdualdrift^2 \right) \\ 
            & \qquad \cdot \|\lowervar_t - \lowersol_t\|^2           
            - \left(\beta \frac{\rho \mu_{\lowerobj}}{2}  - \frac{3\gamma}{2} M_\cC^2 \right) \; \|\dualsol(\uppervar_t) - \dualvar_t\|^2 \\
    \end{align*}}
    Let us now set $\tau = \rho = \frac{1}{L_{\lowerobj, 1}^{\lowervar, \lowervar}}$, and 
    {\small \begin{align*}
        \alpha &= 6 \gamma \kappa_{\lowerobj, \lowervar} \left(L_{\upperobj, 1} + L_{\lowerobj, 2}^{\lowervar, \uppervar} \left(4 \kappa_{\lowerobj, \lowervar} \frac{L_{\upperobj, 0}}{\mu_{\lowerobj}} + \varepsilon_{\dualvar, 0}\right)\right)^2 
                    + 1284 \gamma \kappa_{\lowerobj, \lowervar}^3 M_\cC^2 \constantdualdrift^2 \\
        \beta &= 3 \gamma \kappa_{\lowerobj, \lowervar}  M_\cC^2
    \end{align*}}
    where $\kappa_{\lowerobj, \lowervar} = L_{\lowerobj, 1}^{\lowervar, \lowervar} / \mu_{\lowerobj}$ is the condition number of the lower-level problem w.r.t. $\lowervar$. 
    This turns the previous inequality into
    {\small 
    \begin{align*}
        \lyap_{t+1}
            \le \lyap_t &- \frac{\gamma}{8} \|\pgstep_{\uppervar, t}\|^2 - \tfrac{\alpha}{4 \kappa_{\lowerobj,\lowervar}} \|\lowervar_t - \lowersol_t\|^2 \\
            & - \frac{\gamma}{4} \Bigg( 1 - 2 \gamma L_{\implicitobj} - 48 \gamma^2 \kappa_{\lowerobj, \lowervar}^2 \left(\tfrac{M_{\cC}}{\mu_{\lowerobj}} \right)^2 \left(L_{\upperobj, 1} + L_{\lowerobj, 2}^{\lowervar, \uppervar} \left(4 \kappa_{\lowerobj, \lowervar} \frac{L_{\upperobj, 0}}{\mu_{\lowerobj}} + \varepsilon_{\dualvar, 0}\right)\right)^2   \\
                    & \qquad - 10272 \gamma^2 \kappa_{\lowerobj, \lowervar}^4 \frac{M_\cC^4}{\mu_{\lowerobj^2}} \constantdualdrift^2 \\
                    & \qquad + 300 \gamma^2 \kappa_{\lowerobj, \lowervar}^2 M_\cC^2  \; \constantdualdrift^2 
                                            \bigg[2 \kappa_{\lowerobj, \lowervar} \left(\frac{M_\cC}{\mu_{\lowerobj}}\right)^2 + 3 \bigg]  \Bigg) \|\apgstep_{\uppervar, t}\|^2
    \end{align*}}
    which simplifies to $\lyap_{t+1} \le \lyap_t - \frac{\gamma}{8} \|\pgstep_{\uppervar, t}\|^2 - \tfrac{\alpha}{4 \kappa_{\lowerobj,\lowervar}} \|\lowervar_t - \lowersol_t\|^2$ as soon as $\gamma \le \min(\frac{1}{\kappa_{\lowerobj, \lowervar}}, \bar \gamma)$, where 
    {\small \begin{align}\label{eq:def_bar_gamma_2}            
        \bar \gamma^{-1} = & 2{L_{\implicitobj}} + 48 \; \kappa_{\lowerobj, \lowervar} \left(\tfrac{M_{\cC}}{\mu_{\lowerobj}} \right)^2 \left(L_{\upperobj, 1} + L_{\lowerobj, 2}^{\lowervar, \uppervar} \left(4 \kappa_{\lowerobj, \lowervar} \frac{L_{\upperobj, 0}}{\mu_{\lowerobj}} + \varepsilon_{\dualvar, 0}\right)\right)^2 \notag \\
                    & + 10272 \; \kappa_{\lowerobj, \lowervar}^3 \frac{M_\cC^4}{\mu_{\lowerobj^2}} \constantdualdrift^2 \\
                    & + 300 \; \kappa_{\lowerobj, \lowervar} M_\cC^2  \; \constantdualdrift^2 
                                            \bigg[2\kappa_{\lowerobj, \lowervar} \left(\frac{M_\cC}{\mu_{\lowerobj}}\right)^2 + 3 \bigg]. \notag
    \end{align}}
\end{proof}

\subsection{Parameter policy for the stochastic setting}

\begin{lemma}\label{sup:lem:descent_lyap_stochastic}
    For any $t \ge 0$, it holds that
    {\small 
    \begin{align}\label{eq:lyap_descent_stochastic1}
    \lyap_{t+1} \le 
        \begin{aligned}[t]
            \lyap_t &- \expectation\left[\frac{\gamma_t}{8} \|\apgstep_{\uppervar, t}\|^2\right] - \expectation\left[\frac{\gamma_t}{8} \|\pgstep_{\uppervar, t}\|^2\right] - \frac{\alpha \tau \mu_{\lowerobj}}{4} \expectation[\|\lowervar_t - \lowersol_t\|^2] \\
                    &- \expectation\left[\constant_{D, t} \|\apgstep_{\uppervar, t}\|^2 \right] \\
                    &- \expectation\left[\constant_{\lowervar, t} \|\lowervar_t - \lowersol_t\|^2 \right] \\
                    &- \expectation\left[\constant_{\dualvar, t} \|\dualvar_t - \dualsol(\uppervar_t)\|^2 \right] \\
                    &- \expectation\left[\constant_{\lowervar, \dualvar, t} \|\dualvar_t - \dualsol(\uppervar_t)\|^2 \cdot \|\dualvar_t\|^2 \right] \\            
                    &+ \expectation[\constant_{\lyap_t}].
        \end{aligned}
    \end{align}}
    where the constants $\constant_{D, t}, \constant_{\lowervar, t}, \constant_{\dualvar, t}, \constant_{\lowervar, \dualvar, t}$ are defined as
    {\small \begin{align*}
    \constant_{D, t} 
        &\defineq \frac{\gamma_t}{4} 
            \begin{aligned}[t]
                \Bigg(1 &- 2 \gamma_t L_{\implicitobj} - 12 \alpha \gamma_t \tau^{-1} \mu_{\lowerobj}^{-3} (M_\cC)^2 \\
                        &-  108 \beta \gamma_t \rho^{-1} \mu_{\lowerobj}^{-1} L_{\dualvar, \star} \left(1 + 3 \tau^{-1} \mu_{\lowerobj}^{-3} M_{\cC}^2\right) \\
                        &- 12 \delta \gamma_t \tau^{-1} \mu_{\lowerobj}^{-3} M_{\cC}^2 \left(\frac{15}{\rho \mu_{\lowerobj}} \left(\frac{L_{\upperobj, 0}}{\mu_{\lowerobj}}\right)^2 + 2 \rho^2 \cV_{\upperobj, 1}^{\lowervar}\right) \Bigg)
            \end{aligned} \\
    \constant_{\lowervar, t} 
        &\defineq \alpha \frac{\tau \mu_{\lowerobj}}{4} - \frac{3\gamma}{2} (L_{\upperobj, 1})^2 - \beta \cdot 28 \rho^{-1} \mu_{\lowerobj}^{-1} \; L_{\dualvar, \star} \left(2 + \tau^2 \left(L_{\lowerobj, 1}^{\lowervar, \lowervar}\right)^2 \right) - \delta \left(\frac{15}{\rho \mu_{\lowerobj}} \left(\frac{L_{\upperobj, 0}}{\mu_{\lowerobj}}\right)^2
        + 2 \rho^2 \cV_{\upperobj, 1}^{\lowervar}\right) \\
    \constant_{\dualvar, t} 
        &\defineq \beta \frac{\rho \mu_{\lowerobj}}{2} - 3 \gamma M_{\cC}^2 \\
    \constant_{\lowervar, \dualvar, t}     
        &\defineq \delta \frac{\tau \mu_{\lowerobj}}{2} - 3 \gamma (L_{\lowerobj, 2}^{\lowervar, \lowervar})^2 \\
    \constant_{\lyap, t}
        &\defineq \begin{aligned}[t]
            & \frac{3\gamma}{4} \left(\cV_{\upperobj, 1}^{\uppervar} + \cV_{\lowerobj, 2}^{\lowervar, \uppervar} \; \constant_{\dualvar, \star}\right) + \alpha \tau^2 \frac{\kappa_{\lowerobj, \lowervar}}{\kappa_{\lowerobj, \lowervar} - 1} \cV_{\lowerobj, 1} \\
            & + \beta \left(27\; \rho^{-1} \mu_{\lowerobj}^{-1} \; L_{\dualvar, \star}\; \tau^2 \frac{2\kappa_{\lowerobj, \lowervar}}{\kappa_{\lowerobj, \lowervar} - 1} \cV_{\lowerobj, 1} + 4\; \rho^2 \left(\cV_{\upperobj, 1}^{\lowervar} + \cV_{\lowerobj, 2}^{\lowervar, \lowervar} \constant_{\dualvar, \star} \right)\right) \\
            & + \delta\; \begin{aligned}[t]
                & \Bigg( 3 \tau^{-1} \mu_{\lowerobj}^{-3} M_{\cC}^2 \expectation \left[ \expectation[\|\gamma_t \apgstep_{\uppervar, t}\|^2 \mid \mathcal{F}_t] \cdot \|v_t\|^2 \right]\\
                    &+ \tau^2 \frac{\kappa_{\lowerobj, \lowervar}}{\kappa_{\lowerobj, \lowervar} - 1} \cV_{\lowerobj, 1} \expectation \left[ \|v_t\|^2 \right]\\                
                    &+ \tau^2 \left(\frac{15}{\rho \mu_{\lowerobj}} \left(\frac{L_{\upperobj, 0}}{\mu_{\lowerobj}}\right)^2 + 2 \rho^2 \cV_{\upperobj, 1}^{\lowervar}\right) \frac{\kappa_{\lowerobj, \lowervar}}{\kappa_{\lowerobj, \lowervar} - 1} \cV_{\lowerobj, 1} \Bigg)        
            \end{aligned}  
        \end{aligned}
\end{align*}}
\end{lemma}
\begin{proof}  
First observe that, following the exact same steps as in the proof of Lemma~\ref{lem:implicit_descent}, we have for any $t \ge 0$,
{\small \begin{displaymath}
    \implicitobj(\uppervar_{t+1}) \le \implicitobj(\uppervar_{t}) - \frac{\gamma_t}{4} \|\pgstep_{\uppervar, t}\|^2 - \frac{\gamma_t}{8} \|\apgstep_{\uppervar, t}\|^2 + \frac{\gamma_t}{4} \|\pgstep_{\uppervar, t} - \apgstep{\uppervar, t}\|^2 +  \frac{\gamma_t}{2} \|\nabla \implicitobj(\uppervar_t) - \agrad_{\uppervar, t}\|^2 + \frac{\gamma_t^2 L_{\implicitobj}}{2} \|\apgstep_{\uppervar, t}\|^2.
\end{displaymath}}
Thus, taking the expectation and using Lemma~\ref{lem:implicit_gradient_error_stochastic}, we obtain
{\small \begin{align*}
    \expectation[\implicitobj(\uppervar_{t+1})]
        &\le
            \expectation[\implicitobj(\uppervar_{t})] 
            - \expectation\left[\frac{\gamma_t}{8} \|\apgstep_{\uppervar, t}\|^2] - \expectation[\frac{\gamma_t}{8} \|\pgstep_{\uppervar, t}\|^2\right] - \expectation\left[\frac{\gamma_t}{8}\left(1 - 4 \gamma_t L_{\implicitobj}\right) \|\apgstep_{\uppervar, t}\|\right]^2 \\
            &+ \expectation\left[\frac{3 \gamma_t}{4} \|\pgstep_{\uppervar, t} - \apgstep_{\uppervar, t}\|^2\right] \\
        & \le 
            \expectation[\implicitobj(\uppervar_{t})] 
            - \expectation\left[\frac{\gamma_t}{8} \|\apgstep_{\uppervar, t}\|^2\right] - \expectation\left[\frac{\gamma_t}{8} \|\pgstep_{\uppervar, t}\|^2\right] - \expectation\left[\frac{\gamma_t}{8}\left(1 - 4 \gamma_t L_{\implicitobj}\right) \|\apgstep_{\uppervar, t}\|\right]^2 \\
            & + \frac{3\gamma}{2} (L_{\upperobj, 1})^2 \expectation\left[\|\lowervar_t - \lowersol_t\|^2\right] \\
            & + 3 \gamma  M_\cC^2 \; \expectation\left[\|\dualvar_t - \dualsol(\uppervar_t)\|^2\right] \\
            & + 3 \gamma (L_{\lowerobj, 2}^{\lowervar, \uppervar})^2 \; \expectation\left[\|\lowervar_t - \lowersol_t\|^2 \cdot \|v_t\|^2\right] \\
            & + \frac{3\gamma}{4} \left(\cV_{\upperobj, 1}^{\uppervar} + \cV_{\lowerobj, 2}^{\lowervar, \uppervar} \; \constant_{\dualvar, \star}\right).
\end{align*}} 
where we used that $\gamma_t \defineq \gamma \min(1, \constant_\gamma / \|\apgstep_{\uppervar, t}\|) \le \gamma$.
Therefore, we may invoke Lemmas~\ref{lem:lower-level-descent-stochastic}, \ref{lem:dualsol_distance_stochastic} and~\ref{lem:expected_product} to obtain
{\small \begin{align*}
    \lyap_{t+1}       
    & \le
            \expectation[\implicitobj(\uppervar_{t})] 
            - \expectation\left[\frac{\gamma_t}{8} \|\apgstep_{\uppervar, t}\|^2\right] - \expectation\left[\frac{\gamma_t}{8} \|\pgstep_{\uppervar, t}\|^2\right] - \expectation\left[\frac{\gamma_t}{8}\left(1 - 4 \gamma_t L_{\implicitobj}\right) \|\apgstep_{\uppervar, t}\|\right]^2 \\
            & + \frac{3\gamma}{2} (L_{\upperobj, 1})^2 \expectation\left[\|\lowervar_t - \lowersol_t\|^2\right] \\
            & + 3 \gamma  M_\cC^2 \; \expectation\left[\|\dualvar_t - \dualsol(\uppervar_t)\|^2\right] \\
            & + 3 \gamma (L_{\lowerobj, 2}^{\lowervar, \uppervar})^2 \; \expectation\left[\|\lowervar_t - \lowersol_t\|^2 \cdot \|v_t\|^2\right] \\
            & + \frac{3\gamma}{4} \left(\cV_{\upperobj, 1}^{\uppervar} + \cV_{\lowerobj, 2}^{\lowervar, \uppervar} \; \constant_{\dualvar, \star}\right) \\
        & + \alpha \; 
                \Bigg( \left(1 - \tau \mu_{\lowerobj}/2\right) \expectation[\|\lowervar_t - \lowersol_t\|^2] + 3 \tau^{-1} \mu_{\lowerobj}^{-3} M_{\cC}^2 \expectation[\|\gamma_t \apgstep_{\uppervar, t}\|^2] + \tau^2 \frac{\kappa_{\lowerobj, \lowervar}}{\kappa_{\lowerobj, \lowervar} - 1} \cV_{\lowerobj, 1}\Bigg) \\
        & + \beta \;
                \Bigg( (1-\rho \mu_{\lowerobj}/2) \expectation[\|\dualvar_t - \dualsol(\uppervar_t)\|^2]
                + 28\; \rho^{-1} \mu_{\lowerobj}^{-1} \; L_{\dualvar, \star} \left(2 + \tau^2 \left(L_{\lowerobj, 1}^{\lowervar, \lowervar}\right)^2 \right) \expectation[\|\lowervar_t - \lowersol_t \|^2] \\
                &\qquad + 27\; \rho^{-1} \mu_{\lowerobj}^{-1} \; L_{\dualvar, \star} \; \left(1 + 3 \tau^{-1} \mu_{\lowerobj}^{-3} M_{\cC}^2 \right) \expectation[\|\gamma_t \apgstep_{\uppervar, t}\|^2]
                + 27\; \rho^{-1} \mu_{\lowerobj}^{-1} \; L_{\dualvar, \star} \tau^2 \frac{2\kappa_{\lowerobj, \lowervar}}{\kappa_{\lowerobj, \lowervar} - 1} \cV_{\lowerobj, 1} \\
                &\qquad + 4\; \rho^2 \left(\cV_{\upperobj, 1}^{\lowervar} + \cV_{\lowerobj, 2}^{\lowervar, \lowervar} \constant_{\dualvar, \star} \right) \Bigg) \\
        & + \delta
            \Bigg( \left(1 - \frac{\tau \mu_{\lowerobj}}{2}\right) \expectation \left[ \|\lowervar_t - \lowersol_t\|^2 \cdot \|v_t\|^2 \right]
            + 3 \tau^{-1} \mu_{\lowerobj}^{-3} M_{\cC}^2 \expectation \left[ \expectation[\|\gamma_t \apgstep_{\uppervar, t}\|^2 \mid \mathcal{F}_t] \cdot \|v_t\|^2 \right]\\
            &\qquad + \tau^2 \frac{\kappa_{\lowerobj, \lowervar}}{\kappa_{\lowerobj, \lowervar} - 1} \cV_{\lowerobj, 1} \expectation \left[ \|v_t\|^2 \right]
            + \left(\frac{15}{\rho \mu_{\lowerobj}} \left(\frac{L_{\upperobj, 0}}{\mu_{\lowerobj}}\right)^2
                + 2 \rho^2 \cV_{\upperobj, 1}^{\lowervar}\right)  \expectation \left[ \|\lowervar_t - \lowersol_t\|^2 \right]\\
            &\qquad + 3 \tau^{-1} \mu_{\lowerobj}^{-3} M_{\cC}^2  \left(\frac{15}{\rho \mu_{\lowerobj}} \left(\frac{L_{\upperobj, 0}}{\mu_{\lowerobj}}\right)^2
                + 2 \rho^2 \cV_{\upperobj, 1}^{\lowervar}\right)  \expectation \left[ \expectation[\|\gamma_t \apgstep_{\uppervar, t}\|^2 \mid \mathcal{F}_t] \right] \\
            &\qquad + \tau^2 \left(\frac{15}{\rho \mu_{\lowerobj}} \left(\frac{L_{\upperobj, 0}}{\mu_{\lowerobj}}\right)^2
                + 2 \rho^2 \cV_{\upperobj, 1}^{\lowervar}\right) \frac{\kappa_{\lowerobj, \lowervar}}{\kappa_{\lowerobj, \lowervar} - 1} \cV_{\lowerobj, 1} \Bigg).
\end{align*}}
where we used in several places $1-\rho \mu_\lowerobj/8 \le 1$ and $1 - \frac{\tau \mu_{\lowerobj}}{2} \le 1$.
After some rearrangement, we deduce that
{\small \begin{align}\label{eq:lyap_descent_stochastic2}
    \lyap_{t+1} \le 
        \begin{aligned}[t]
            \lyap_t &- \expectation\left[\frac{\gamma_t}{8} \|\apgstep_{\uppervar, t}\|^2\right] - \expectation\left[\frac{\gamma_t}{8} \|\pgstep_{\uppervar, t}\|^2\right] - \frac{\alpha \tau \mu_{\lowerobj}}{4} \expectation[\|\lowervar_t - \lowersol_t\|^2] \\
                    &- \expectation\left[\constant_{D, t} \|\apgstep_{\uppervar, t}\|^2 \right] \\
                    &- \expectation\left[\constant_{\lowervar, t} \|\lowervar_t - \lowersol_t\|^2 \right] \\
                    &- \expectation\left[\constant_{\dualvar, t} \|\dualvar_t - \dualsol(\uppervar_t)\|^2 \right] \\
                    &- \expectation\left[\constant_{\lowervar, \dualvar, t} \|\dualvar_t - \dualsol(\uppervar_t)\|^2 \cdot \|\dualvar_t\|^2 \right] \\            
                    &+ \expectation[\constant_{\lyap_t}].
        \end{aligned}
\end{align}}
where the quantities $\constant_{D,t}, \constant_{\lowervar, t}, \constant_{\dualvar, t}, \constant_{\lowervar, \dualvar, t}$ and $\constant_{\lyap, t}$ are defined as 
{\small \begin{align*}
    \constant_{D, t} 
        &\defineq \frac{\gamma_t}{4} 
            \begin{aligned}[t]
                \Bigg(1 &- 2 \gamma_t L_{\implicitobj} - 12 \alpha \gamma_t \tau^{-1} \mu_{\lowerobj}^{-3} (M_\cC)^2 \\
                        &-  108 \beta \gamma_t \rho^{-1} \mu_{\lowerobj}^{-1} L_{\dualvar, \star} \left(1 + 3 \tau^{-1} \mu_{\lowerobj}^{-3} M_{\cC}^2\right) \\
                        &- 12 \delta \gamma_t \tau^{-1} \mu_{\lowerobj}^{-3} M_{\cC}^2 \left(\frac{15}{\rho \mu_{\lowerobj}} \left(\frac{L_{\upperobj, 0}}{\mu_{\lowerobj}}\right)^2 + 2 \rho^2 \cV_{\upperobj, 1}^{\lowervar}\right) \Bigg)
            \end{aligned} \\
    \constant_{\lowervar, t} 
        &\defineq \alpha \frac{\tau \mu_{\lowerobj}}{4} - \frac{3\gamma}{2} (L_{\upperobj, 1})^2 - \beta \cdot 28 \rho^{-1} \mu_{\lowerobj}^{-1} \; L_{\dualvar, \star} \left(2 + \tau^2 \left(L_{\lowerobj, 1}^{\lowervar, \lowervar}\right)^2 \right) - \delta \left(\frac{15}{\rho \mu_{\lowerobj}} \left(\frac{L_{\upperobj, 0}}{\mu_{\lowerobj}}\right)^2
        + 2 \rho^2 \cV_{\upperobj, 1}^{\lowervar}\right) \\
    \constant_{\dualvar, t} 
        &\defineq \beta \frac{\rho \mu_{\lowerobj}}{2} - 3 \gamma M_{\cC}^2 \\
    \constant_{\lowervar, \dualvar, t}     
        &\defineq \delta \frac{\tau \mu_{\lowerobj}}{2} - 3 \gamma (L_{\lowerobj, 2}^{\lowervar, \lowervar})^2 \\
    \constant_{\lyap, t}
        &\defineq \begin{aligned}[t]
            & \frac{3\gamma}{4} \left(\cV_{\upperobj, 1}^{\uppervar} + \cV_{\lowerobj, 2}^{\lowervar, \uppervar} \; \constant_{\dualvar, \star}\right) + \alpha \tau^2 \frac{\kappa_{\lowerobj, \lowervar}}{\kappa_{\lowerobj, \lowervar} - 1} \cV_{\lowerobj, 1} \\
            & + \beta \left(27\; \rho^{-1} \mu_{\lowerobj}^{-1} \; L_{\dualvar, \star}\; \tau^2 \frac{2\kappa_{\lowerobj, \lowervar}}{\kappa_{\lowerobj, \lowervar} - 1} \cV_{\lowerobj, 1} + 4\; \rho^2 \left(\cV_{\upperobj, 1}^{\lowervar} + \cV_{\lowerobj, 2}^{\lowervar, \lowervar} \constant_{\dualvar, \star} \right)\right) \\
            & + \delta\; \begin{aligned}[t]
                & \Bigg( 3 \tau^{-1} \mu_{\lowerobj}^{-3} M_{\cC}^2 \expectation \left[ \expectation[\|\gamma_t \apgstep_{\uppervar, t}\|^2 \mid \mathcal{F}_t] \cdot \|v_t\|^2 \right]\\
                    &+ \tau^2 \frac{\kappa_{\lowerobj, \lowervar}}{\kappa_{\lowerobj, \lowervar} - 1} \cV_{\lowerobj, 1} \expectation \left[ \|v_t\|^2 \right]\\                
                    &+ \tau^2 \left(\frac{15}{\rho \mu_{\lowerobj}} \left(\frac{L_{\upperobj, 0}}{\mu_{\lowerobj}}\right)^2 + 2 \rho^2 \cV_{\upperobj, 1}^{\lowervar}\right) \frac{\kappa_{\lowerobj, \lowervar}}{\kappa_{\lowerobj, \lowervar} - 1} \cV_{\lowerobj, 1} \Bigg)        
            \end{aligned}  
        \end{aligned}
\end{align*}}
\end{proof}
\begin{lemma}\label{sup:lem:parameter_choices_stochastic}
    Assume that the step-sizes $\tau$ and $\rho$ are set as 
    {\small \[
        \tau=1/L_{\lowerobj, 1}^{\lowervar, \lowervar}, \rho=\min(1/L_{\lowerobj, 1}^{\lowervar, \lowervar}, \mu_{\lowerobj}/(4 \cV_{\lowerobj, 2}^{\lowervar, \lowervar}))
    \]}
    and $\gamma$ is set as $\gamma = \min\left(\bar \gamma, L_{\implicitobj}^{-1}\right)$, where
    {\small \begin{displaymath}
    {\bar\gamma}^{-1} \defineq \begin{aligned}[t]
                        2 \; L_{\implicitobj} 
                            &+ 72\; M_\cC^2 
                                 \kappa_{\lowerobj, \lowervar}^2 (L_{\upperobj, 1})^2 L_{\implicitobj}^{-1} \mu_{\lowerobj}^{-2} \\
                            &+ 32256 \;  L_{\dualvar, \star} M_\cC^4 
                                \kappa_{\lowerobj, \lowervar}^2 L_{\implicitobj}^{-1} \mu_{\lowerobj}^{-2} \left(\kappa_{\lowerobj, \lowervar} + 4 \mu_\lowerobj^{-2} \cV_{\upperobj, 1}^{\lowervar}\right)^2   \\     
                            &+ 288\; \kappa_{\lowerobj, \lowervar}^2  M_\cC^2 (L_{\lowerobj, 2}^{\lowervar, \lowervar})^2 L_{\implicitobj}^{-1} \mu_{\lowerobj}^{-2} \left(15 \left(\kappa_{\lowerobj, \lowervar} + 4 \mu_\lowerobj^{-2} \cV_{\upperobj, 1}^{\lowervar}\right) \left(\frac{L_{\upperobj, 0}}{\mu_{\lowerobj}}\right)^2 + 2  \frac{\cV_{\upperobj, 1}^{\lowervar}}{(L_{\lowerobj, 1}^{\lowervar, \lowervar})^2}\right) \Bigg) \\
                            &+ 648 \; L_{\dualvar, \star} M_{\cC}^2 L_{\implicitobj}^{-1}  \left(1 + 3\; \kappa_{\lowerobj, \lowervar} \mu_{\lowerobj}^{-2} M_{\cC}^2\right) \left(\kappa_{\lowerobj, \lowervar} + 4 \mu_\lowerobj^{-2} \cV_{\upperobj, 1}^{\lowervar}\right)^2 \\
                            &+ 72 \; \kappa_{\lowerobj, \lowervar}^2 (L_{\lowerobj, 2}^{\lowervar, \lowervar})^2 M_{\cC}^2 L_{\implicitobj}^{-1} \mu_{\lowerobj}^{-2}  \left(15 \left(\kappa_{\lowerobj, \lowervar} + 4 \mu_\lowerobj^{-2} \cV_{\upperobj, 1}^{\lowervar}\right) \left(\frac{L_{\upperobj, 0}}{\mu_{\lowerobj}}\right)^2 + 2\; \frac{\cV_{\upperobj, 1}^{\lowervar}}{(L_{\lowerobj, 1}^{\lowervar, \lowervar})^2}\right) 
            \end{aligned}
    \end{displaymath}}
    Assume furthermore that the Lyapynov parameters $\alpha, \beta$, and $\delta$ are set as:
    {\small \begin{align*}
    \alpha &= \gamma \begin{aligned}[t]
            \Bigg( 6\; \kappa_{\lowerobj, \lowervar} (L_{\upperobj, 1})^2  
                &+ 2688\; \kappa_{\lowerobj, \lowervar} \left(\kappa_{\lowerobj, \lowervar} + 4 \mu_\lowerobj^{-2} \cV_{\upperobj, 1}^{\lowervar}\right)^2 L_{\dualvar, \star} M_{\cC}^2   \\
                &+ 24\; \kappa_{\lowerobj, \lowervar} (L_{\lowerobj, 2}^{\lowervar, \lowervar})^2 \left(15 \left(\kappa_{\lowerobj, \lowervar} + 4 \mu_\lowerobj^{-2} \cV_{\upperobj, 1}^{\lowervar}\right) \left(\frac{L_{\upperobj, 0}}{\mu_{\lowerobj}}\right)^2 + 2  \frac{\cV_{\upperobj, 1}^{\lowervar}}{(L_{\lowerobj, 1}^{\lowervar, \lowervar})^2}\right) \Bigg)
        \end{aligned} \\
    \beta &= 6 \gamma \left(\kappa_{\lowerobj, \lowervar} + 4 \mu_\lowerobj^{-2} \cV_{\upperobj, 1}^{\lowervar}\right) M_{\cC}^2, \quad \delta = 6 \gamma \kappa_{\lowerobj, \lowervar} (L_{\lowerobj, 2}^{\lowervar, \lowervar})^2. 
    \end{align*}}
    Then, the constants $\constant_{D, t}, \constant_{\lowervar, t}, \constant_{\dualvar, t}, \constant_{\lowervar, \dualvar, t}$ defined in Lemma~\ref{sup:lem:descent_lyap_stochastic} are all non-negative.
\end{lemma}
\begin{proof}
    It suffices to plug the expressions given for $\tau, \rho, \gamma, \alpha, \beta,$ and $\delta$ into the definitions of the constants $\constant_{D, t}, \constant_{\lowervar, t}, \constant_{\dualvar, t}, \constant_{\lowervar, \dualvar, t}$ given in Lemma~\ref{sup:lem:descent_lyap_stochastic}.
    One should start first with $\constant_{\dualvar, t}$ and $\constant_{\lowervar, \dualvar, t}$, then $\constant_{\lowervar, t}$, and finally $\constant_{D, t}$.
\end{proof}

\begin{lemma}\label{sup:lem:constant_lyap_stochastic}
    Under this parameterization given in Lemma~\ref{sup:lem:parameter_choices_stochastic}, the remaining term $\expectation[\constant_{\lyap, t}]$ may be upperbounded as 
    {\small \begin{align*}
    \expectation[\constant_{\lyap, t}] \le \constant_{\lyap} 
        &\defineq \gamma 
        \begin{aligned}[t]
            & \Bigg[ \frac{3}{4} \left(\cV_{\upperobj, 1}^{\uppervar} + \cV_{\lowerobj, 2}^{\lowervar, \uppervar} \; \constant_{\dualvar, \star}\right)  \\
                    &+ 6\; (L_{\upperobj, 1})^2 (\kappa_{\lowerobj, \lowervar} - 1)^{-1} \mu_{\lowerobj}^{-2}\; \cV_{\lowerobj, 1} \\
                    &+ 2688 \; L_{\dualvar, \star} \left(\kappa_{\lowerobj, \lowervar} + 4 \mu_\lowerobj^{-2} \cV_{\upperobj, 1}^{\lowervar}\right)^2  M_{\cC}^2  (\kappa_{\lowerobj, \lowervar} - 1)^{-1} \mu_{\lowerobj}^{-2}\; \cV_{\lowerobj, 1}  \\
                    &+ 24 \; \frac{\kappa_{\lowerobj, \lowervar}^2}{\kappa_{\lowerobj, \lowervar} - 1} \left(15 \left(\kappa_{\lowerobj, \lowervar} + 4 \mu_\lowerobj^{-2} \cV_{\upperobj, 1}^{\lowervar}\right) \left(\frac{L_{\upperobj, 0}}{\mu_{\lowerobj}}\right)^2 + 2  \frac{\cV_{\upperobj, 1}^{\lowervar}}{(L_{\lowerobj, 1}^{\lowervar, \lowervar})^2}\right) \cV_{\lowerobj, 1} \\
                    &+ 324  \; L_{\dualvar, \star}\; M_{\cC}^2 \left(\kappa_{\lowerobj, \lowervar} + 4 \mu_\lowerobj^{-2} \cV_{\upperobj, 1}^{\lowervar}\right)^2 (\kappa_{\lowerobj, \lowervar} - 1)^{-1} \mu_{\lowerobj}^{-1} (L_{\lowerobj, 1}^{\lowervar, \lowervar})^{-1} \cV_{\lowerobj, 1}\\
                    &+ 24\; (L_{\lowerobj, 1}^{\lowervar, \lowervar})^{-1} \mu_{\lowerobj}^{-1} M_{\cC}^2 \left(\cV_{\upperobj, 1}^{\lowervar} + \cV_{\lowerobj, 2}^{\lowervar, \lowervar} \constant_{\dualvar, \star} \right) \\                    
                    &+ 6\; \cC_{\dualvar, \star} (L_{\lowerobj, 2}^{\lowervar, \lowervar})^2 (\kappa_{\lowerobj, \lowervar} - 1)^{-1} \mu_{\lowerobj}^{-2} \cV_{\lowerobj, 1} \\                
                    &+ 6 (L_{\lowerobj, 2}^{\lowervar, \lowervar})^2 \mu_{\lowerobj}^{-2} \left(15 \left(\kappa_{\lowerobj, \lowervar} + 4 \mu_\lowerobj^{-2} \cV_{\upperobj, 1}^{\lowervar}\right) \left(\frac{L_{\upperobj, 0}}{\mu_{\lowerobj}}\right)^2 + 2  \frac{\cV_{\upperobj, 1}^{\lowervar}}{(L_{\lowerobj, 1}^{\lowervar, \lowervar})^2}\right) (\kappa_{\lowerobj, \lowervar} - 1)^{-1} \cV_{\lowerobj, 1} \Bigg] \\
                    &+ 18\;\gamma^3 (L_{\lowerobj, 2}^{\lowervar, \lowervar})^2 \mu_{\lowerobj}^{-4} M_{\cC}^2 \cC_{\dualvar, \star} \cC_{\gamma}^2
        \end{aligned}         
\end{align*}}
\end{lemma}
\begin{proof}
    This directly follows again from plugging in the parameterization given in Lemma~\ref{sup:lem:parameter_choices_stochastic} into the definition of $\constant_{\lyap, t}$ given in Lemma~\ref{sup:lem:descent_lyap_stochastic}.
\end{proof}
}

\end{document}